\documentclass[preprint]{article}

\pdfoutput=1
\usepackage{lineno,hyperref}
\usepackage{nameref}
\modulolinenumbers[5]

\usepackage[T1]{fontenc}
\usepackage[htt]{hyphenat}
\usepackage{amsmath}
\usepackage{amsfonts}
\usepackage{multirow}
\usepackage{graphics}
\usepackage{graphicx}
\usepackage{adjustbox}
\usepackage{float}
\usepackage{booktabs, tabularx, rotating}
\usepackage{multirow}
\usepackage{subcaption}
\usepackage{changepage}
\usepackage{etoolbox,siunitx}
\usepackage{wrapfig}
\usepackage{xcolor}
\usepackage{authblk}
\robustify\bfseries

\newcommand{\etal}{\textit{et al}. }
\newcommand{\ie}{i.\,e.\ }

\begin{document}


\title{Recent Trends in the Use of Statistical Tests for Comparing Swarm and Evolutionary Computing Algorithms: Practical Guidelines and a Critical Review}

\author[1]{J.~Carrasco}
\author[1]{S.~Garc\'{i}a}
\author[2]{M.M.~Rueda}
\author[3]{S.~Das}
\author[1]{F.~Herrera}

\affil[1]{\small{Department of Computer Science and AI, Andalusian Research Institute in Data Science and Computational Intelligence, University of Granada, Granada, Spain}}

\affil[2]{Department of Statistic and Operational Research,
  University of Granada, Andalusian Research Institute of Mathematics,
  Granada, Spain}

\affil[3]{Electronics and Communication Sciences Unit, Indian
  Statistical Institute, 203 B.T.Road, Kolkata 700108, West Bengal,
  India}

\maketitle

\begin{abstract}
  A key aspect of the design of evolutionary and
  swarm intelligence algorithms is studying their
  performance. Statistical comparisons are also a crucial part which
  allows for reliable conclusions to be drawn. In the present paper we
  gather and examine the approaches taken from different perspectives
  to summarise the assumptions made by these statistical tests, the
  conclusions reached and the steps followed to perform them
  correctly. In this paper, we conduct a survey on the current trends
  of the proposals of statistical analyses for the comparison of
  algorithms of computational intelligence and include a description
  of the statistical background of these tests. We illustrate the use
  of the most common tests in the context of the Competition on
  single-objective real parameter optimisation of the IEEE Congress on
  Evolutionary Computation (CEC) 2017 and describe the main advantages
  and drawbacks of the use of each kind of test and put forward some
  recommendations concerning their use.

\end{abstract}



\section{Introduction}
\label{sec:introduction}
Over the few last years the comparison of evolutionary optimisation
algorithms and statistical analysis have undergone
some changes. The classic paradigm consisted of the application of
classic frequentist tests on the final results over a set of benchmark
functions, although different trends have been proposed since then
\cite{2019-Hellwig-Benchmarkingevolutionaryalgorithms}, such as:

\begin{itemize}
\item The first amendment after the popularisation of the use of
  statistical tests was the proposal of non-parametric tests that
  consider the underlying distribution of the analysed results,
  improving the robustness of the drawn conclusions
  \cite{2008-Demsar-appropriatenessstatisticaltests}.
\item With this new perspective, some non-parametric tests
  \cite{2003-Sheskin-Handbookparametricnonparametric}, whose
  assumptions were less restrictive than the previous ones, were
  suggested for the comparison of computational intelligence
  algorithms \cite{2009-Garcia-studyusenonparametric}. 
\item However, there are other approaches and considerations, like the
  convergence of the solution
  \cite{2014-Derrac-Analyzingconvergenceperformance}, robustness with
  respect to the seed and results over different runs and the
  computation of the confidence interval and confidence curves
  \cite{2017-Berrar-Confidencecurvesalternative}.
\item As has already occurred in several research fields, a Bayesian
  trend \cite{2003-Gelman-BayesianDataAnalysis} has emerged with some
  criticisms to the well known Null Hypothesis Statistical Tests
  (NHST) and there are some interesting proposals of Bayesian tests
  analogous to the classic frequentist tests
  \cite{2017-Benavoli-TimeChangeTutorial}.
\end{itemize}

Inferential statistics make predictions and obtain conclusions from
data, and these predictions are the basis for the performance
comparisons made between algorithms
\cite{2011-Japkowicz-EvaluatingLearningAlgorithms}. The procedure
followed to reach relevant information is detailed below:

\begin{enumerate}
\item The process begins with the results of the runs from an
  algorithm in a single benchmark function. These
    results assume the role of a sample from an unknown distribution
  whose parameters we can just estimate to compare
  it with another algorithm's distribution.
\item Depending on the nature and purpose of the test, the results of
  the algorithms involved in the comparison will be aggregated in
  order to compute a statistic.
\item The statistic is used as an estimator of a characteristic,
  called parameter, of the distribution of interest, either the
  distribution of the results of our algorithm or the distribution of
  the algorithms' performance difference when we are comparing
  the results from a set of algorithms.
\end{enumerate}

This procedure allows us to get information from experimental results,
although it must be followed in order to obtain impartial conclusions
which can also be reached by other researchers. Statistical tests
should be considered as a toolbox to collect relevant information, not
as a set of methods to confirm the previously stated conclusion. There
are two main approaches:
\begin{description}
\item[Frequentist] These are the most common tests. Both Parametric
  and Non-Parametric tests follow this approach. Here, a non-effect
  hypothesis (in the sense of non-practical
  differences between algorithms) $\mathcal{H}_{0}$ and an alternative
  hypothesis $\mathcal{H}_{1}$ are set up and a test, known as Null
  Hypothesis Statistical Test (NHST) is performed. With the results of
  this test, we determine if we should reject the null hypothesis in
  favour of the alternative one or if we do not have enough evidence
  to reject the null hypothesis. This decision is made according to
  two relevant concepts in the Frequentist paradigm
  \cite{2013-Odile-StatisticalTestsNonparametric,
    2003-Sheskin-Handbookparametricnonparametric,
    2010-Gibbons-NonparametricStatisticalInference}:
  \begin{itemize}
  \item $\alpha$ or confidence coefficient: While estimating the
    difference between the populations, there is a certain confidence
    level about how likely the true statistic is to lie in the range
    of estimated differences between the samples. This confidence
    level is denoted as $(1-\alpha)$, where $\alpha \in [0,1]$, and
    usually $\alpha = 0.05$. This is not the same situation as if the
    probability of the parameter of interest lay in
    the range of estimated differences, but the percentage of times
    that the true parameter would lie in this interval if repeated
    samples from the population had been extracted.
  \item $p$-value: Given a sample $D$ and considering a null
    hypothesis $\mathcal{H_0}$, the associated $p$-value is the
    probability of obtaining a new sample as far from the null
    hypothesis as the collected data. This means that if the gathered
    data is not consistent with the assumed hypothesis, the obtained
    probability will be lower. Then, in the context of hypothesis
    testing, $\alpha$ represents the established threshold which the
    $p$-value is compared with. If the $p$-value is lower than this
    value, the null hypothesis is rejected.
  \item The confidence interval, the counterpart of the $p$-values,
    represents the certainty about the difference between the samples
    at a fixed confidence level. Berrar proposes the confidence
    curves, as a graphic representation that generalise the confidence
    intervals for every confidence level
    \cite{2017-Berrar-Confidencecurvesalternative}.
  \end{itemize}
\item[Bayesian] Here, we do not compute a single probability but a
  distribution of the parameter of interest itself. With this
  approach, we avoid some main drawbacks of NHST, although they
  require a deeper understanding of the underlying statistics and
  conclusions are not as direct as in frequentist tests.
\end{description}

\begin{figure}
  \centering
  \includegraphics[width=.9\textwidth]{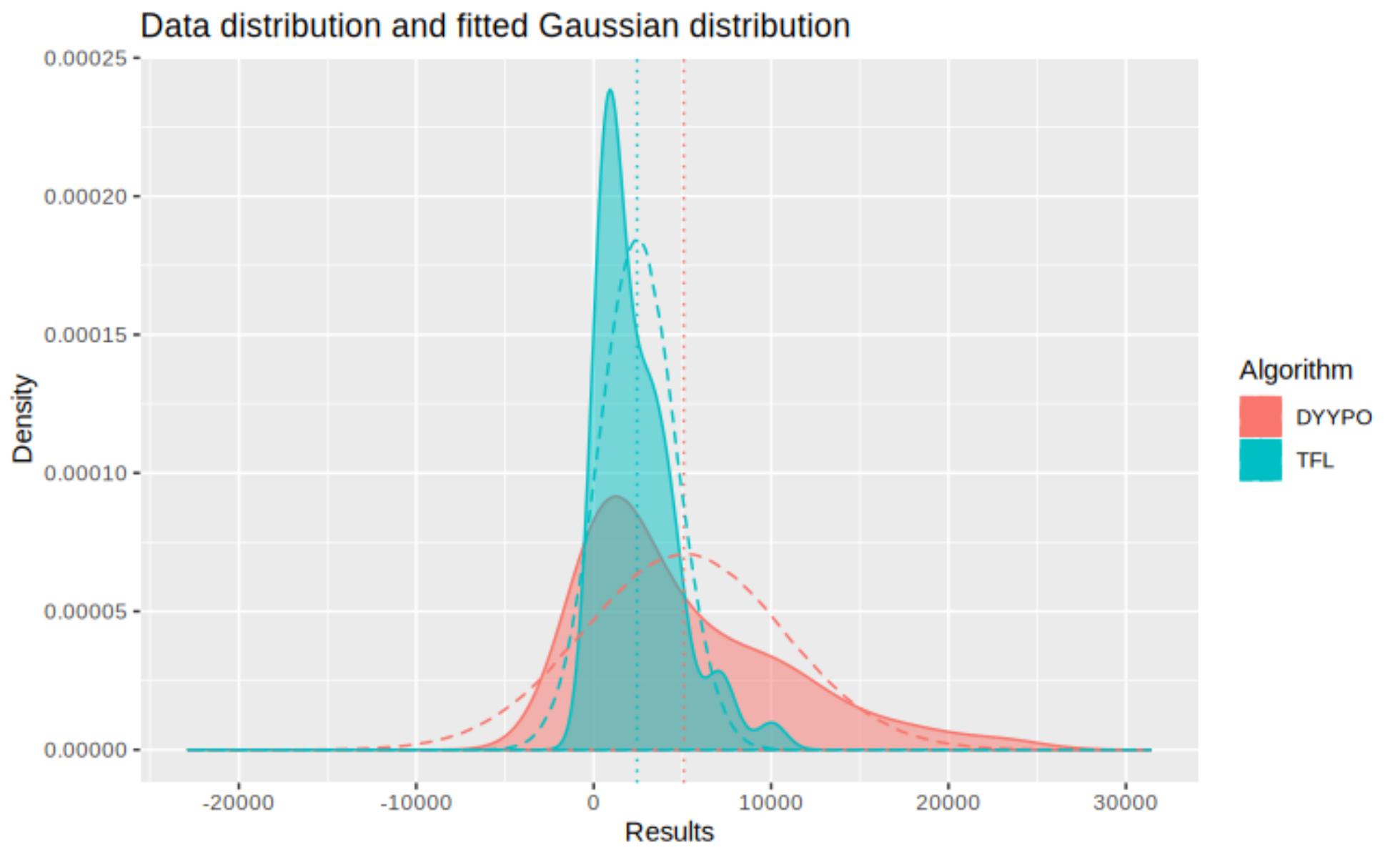}
  \caption[Gaussian distribution]{Comparison between real and fitted
    Gaussian distribution of results.}
  \label{fig:gaussian}
\end{figure}

The main concepts of these families of statistical tests can be
explained using \autoref{fig:gaussian}. In this figure we have plotted
the density of distribution of the results of the different runs of
two algorithms (DYYPO and TLBO-FL, which are presented in
\autoref{sec:cont-algor}) for a single benchmark. The density
represents the relative probability of the random variable of the
results of these algorithms for each value in the x-axis. In this
scenario, a Parametric NHST would set up a null hypothesis about the
means of both populations (plotted with a dotted line), assuming that
their distribution follows a Gaussian distribution. We have also
plotted the Gaussian distributions with the mean and standard
deviation of each population. In this context, there is a difference
between the two means and the test could reject the null
hypothesis. However, if we consider the estimated distributions, they
are overlapped. This overlapping lead to the concept of \textit{effect
  size}, which indicates not only if the population means can be
considered as different, but quantifies this difference. In
\autoref{sec:confidence-curves} and \autoref{sec:null-hypoth-stat} we
will go in-depth in the study of this issue.

Nonetheless, there is a difference between the estimated Gaussian
distributions and the real densities. This is the reason why
Non-Parametric Tests arose as an effective alternative to
their parametric counterparts, as they do not
suppose the normality of the input data. In the
  Bayesian procedure, we would make an estimation of the distribution
  of the differences between the algorithms' results as we have made
  with the results themselves in \autoref{fig:gaussian}. Then, with
  the estimated distribution of the parameter of interest, \ie the
  difference between the performances of the algorithms, we could extract the
  desired information.

Bayesian paradigm makes statements about the distribution of the
difference between the two algorithms, which can
reduce the burden of the researchers the NHST do not
find significant differences between the algorithms.

Other significant concepts in the description of statistical tests
come with the comparison of the tests themselves. Type I error occurs
when $\mathcal{H}_{0}$ is rejected but it should not be. In our
scenario, this means that the equivalence of the algorithms has been
discarded although there is not a significant difference between
them. Otherwise, type II error arises when there is a significant
difference but this has not been detected. We denote $alpha$ as the
probability of making a type I error (often called significance level
or size of test) and $beta$ as the probability of make a type II
error. The power of a test is $1 - \beta$, \ie the probability of
rejecting $\mathcal{H}_{0}$ when it is false, so we are interested in
comparing the power of the tests, because with the same probability of
making a type I error, the more powerful a test is, the more
differences that will be regarded as significant.

There is some notation that is used along this work and it is
summarised in \autoref{tab:notation}.

\begin{table}[H]
  \centering
  \begin{tabular}{rp{7cm}l}
    Notation & Description & Location \\ \hline 
    $\mathcal{H}_0$ & Null hypothesis & \autoref{sec:frequentist-tests} \\
    $\mathcal{H}_1$ & Alternative hypothesis & \autoref{sec:frequentist-tests} \\
    $\mathcal{H}_j$ & When comparing multiple algorithms, each one of
                      the multiple comparisons. &  Subsection \ref{sec:post-hoc-procedures}\\
    $\mathcal{H}_{i,j}$ & When comparing multiple algorithms, the
                          comparison between the algorithm $i$ and
                          $j$. & Subsection \ref{sec:post-hoc-procedures}\\
    $\alpha$ & Significance level & \autoref{sec:frequentist-tests} \\
    $\mu$ & Mean of the performance of an algorithm & Section \ref{sec:param-stat-tests} \\
    $k$ & Number of compared algorithms &
                                          Sections~\ref{sec:frequentist-tests},
                                          \ref{sec:bayesian-tests} \\
    $N_m(\mu, \Sigma)$ & Multivariate Gaussian distribution, with
                         parameters $\mu$ and $\Sigma$ & \autoref{sec:frequentist-tests} \\
    $n$ & Number of benchmark functions & Sections~\ref{sec:frequentist-tests},\ref{sec:bayesian-tests} \\
    $\theta$ & Probability of stated hypothesis & \autoref{sec:bayesian-tests} \\
    $Dir$ & Dirichlet distribution & \autoref{sec:bayesian-tests} \\
    DP & Dirichlet Process & \autoref{sec:bayesian-tests} \\
    $\alpha$ & Prior distribution of DP & \autoref{sec:bayesian-tests} \\
    $\mathcal{E}$ & The expectation with respect to the Dirichlet
                    Process. & Subsection \ref{sec:impr-dirichl-proc} \\
    $\delta_z$ & Dirac's delta centered in $z$ &
                                                 \autoref{sec:bayesian-tests} \\
    $I_{[precondition]}$ & Indicator function. Takes 1 when $precondition$
                           is fulfilled, 0 otherwise. & \autoref{sec:bayesian-tests} \\
    \hline
  \end{tabular}
  \caption{General notations used in this paper}
  \label{tab:notation}
\end{table}

In this paper, we gather the different points of view of the
statistical analysis of results in the computational intelligence
research field with the associated theoretical background and the
properties of the tests. This allows comparisons to be made regarding
the appropriateness of the use of each testing paradigm, depending on
the specific situation. Moreover, we also adapt the calculation of
these confidence curves with a non-parametrical procedure
\cite{1988-Campbell-StatisticsMedicineCalculating}, as it is more
convenient in the context of the comparison of evolutionary
optimisation algorithms.

We have included in this paper the performance and the analyses of the
tests in the context of the 2017 IEEE Congress on Evolutionary
Computation (CEC) Special Session and Competition on Single Objective
Real Parameter Numerical Optimisation
\cite{2016-Awad-ProblemDefinitionsEvaluation}. This
case study allows for a clear illustration of the
use of these tests, their behaviour with different distributions of
data and the conclusions that can be made with their outputs. The
test-suite proposed for this competition can be considered to be a
relevant benchmark for the comparison of any newly proposed
algorithms. Moreover, the results achieved and final rankings of this
competition are supposed to indicate what is the
state-of-the-art in evolutionary
  computation and against which algorithms our proposals should be
  compared.

The code for the tests and the analysis, tables and plots are included
as a vignette in the developed \texttt{R} package \texttt{rNPBST}
\footnote{\url{https://github.com/JacintoCC/rNPBST}}
\cite{2017-Carrasco-rNPBSTPackageCovering}. We have also developed a
\textit{shiny} application to facilitate the use of the aforementioned
tests. This application processes the results of two or more
algorithms and performs the selected test. The results are exported in
\TeX format and an HTML table. Cases in which there
is a plot associated with the test are highlighted.

This paper is organised as follows. In \autoref{sec:surv-all-stat} we
include a survey on the main statistical analyses
  proposed in the literature for evaluating the classification and
  optimisation algorithms. \autoref{sec:frequentist-tests} contains a
depiction of the well-known parametric tests and non-parametric tests
respectively. In \autoref{sec:null-hypoth-stat} we describe the main
criticisms and proposals concerning the traditional
tests. \autoref{sec:bayesian-tests} introduces Bayesian test concepts
and notations. Due to the relevance of the Multi-Objective problem in
the optimisation scientific community, we have gathered the tests that
address this issue in \autoref{sec:mult-meas-tests} regardless of
their statistical nature. In \autoref{sec:exper-fram}, we describe the
setting of the CEC'2017 Special Session, which is performed in
\autoref{sec:experiments-results} and summarised in
\autoref{sec:summ-results-cec2017}. The lessons learnt and test
considerations are presented in \autoref{sec:misc-lessons-learnt}.
\autoref{sec:conclusions} summarises the conclusions obtained in the
analyses.

\section{Survey on Statistical Analyses Proposed}
\label{sec:surv-all-stat}

In this section, we provide a chart with an
extensive survey on the different statistical proposals made for the
comparison of machine learning and optimisation algorithms. The survey
is included in \autoref{tab:survey}. This table includes a
categorisation of the methodologies according to their statistic
nature, a brief description of the underlying idea and the considered
scenario of the comparison of the data.

The proposals are sorted by the year of publication. This order
highlights past and present trends in the statistical comparison. The
first proposals were made from a frequentist and fundamentally
parametric point of view. Later, with the work of Dem\v{s}ar
\cite{2006-Demsar-Statisticalcomparisonsclassifiers} non-parametric
tests arose as the alternative to parametric ones in certain
circumstances. In recent years, some Bayesian proposals were made,
although they still depend on the results' distributions and are
focused on the comparison of classifiers.

Moreover, most of the proposals are oriented to compare
classifiers. The tests and guidelines made for the comparison of
optimisation algorithms are consistent and are constantly used in the
literature. However, there is not a specific test that takes account
of the multiple runs in the same benchmark function and estimates the
correlation between these runs, as the tests suggested for
cross-validation setups do.

\begin{table}[H]
  \centering
\resizebox*{!}{\textheight}{%
  \begin{tabular}{|llp{12cm}|}
    \hline
    Citation & Kind of test & Description \\ \hline
    \cite{1988-Looney-statisticaltechniquecomparing}
             & Non-parametric - Classification
                            & Cochran Test for the distribution of mistakes on the
                              classification of $n$ samples.
    \\ \hline
    \cite{1998-Dietterich-Approximatestatisticaltests}
             & Parametric - Classification
                            & McNemar Test and $t$-test variants for
                              different scenarios of the comparison of two
                              classifiers accuracy. 
    \\ \hline
    \cite{1999-Alpaydin-CombinedcvTest}
             & Parametric - Classification
                            & Proposal of 5$\times$2 cv F test. \\ \hline
    \cite{2002-Castillo-Valdivieso-Statisticalanalysisparameters}
             & Parametric - Optimisation
                            & ANOVA test for parameter analysis in
                              genetic algorithms.
    \\ \hline
    \cite{2002-Pizarro-Multiplecomparisonprocedures}
             & Frequentist - Classification
                            & Introduces corrections for multiple testing (Bonferroni, Tukey, Dunnet, Hsu).
    \\ \hline
    \cite{2003-Nadeau-Inferencegeneralizationerror}
             & Parametric - Classification
                            & Proposal using variance estimators of
                              cross-validation results.
    \\ \hline
    \cite{2003-Chen-StatisticalComparisonsMultiple}
             & Non-parametric - Classification
                            & Proposal of multiple comparisons using
                              Cochran test.
    \\ \hline
    \cite{2004-Czarn-Statisticalexploratoryanalysis}
             & Parametric - Optimisation
                            & Guide on the use of ANOVA test in
                              exploratory analyses of genetics algorithms.
    \\ \hline
    \cite{2006-Moskowitz-Comparingpredictivevalues}
             & Parametric - Classification
                            & Computation of the sample size for a
                              desired confidence interval width of the
                              True Positive Rate and True Negative Rate.
    \\ \hline
    \cite{2006-Yildiz-Orderingfindingbesta}
             & Parametric - Classification
                            & Proposal of MultiTest algorithm that order the
                              competitors using post-hoc test.
    \\ \hline
    \cite{2006-Demsar-Statisticalcomparisonsclassifiers}
             & Frequentist - Classification
                            & Review of the use of previous parametric
                              tests, Wilcoxon Signed-Rank and Friedman test. 
    \\ \hline
    \cite{2007-Smucker-comparisonstatisticalsignificance}
             & Frequentist - Information Retrieval
                            &  Comparison of pairwise non-parametric
                              tests with Student's paired t-test.
    \\ \hline
    \cite{2008-Demsar-appropriatenessstatisticaltests}
             & Frequentist - Classification 
                            & Criticisms and tips on the use of
                              statistical tests. 
    \\ \hline
    \cite{2008-Garcia-ExtensionStatisticalComparisons}
             & Non-parametric - Classification
                            & Extensive proposal of non-parametric
                              tests and associated post-hoc methods.
    \\ \hline
    \cite{2009-Aslan-Statisticalcomparisonclassifiers}
             & Parametric - Classification
                            & $k$-fold cross validated paired $t$-test
                              for AUC values.
    \\ \hline
    \cite{2009-Garcia-studyusenonparametric}
             & Non-parametric - Optimisation
                            & Study of the preconditions for a safe use
                              of the parametric tests and
                              proposal of non-parametric methods for an
                              optimisation scenario.
    \\ \hline
    \cite{2009-Garcia-studystatisticaltechniques}
             & Non-parametric - Classification
                            & Study of test application prerequisite in
                              a classification context and with
                              different measures.
    \\ \hline
    \cite{2009-Luengo-studyusestatistical}
             & Non-parametric - Classification
                            & Study on the use of statistical tests in
                              neural networks' results.
    \\ \hline
    \cite{2010-Garcia-Advancednonparametrictests}
             & Non-parametric - Classification 
                            & Newly proposed post-hoc methods and
                              non-parametric comparison
                              (Li, Holm, Holland,
                              Finner, Hochberg, Hommel and Rom post-hoc procedures).
    \\ \hline
    \cite{2010-Westfall-MultipleMcNemarTests}
             & Parametric - Classification
                            & Repeated McNemar's test.
    \\ \hline
    \cite{2010-Rodriguez-SensitivityAnalysiskFold}
             & Frequentist - Classification
                            & Decomposition of the variance of the
                              $k$-fold CV for prediction error
                              estimation.  
    \\ \hline
    \cite{2010-Ojala-Permutationtestsstudying}
             & Permutational - Classification
                            & Presentation of two permutational tests
                              that study if the algorithm has learnt
                              the data structure and if it uses the
                              attributes distribution and dependencies.
    \\ \hline
    \cite{2011-Carrano-MulticriteriaStatisticalBased}
             & Parametric - Optimisation
                            & A multicriteria comparison
                              algorithm (MCStatComp)
                              using the aggregation of the criteria through a non-dominance analysis.
    \\ \hline
    \cite{2011-Derrac-practicaltutorialuse}
             & Non-parametric - Optimisation
                            & Tutorial on the use of non-parametric
                              tests and post-hoc procedures.
    \\ \hline
    \cite{2012-Trawinski-Nonparametricstatisticalanalysis}
             & Non-parametric
                            & Review on the use of non-parametric
                              tests and post-hoc
                              tests, and the impact
                              of normality.
    \\ \hline
    \cite{2012-Ulas-Costconsciouscomparisonsupervised}
             & Non-parametric - Classification
                            & Multi$^2$Test
                              algorithm, which orders the algorithms
                              using non-parametric pairwise tests.
    \\ \hline
    \cite{2012-Irsoy-DesignAnalysisClassifier}
             & Non-parametric - Classification
                            & Survey on the use of statistical tests
                              in Bioinformatics field.
    \\ \hline 
    \cite{2012-Lacoste-Bayesiancomparisonmachine}
             & Bayesian - Classification
                            & Bayesian Poisson binomial test for pairwise comparison of classifiers.
    \\ \hline 
    \cite{2012-Brodersen-Bayesianmixedeffectsinference}
             & Bayesian - Classification
                            & Hierarchical study through Bayesian inference.
    \\ \hline 
    \cite{2013-Yildiz-StatisticalTestsUsing}
             & Parametric - Classification
                            & Inclusion of a certain level of allowed error in paired
                              $t$-test.
    \\ \hline
    \cite{2013-Bostanci-EvaluationClassificationAlgorithms}
             & Parametric - Classification
                            & Comparison using McNemar test with
                              different measures. 
    \\ \hline
    \cite{2014-Otero-Bootstrapanalysismultiple}
             & Permutation - Classification
                            & Permutation (bootstrap) tests for a cross-validation
                              setup.
    \\ \hline
    \cite{2014-Yu-Blocked3x2CrossValidated}
             & Parametric - Classification
                            & Blocked $3 \times 2$ cross validation
                              estimator of variance.
    \\ \hline
    \cite{2014-Derrac-Analyzingconvergenceperformance}
             & Non-parametric - Optimisation
                            & Analysis of convergence using Page test.
    \\ \hline
    \cite{2014-Garcia-statisticalanalysisparameters}
             &  Non-parametric - Classification
                            & Proposal of Page test for parameter
                              trend study.
    \\ \hline
    \cite{2014-Benavoli-BayesianWilcoxonsignedrank}
             & Bayesian
                            & Bayesian version of Wilcoxon Signed Rank
                              test. 
    \\ \hline
    \cite{2015-Benavoli-ImpreciseDirichletProcess}
             & Bayesian
                            & Bayesian test without prior
                              information using
                              Imprecise Dirichlet Process.
    \\ \hline
    \cite{2015-Corani-Bayesianapproachcomparing}
             & Bayesian - Classification
                            & Bayesian test for two classifiers on
                              multiple data sets accounting the
                              correlation of cross-validation. 
    \\ \hline
    \cite{2015-Benavoli-StatisticalTestsJoint}
             & Bayesian
                            & Proposal of Multiple Measures tests.
    \\ \hline
    \cite{2015-Benavoli-Bayesiannonparametricprocedure}
             & Bayesian
                            & Presentation of the Bayesian Friedman test.
    \\ \hline
    \cite{2015-Wang-ConfidenceIntervalMeasure}
             & Parametric - Information retrieval
                            & Confidence interval for $F_{1}$ measure
                              using blocked $3 \times 2$ cross
                              validation.
    \\ \hline
    \cite{2015-Perolat-GeneralizingWilcoxonranksum}
             & Non-parametric - Classification
                            & Generalisation of Wilcoxon rank-sum test
                              for interval data.
    \\ \hline
    \cite{2015-Singh-Statisticalvalidationmultiple}
             & Non-parametric - Classification
                            & Proposal off Mann-Whitney U test for two
                              classifiers and the Kruskal-Wallis H
                              test for multiple classifiers with the
                              associated post-hoc corrections. 
    \\ \hline
    \cite{2016-Gondara-Classifiercomparisonusing}
             & Frequentist - Classification 
                            & Proposal of Wald and Score tests for
                              precision comparison.
    \\ \hline
    \cite{2016-Corani-Statisticalcomparisonclassifiers}
             & Bayesian - Classification
                            & Bayesian hierarchical model for the joint
                              comparison of multiple classifiers on
                              multiple data sets with the
                              cross-validation results. 
    \\ \hline
    \cite{2017-Berrar-Confidencecurvesalternative}
             & Parametric - Classification
                            & Presentation of the confidence curves as the
                              confidence interval generalisation.
    \\ \hline
    \cite{2017-Eisinga-Exactpvaluespairwise}
             & Non-parametric - Classification
                            & Proposal of exact computation of
                              Friedman test. 
    \\ \hline
    \cite{2017-Benavoli-TimeChangeTutorial}
             & Bayesian
                            & Extensive tutorial on the use of
                              Bayesian tests.
    \\ \hline
    \cite{2017-Yu-NewKindNonparametric}
             & Non-parametric - Classification
                            & New proposals of non-parametric tests
                              that introduce weights.
    \\ \hline
    \cite{2017-Eftimov-Comparingmultiobjectiveoptimization}
             & Frequentist - Optimisation
                            & Application of Deep Statistical
                              Comparison of Multi-Objective
                              Optimisation algorithms for an ensemble
                              of quality indicators. 
    \\ \hline
    \cite{2018-Calvo-BayesianInferenceAlgorithm}
             & Bayesian
                            & Bayesian analysis based on a model over
                              the algorithms' rankings.
    \\ \hline
    \cite{2019-Campelo-Samplesizeestimation}
             & Parametric - Classification
                            & Methodology for the definition of the
                              sample sizes. 
    \\ \hline
    \cite{2019-Eftimov-novelstatisticalapproach}
             & Frequentist - Optimisation
                            & Extension of Deep
                              Statistical Comparison, a two-step comparison that select the
                              appropiate parametric or non-parametric
                              test according to the normality of the
                              data. 
    \\ \hline
  \end{tabular}}%
\caption{Survey on different statistical proposals for results analysis}
\label{tab:survey}
\end{table}

\section{Frequentist tests}
\label{sec:frequentist-tests}

In this section, we describe the classic frequentist tests: the
properties of the parametric tests and their assumptions, and the
different non-parametric tests and their application in the context of
the comparison of single-objective and
  bound-constrained evolutionary optimisation algorithms. Although
there are other tests, algorithms and proposals, as reflected in
\autoref{tab:survey}, the tests presented in this section represent the
core of the statistical comparison methodology.

\subsection{Parametric Statistical Tests}
\label{sec:param-stat-tests}

Parametric tests make the assumption that our sample comes from a
distribution that belongs to a known family, usually the Gaussian
family, and it is described with a little number of parameters
\cite{2003-Sheskin-Handbookparametricnonparametric}. 

\paragraph{t-test}

This classic test is used to compare two samples. Null hypothesis
consists in the equivalence of the means of both populations. The main
assumptions made by t-test is that the samples have been extracted
randomly and the distribution of the populations of the samples are
normal.

The required input of this test is the group of observations of the
different runs of the pair algorithms that will be compared
for a single problem. 

\paragraph{Analysis of Variance}

When we are interested in the comparison of $k$ distinct algorithms,
we need another test, because repeating the t-test for every pair of
algorithms would increment the type I error. The
  Analysis of Variance (ANOVA) test of the null hypothesis consists
in the equivalence of all the means:
\[ \mathcal{H}_0: \mu_1 = \dots = \mu_k, \] against
$\mathcal{H}_1: \exists i \neq j,\ \mu_i \neq \mu_j$. ANOVA test deals
with the variance within a group, between groups or a combination of
the two types.

Here the input consists of a matrix where each column represents an
algorithm and each row is a single benchmark function, while the cells
contain the mean of the performance for all runs of each algorithm in
each benchmark
\cite{2002-Castillo-Valdivieso-Statisticalanalysisparameters}.

These tests are very relevant in the statistical comparison, although
they have troublesome prerequisites in the field of comparison of
optimisation algorithms. Then, non-parametric statistical tests were
proposed to address this issue with a known
methodology.

\subsection{Non-Parametrical Statistical Tests}
\label{sec:non-param-stat}

According to Pesarin \cite{2010-Pesarin-Permutationtestscomplex},
$\mathcal{P}$ is a non-parametric family of distributions if it is not
possible to find a finite-dimensional space $\Theta$ in which there is
a one-to-one relationship between $\Theta$ and $\mathcal{P}$. This
means that we do not have to assume that the
underlying distribution belongs to a known family of
distributions. Consequently, the prerequisites for non-parametric
tests such as symmetry or continuity, are less restrictive than
parametric ones and non-parametric tests are more robust and less
sensitive to dirty data
\cite{2006-Demsar-Statisticalcomparisonsclassifiers,
  2008-Garcia-ExtensionStatisticalComparisons}.

\subsubsection{Check of the preconditions}
\label{sec:check-preconditions}

The goodness of fit tests, for example, Kolmogorov-Smirnov,
Shapiro-Wilk or D'Agostino-Pearson, are used to determine whether the
normality of the distribution can be rejected
\cite{2003-Sheskin-Handbookparametricnonparametric,
  2009-Garcia-studyusenonparametric}. This can be used to assure a
safer use of parametric tests like ANOVA or t-test. Another usual
prerequisite of parametric tests is the homoscedasticity, that is, the
equivalence of the variances of the populations. Levene's test is used
to this purpose, so that we can reject the hypothesis of the
equivalence of the variances at a level of significance, and then one
of the alternatives would be to use a non-parametric test
\cite{2010-Nordstokke-newnonparametricLevene,
  2009-Garcia-studystatisticaltechniques}. Although it is correct to
use non-parametric tests to achieve our main goal before even checking
these preconditions, it is worth doing it because parametric tests
make more assumptions about the population and when they are satisfied
they have stronger information and consequently in these circumstances
they are more powerful.

However, non-parametric tests also make some assumptions that are
easier to fulfil than parametric tests, like
symmetry in the Wilcoxon Signed-Ranks test. This means that a safe use
of this test is depending on the symmetry of the
population, \ie the probability density function is
  reflected around a vertical line at some value
\cite{2010-Kasuya-Wilcoxonsignedrankstest}.

\subsubsection{Pairwise comparisons}
\label{sec:pairwise-comparisons}

The first kind of experimental study would consist in the comparison
of a pair of algorithms' performance in several benchmarks. A pair of
non-parametric tests are presented in this section.

\begin{description}
\item[Sign-test] This simple test developed by Dixon and Mood
  \cite{1946-Dixon-statisticalsigntest} gives an idea of the way that
  non-parametric tests are developed. The assumption made here is that
  if the two algorithms involved in the comparison had an equivalent
  performance, the number of cases where each algorithm outperforms
  the other would be similar. 

\item[Wilcoxon signed-rank test] The underlying idea of this test
  \cite{1945-Wilcoxon-IndividualComparisonsRanking} is the next step
  in non-parametric tests, not just making a count of the wins of each
  algorithm but ranking the differences between the performance and
  developing the statistic over them
  \cite{1981-Conover-Ranktransformationsbridge}. This is already a
  widespread test used to compare two algorithms when t-test
  prerequisites are not fulfilled.
\end{description}

These tests receive the means of the runs for each benchmark function
of two algorithms. Unlike the parametric tests, in which null
hypothesis consists in the equivalence of the means, in these tests
$\mathcal{H}_{0}$ is the equivalence of the medians.

\subsubsection{Multiple comparisons}
\label{sec:multiple-comparisons}

Sometimes we are interested in the joint analysis of multiple
algorithms. We have the result of every algorithm for every problem,
but we are not allowed to make a pairwise comparison for each pair of
algorithms without losing the control on the
Family-Wise Error Rate (FWER), \ie augmenting the
probability of making at least one type I error. This effect is easily
computed. If we allow an error of $\alpha$ in each test (the
significance) and conduct $k$ tests, the probability of making at
least one error would be $1-(1-\alpha)^k$. Some methods are described
below so that they can be used instead of repeated pairwise
comparisons:

\begin{description}
\item[Multiple Sign test] This test is an extension of Sign test when
  we compare $k$ algorithms with a control algorithm. Differences for
  every pair $(algorithm,$ $problem)$ with the control algorithm are
  computed and the number of each sign is counted. Then, the median of
  the performance of every algorithm is compared with the median of
  the control algorithm. A directional null hypothesis is rejected if
  the number of minus (or plus) signs is less than or equal to the
  corresponding critical value available in the table for a treatment
  versus control multiple comparisons sign test
  \cite{1965-Rhyne-TablesTreatmentsControl} for specific values of $n$
  (the number of benchmarks), $m=k-1$ and level of significance
  $\alpha$.

\item[Friedman test] Friedman test is the non-parametric test that
  accomplishes the ANOVA test purpose. The null hypothesis in this
  test is the equivalence of the medians of the different algorithms
  and the alternative hypothesis consists in the difference of two or
  more algorithms medians. To calculate the statistic, we rank the
  algorithm performance for each problem and compute the average of
  each algorithm between problems. Iman and Davenport proposed a more
  powerful version of this test using a modified statistic
  \cite{1980-Iman-Approximationscriticalregion}.

\item[Friedman Aligned-Ranks test] The main inconvenient of the
  Friedman test is that ranks are just considered in each problem,
  which leads to a conservative test when comparing few
  algorithms. However, Friedman Aligned-Ranks test
  \cite{1962-Hodges-RankMethodsCombination} considers every
  algorithm-problem pair to produce a rank from 1 to $k \cdot n$.

\item[Quade test] The Quade test
  \cite{1967-Quade-Rankanalysiscovariance} takes into account the
  different difficulties of the problems. This is made calculating the
  range of the problems as the maximum differences between the samples
  and ranking the $n$ ranges. Then, these ranks are assigned to the
  problems and represent a weight to the ranks
    obtained by the Friedman method. With these weighted ranks, the
  procedure follows as usual.
\end{description}

\subsubsection{Comparison of Convergence Performance}
\label{sec:comp-conv-perf}

A fundamental aspect that should be considered in the comparison of
optimisation algorithms is the convergence of their results, as it is
a desirable property. Derrac \etal suggest the use of the Page trend
test for this purpose
\cite{2014-Derrac-Analyzingconvergenceperformance}. In the original
definition of the Page test, this test is designed to detect
increasing trends in the rankings of the original data. The idea used
for the comparison of the convergence is to apply the test in the
differences of the results between two algorithms $\mathbf{A}$ and
$\mathbf{B}$ at $c$ equidistant cut points of the search on the
corresponding $n$ benchmark functions. The null hypothesis is the
equivalence of the means for each cut point and this hypothesis is
rejected in favor of an ordered alternative, which states that there
is an increasing trend in the ranks of the differences between
$\mathbf{A}$ and $\mathbf{B}$ ($\mathbf{A}-\mathbf{B}$). Under these
preconditions and assuming that the objective is the minimisation of
the functions, the interpretation to be made is as follows:

\begin{enumerate}
\item If $\mathcal{H}_{0}$ is rejected, then an increasing trend is
  detected. This could mean that the optimum of $\mathbf{A}$ is
  growing faster than the result of the $\mathbf{B}$
  algorithm (which lacks of logic because when an algorithm finds a
  minimum, it does not grow with the iterations) or the fitness of
  $\mathbf{B}$ is decreasing faster than the fitness of $\mathbf{A}$,
  that is, the algorithm $\mathbf{B}$ converges
    faster than the algorithm $\mathbf{A}$.
\item If the test is made over $\mathbf{B}-\mathbf{A}$ and the null
  hypothesis is rejected, the reasoning is analogous to the previous
  one and the algorithm $\mathbf{A}$ converges
    faster.
\item If $\mathcal{H}_{0}$ is not rejected, we cannot say anything
  about the relative convergence of the two compared algorithms.
\end{enumerate}

As we are analyzing the differences in the trend
between the algorithms $\mathbf{A}$ and $\mathbf{B}$, a modification
in the rankings of the difference is needed when an algorithm reaches
the optimum before the end depending on the circumstance:
\begin{itemize}
\item If no algorithm reaches the optimum, there is not any further
  change.
\item If algorithm $\mathbf{A}$ reaches the optimum before the end,
  ranks should increase from the cut point where algorithm
  $\mathbf{A}$ reaches the optimum to the last cut point in a way that
  the highest ranks are in the last positions.
\item If algorithm $\mathbf{B}$ reaches the optimum before the end,
  ranks should decrease from the cut point where algorithm
  $\mathbf{B}$ reaches the optimum and include the lowest values.
\item If both algorithms reach the optimum at the same point, the
  ranks are computed as normal in the original version of the test.
\end{itemize}

It is important to note that this test focuses on the trends of the
differences, so the conclusions are submitted to the sign of the
difference between the firsts observations. This means that if the
algorithm $\mathbf{A}$ starts at a point near the objective and then
decrease its score function slowly and algorithm $\mathbf{B}$ starts
with a bad score, the test could reject the null hypothesis because
the algorithm $\mathbf{B}$ reduce its score at a greater pace. Then,
the comparison between the convergences should be made in the context
of algorithms with similar initial performance, where the convergence
is the key aspect of the performance.

Another relevant issue of the Page test is that we can only compare
two algorithms at a time. Then, if we are interested in the comparison
of the convergence (or final results) of multiple algorithms or the
fine-tuning of the parameters, we should incorporate the post-hoc
procedures to the statistical background described in this paper.

\subsection{Post-hoc Procedures}
\label{sec:post-hoc-procedures}

We should, however, be conscious that these
multiple-comparison tests, as in ANOVA, only can
detect significant differences between the whole group of algorithms,
but cannot tell where these differences are. Then, we declare a family
of hypotheses of the comparison between a control algorithm and the
$k-1$ remaining algorithms or among the $k$ algorithms, with $k-1$ or
$k(k-1)/2$ hypotheses, respectively. Post-hoc tests aim to obtain a
set of $p$-values for this family of hypotheses.

We compute the $p$-value of every hypothesis using a normal
approximation for a statistic that depends on the non-parametric test
used. However, as we previously stated, singular $p$-values should not
be used for multiple comparison due to the loss of the control over
the FWER, so an adjustment is made to obtain the adjusted $p$-values
(APVs). Several methods are available for the computation of the adjusted
$p$-values are for example Bonferroni-Dunn, Li, Holm, Holland,
Hochberg and Rom methods
\cite{2010-Garcia-Advancednonparametrictests}.

\subsubsection{One vs. $n$ algorithms} In the context of a comparison
of multiple algorithms versus one control algorithm, we can follow the
Bergmann and Hommel procedure, where we say that an index set of
hypotheses $I \subseteq \{1, \dots, m\}$ is exhaustive if all
associated hypothesis $H_j, j \in I$ could be true.  It rejects every
hypothesis $H_j$ as $j$ is not included in the acceptance set
$A = \cup\{I: I \ \mathrm{exhaustive}, \min\{ p_i: i \in I\} >
\alpha/|I|\}$, where $|\cdot|$ represents the cardinality of a
set. Due to the high cost of checking if each $I$ is exhaustive, a set
$E$ with all the exhaustive sets of hypotheses is previously
calculated and then the hypotheses that are not in $A$ are rejected.
For the Bergmann-Hommel test, $APV_i = \min\{\upsilon, 1\}$ where
$\upsilon = \max\{|I| \cdot \min\{p_j, j \in I\}: I \
\mathrm{exhaustive,} \ i \in I\}$.

\subsubsection{$n$ vs. $n$ algorithms}
Although post-hoc procedures provide an adjusted and reliable
$p$-value for the comparison between algorithms even when we are
making the comparison among all the algorithms, we must consider that
not all combinations of true and false in the hypotheses are
allowed. For example, if the hypothesis of the equivalence between
algorithms $i$ and $j$, say $H_{i,j}$ is rejected, at least one of the
hypotheses $H_{i,k}$ and $H_{j,k}$ must be also rejected.

When we are interested in carrying out all $m = \frac{k(k-1)}{2}$
pairwise comparisons, we can simply use the Holm's test (with a little
modification) or the Nemenyi procedure. These variants make an
adjustment on the signification level through the computed $p$-values:
Holm $p$-value is computed as $APV_i = \min\{\upsilon,1\},$ with
$\upsilon = \max\{(m-j+1)p_j: 1 \leq j \leq i\}$; and Nemenyi
$APV_i = \min\{\upsilon, 1\}$, where $\upsilon = m p_i$). The latter
procedure is equivalent to if $\alpha$ was divided by the number of
comparisons performed ($m$). There is an associated graphic with this
test proposed by Dem\v{s}ar, known as Critical Difference plots
\cite{2006-Demsar-Statisticalcomparisonsclassifiers}, where the
algorithms are ordered by their mean ranks and linked when the null
hypothesis of their equivalence is not rejected by the test. Then, in
a single plot, it can be seen what algorithms equivalence is discarded
by the post-hoc procedure and which groups are formed with a not
sufficiently different performance. This method is simple but is less
powerful than the other ones \cite{2011-Derrac-practicaltutorialuse}:

\begin{description}
\item[Shaffer's static procedure] This method is based on the Holm's
  test although it takes into account the possible hypotheses that can
  be simultaneously true. Obtained $p$-values are ordered and
  hypotheses $H_1, \dots, H_i$ are rejected if $p_i < \alpha / t_i$,
  with $t_i$ the maximum number of hypotheses that can be true when
  any of $H_1, \dots, H_i$ are false.  This method does not depend on
  the $p$-values but it is determined by the given hypotheses. $t_i$
  values are the maximum of the set
\[
  S(k) = \bigcup\limits_{j=1}^k
		\left\lbrace
			{j \choose 2} + x : x \in S(k-j)
		\right\rbrace  \ \mathrm{if}\ k>2, S(0)=S(1)=\{0\},
\]
with $S(k)$ the set of numbers of hypotheses which can be true when
$k$ algorithms are compared.

\item[Shaffer's dynamic procedure] This is a modification to
increase the power of the method. $\alpha=t_i$ is substituted by
$\alpha=t_i^*$, the maximum number of hypothesis that can be true with
the current assignment. That means that $t_i^*$ also depends on the
already rejected hypothesis.  Shaffer $APV_i = \min\{\upsilon, 1\}$
where $\upsilon = \max\{t_jp_j: 1 \leq j \leq i\}$.
\end{description}

\subsection{Confidence Intervals and Beyond}
\label{sec:conf-interv-beyond}

Confidence Intervals (CIs) are usually used in the context of
statistical analysis, presented as a counterpart for the $p$-value. As
classic NHST, CIs are defined at an arbitrary although fixed level of
confidence (usually $95\%$), but they represent a suitable range for
the location of the parameter of interest. The information given by
the effect size concept is then included in the width of the CI: a
narrower interval implies more certainty about the real difference. CI
also includes the null-hypothesis test information. In general, if a
$95\%$-CI does not include the value that represents the equivalence
of the performance, the NHST gives a $p$-value less than $0.05$ and
the hypothesis is rejected
\cite{2013-Berrar-Significancetestsconfidence}. On the other hand, if
the null value belongs to the $95\%$-CI, the equivalence of the
algorithms cannot be discarded.

Although in the Berrar's work
\cite{2013-Berrar-Significancetestsconfidence} the criticisms are
applicable for NHST for both classification and optimisation problems,
the proposal of the computation of the confidence intervals described
in this paper is not valid for optimisation problems because it
expects the data to come from a cross-validation experiment.

Next, we propose here a modification over the computation of the
confidence intervals from a non-parametric and ranking perspective,
adducing the reasons that justify the use of the Wilcoxon test but
without the drawbacks of NHST
\cite{1988-Campbell-StatisticsMedicineCalculating}.


If we have $l$ observations from two algorithms over one benchmark
function, $\{x_{1}, \dots, x_{l}\}, \{y_{1}, \dots, y_{l}\}$, then we
estimate the approximate $100(1-\alpha)\%$ confidence interval of the
difference of the two medians with the median of all the $l^{2}$
differences with the following process:
\begin{enumerate}
\item $K = W_{\alpha/2} - \frac{l(l+1)}{2}$, where $W_{\alpha/2}$ is
  the $100\alpha/2$ percentile of the Wilcoxon two-sample test
  statistic distribution (see pp 156-162 of Center
  \cite{1982-Seldrup-Geigyscientifictables}).
\item When sample size is greater than 20, $K$ can be approximated by
  $K = \frac{l^{2}}{2} - z_{1-\alpha/2}
  \sqrt{\frac{l^{2}(2l+1)}{12}}$, and then rounded to the next integer
  value, with $z_{1-\alpha/2}$ is the $1-\alpha/2$ percentile of the
  standard Gaussian distribution.
\item The confidence interval is defined from the $K$th smallest
  difference to the $K$th greatest of the $l^{2}$ differences of
  the sample.
\end{enumerate}
  
\subsubsection{Confidence Curves}
\label{sec:confidence-curves}

Berrar undertook an in-depth study on the use of confidence intervals
through the concept of the confidence curve
\cite{2017-Berrar-Confidencecurvesalternative}, which is a plot of the
confidence intervals for all levels for a point estimate, \ie the
observed difference of the samples. Thus, criticisms like the
arbitrariness of the percentage of the level of significance are
turned down.

In this confidence curve, the $x$-axis represents each null hypothesis
while the $y$-axis shows the associated $p$-value. With this
representation, for each level $\alpha$ in the $y$-axis, in the
$x$-axis we obtain a $100(1-\alpha)\%$ confidence
interval. Therefore, a wider curve means that there is less certainty
about the estimated difference between the algorithms.

An interesting advantage of the use of confidence curves in the
comparison of optimisation algorithms is the availability in a single
comparison of the information given by an NHST (if
the null hypothesis is rejected), the classic confidence interval (and
the effect size), and the different confidence intervals at
other significance levels. However, the information
provided by this method requires more attention than a single
$p$-value.



\section{Known Criticisms to Null Hypothesis Statistical Tests}
\label{sec:null-hypoth-stat}

The need of the use of statistical tests in the analyses and
experimentation is unquestioned. However, certain controversy exists
regarding the types of these tests and the implications of their
results, mainly from a Bayesian perspective
\cite{2012-Good-Commonerrorsstatistics}. The most repeated criticisms
about the NHST are:

\begin{itemize}
\item They do not provide the answer that we expected
  \cite{2010-Kruschke-Bayesiandataanalysis,
    2017-Benavoli-TimeChangeTutorial}. Commonly, when we are using
  statistical tests to compare optimisation algorithms, we want to
  prove that our algorithm outperforms existing algorithms. Then, once
  an NHST is carried out, a $p$-value is obtained and this is often
  misunderstood as the probability of the null hypothesis being
  true. As we previously mentioned, the $p$-value is the probability
  of getting a sample as extreme as the observed one, \ie the
  probability of getting that data if the null hypothesis is true,
  $P(D|\mathcal{H}_0)$, instead of $P(\mathcal{H}_0|D)$.
	
\item We can almost always reject $\mathcal{H}_0$ if we get enough
  data, making a little difference significant through a high number
  of experiments. This implies that a test could determine a
  statistically significant difference but without practical
  implications. On the other hand, a significant difference may not be
  detected when there is insufficient data. Usually when an NHST is
  made, all hypotheses with $p$-value $p \leq 0.05$ are rejected, and
  those with $p > 0.05$ are considered not significant. This issue
  could lead to the inclusion in the experimentation of a new proposal
  of many benchmark functions to cause the rejection of the
  null hypothesis although the differences between the algorithms were
  random and very small.
	
\item There is an important misconception related to the
  reproducibility of the experiments, as a lower $p$-value does not
  mean a higher probability of obtaining the same results if the
  experiments are replicated as is described by Berrar and Lozano
  \cite{2013-Berrar-Significancetestsconfidence}. This is because, if
  the null hypothesis is rejected, the probability of obtaining a
  significant $p$-value is determined by the power of the test, which
  depends on the $\alpha$ level, the true effect size in the target
  distribution and the size of the test set, but not on the $p$-value.

\item A crucial issue is that we have no information when
  an NHST does not reject the null hypothesis and
  not finding a significant difference does not mean that the
  algorithms performance is equivalent.
\end{itemize}

\section{Bayesian Paradigm and Distribution Estimation}
\label{sec:bayesian-tests}

In recent years, the Bayesian approach has been proposed as an
alternative of frequentist statistics for comparing algorithms
performance in optimisation and classification problems
\cite{2017-Benavoli-TimeChangeTutorial}. In the following we also
describe some differences with respect to the NHST and their problems
addressed in the previous section.

\subsection{Bayesian Parameter Estimation and Notation} 
\label{sec:bayes-param-estim}

In this Bayesian approach, a set of candidate values of a parameter,
which includes the possibility of no difference between samples, is
set up \cite{2003-Gelman-BayesianDataAnalysis}. Then we compute the
relative credibility of the values in the candidate set according to
the observations and using Bayesian inference. This approach is
preferable in optimisation comparison to Bayesian model comparison
approach, because this kind of comparison does not have a null value
that brings the NHST objections back to Bayesian analysis.

Bayesian analysis is executed in three steps:
\begin{description}
\item[First step] Establish a mathematical model of the data. In a
  parametric model it will be the likelihood function
  $P(Data | \theta)$.
\item[Second step] Establish the prior distribution $P(\theta)$. The
  common procedure is to select a prior distribution whose posterior
  distribution is known.
\item[Third step] Use Bayes' rule to obtain the posterior distribution
  $P(\theta | Data)$ from the combination of likelihood and prior
  distribution. This means that we can see the distribution of the
  difference of performance between the algorithms, which may reveal
  that their results differ but that there is no algorithm that
  outperforms the other. For instance, if one algorithm gets better
  results in one problem, but worse in another one, using Frequentist
  statistics we would not reject the hypothesis of
  equivalence, but we could not know if their results are similar in
  all the observations.
\end{description}

The tests described in this section are oriented to obtain the
posterior distribution of the difference of performance between two
algorithms, noted as $z$.  We set a prior Dirichlet Process (DP) with
base measure $\alpha$. This process is a probability measure on the
set of probability measures on a determined space (as we are
interested in the distribution of $z$, this space will be
$\mathbb{R}$). This means that if we make a sample from a DP, we do
not obtain a number, like we would do if we make a sample from a
uni-dimensional Gaussian distribution, but a probability measure on
$\mathbb{R}$. Moreover, as a property of the DP, if
$P \sim DP(\alpha)$, for any measurable partition of $\mathbb{R}$
$B = \{ B_{1}, \dots, B_{m}\}$, we obtain
$P(B_{1}), \dots, P(B_{m}) \sim \mathnormal{Dir}(\alpha(B_{1}), \dots,
\alpha(B_{m}))$.

This procedure could be thought as if we started
with a distribution of the probability of an algorithm outperforming
another algorithm. Then, using the observations we change this
distribution: so that while before it was a wider distribution with no
certainty regarding the results, now it is closer to the observations
made.

\subsection{Bayesian Sign and Signed-Rank Tests}
\label{sec:bayesian-sign-signed}

This test is presented by Benavoli \etal
\cite{2014-Benavoli-BayesianWilcoxonsignedrank} as a Bayesian version
of sign test. We consider a DP to be the prior distribution of the
scalar variable $z$. The DP is determined by the prior strength
parameter $s > 0$ and the prior mean $\alpha^* = \frac{\alpha}{s}$,
which is a probability measure for DPs. We usually use the measure
$\delta_{z_0}$, \ie a Dirac's delta centered on the
pseudo-observation $z_0$. Then the posterior probability density
function of the difference between algorithms $Z$ follows the
expression \cite{2017-Benavoli-TimeChangeTutorial}:
\[
  P(z) =
  	w_0 \delta_{z_0} (z) +
	  \sum_{j = 1}^n w_j \delta_{z_j}(z), \quad 
\]
where $(w_0, w_1, \dots, w_n) \sim Dir(s,1,\dots,1),$ that is a
combination of Dirac delta functions centered on the
observations with Dirichlet distributed weights. In the sign test, we
compute $\theta_l, \theta_e$ and $\theta_r$, the probabilities of the
mean difference between algorithms being in the intervals
$(-\infty, -r),$ $[-r,r]$ or $(r, +\infty)$ respectively, where $r$ is
the limit of the region of practical equivalence (rope) as a linear
combination of the observations with the weights $w_i$. Then,
\[
  \theta_l, \theta_e, \theta_r \sim Dir(n_l + sI_{(-\infty,-r)}(z_0),
  n_e + sI_{[-r,r]}(z_0), n_r + sI_{(r,+\infty)}(z_0)),
\]
where $n_l, n_e, n_r$ are the number of observations that fall in each
interval.

For a version of signed rank test the computation of the probabilities
$\theta_l, \theta_e, \theta_r$ is similar to that of the sign test,
although it involves pairs of observations in the modification of the
probabilities. This time there is not a simple distribution for the
probabilities, although we can estimate it using Monte Carlo sampling
$(w_0, w_1, \dots, w_n) \sim Dir(s,1, \dots, 1)$.

\subsection{Imprecise Dirichlet Process}
\label{sec:impr-dirichl-proc}

In the Bayesian paradigm we must fix the prior strength $s$ and the
prior measure $\alpha^*$ according to the available prior
information. If we do not have any information, we should specify a
non-informative Dirichlet Process. In our problem we could use the
limiting DP when $s \rightarrow 0$, but it brings out numerical
problems like instability in the inversion of Bayesian Friedman
covariance matrix. A solution then is assuming $s>0$ and
$\alpha^* = \delta_{X_1 = \dots = X_m}$, so we get
$\mathcal{E}[E[X_j-X_i]] = 0$ for each pair $i,j$ and
$\mathcal{E}[E[R_j]] = m(m-1)/2$. Then we are presuming that all the
algorithms' performances are equal, which is not non-informative. The
alternative proposed in
\cite{2014-Benavoli-BayesianWilcoxonsignedrank} is to use a prior
near-ignorance DP (IDP). This involves fixing $s>0$ and letting
$\alpha^*$ vary in the set of all probability measures, \ie
considering all the possible prior ranks. Posterior inference is then
computed taking $\alpha^*$ into account, obtaining lower and upper
bounds for the hypothesis probability.

\subsection{Bayesian Friedman Test}
\label{sec:bayes-friedm-test}

Bayesian sign test is generalised for the comparison
among $m \geq 3$ algorithms by the Bayesian Friedman test
\cite{2015-Benavoli-Bayesiannonparametricprocedure}. In this section
$\gamma$ represents the level of significance, as $\alpha$ denotes the
measure. We consider the function
$R(X_i) = \sum\limits_{i \neq k = 1} I_{\{X_i > X_k\}} + 1$, the
$i$-th rank. The main goal here is to test if the point
$\mu_0 = [(m+1)/2, \dots, (m+1)/2]$, \ie the point of null hypothesis
where all the algorithms have the same rank, is in
$(1-\gamma) \% SCR(E[R(X_1), \dots, R(X_m)])$, where the $SCR$ is the
symmetric credible region for $E[R(X_1), \dots, R(X_m)]$. If the
inclusion does not happen, there is a difference between algorithms
with probability $1-\gamma$.

For a large number of $n$ observations, we can suppose that the mean
and the covariance matrix tends to the sample mean $\mu$ and
covariance $\Sigma$. Then, we define
$\rho = \text{F}_{\text{inv}}(1-\gamma, m-1, n-m+1)
\frac{(n-1)(m-1)}{n-m+1}\}$, where $\text{F}_{\text{inv}}$ is the
inverse of the $\text{F}$ distribution. So we assume that
$\mu_0 \in (1-\gamma)\% SCR$ if
\[ (\mu-\mu_0)^T \Sigma^{-1} (\mu-\mu_0) |_{m-1} \leq \rho, \] where
the notation $|_{m-1}$ means that we take the first $m-1$ components.
For a small number of observations, we can compute the $SCR$ by
sampling from the posterior DP, resulting in the linear combination of
the weights and the Dirac delta functions centered in
the observations
$P = w_0 \delta_{\mathbf{X}_0} + \sum\limits_{j=1}^n
w_j\delta_{\mathbf{X}_j}$, where $\mathbf{X}_0$ is the
pseudo-observation.

\section{Multiple Measures Tests}
\label{sec:mult-meas-tests}

A relevant concern in the optimisation community is the comparison
between multi-objective algorithms. This issue can be addressed by
considering a weighted sum of the scores of the different objectives
or by calculating the Pareto frontier, selecting the algorithms that
are not worse than others in all the criteria or objectives. The
statistical approach for this scenario consists in the examination of
the relation between two studied elements (algorithms in our case)
through multiple observations (benchmarks) and measuring different
properties. As in the tests oriented to the comparison of
single-objective optimisation algorithms, there exist
  statistical tests from the different paradigms that address this
issue:

\subsection{Hotelling's $T^2$}

The parametric approach to the comparison of multivariate groups is
the Hotelling's $T^2$ statistics
\cite{2002-Willems-robustHotellingtest}. For this test, we would start
with two matrices, gathering the results of two algorithms in $m$
measures. This test assumes that the difference from the samples comes
from a multivariate Gaussian distribution
$\mathcal{N}_m(\mathbf{\mu}, \mathbf{\Sigma})$ and $\mathbf{\mu}$ and
$\mathbf{\Sigma}$ are unknown. This assumption can be checked with the
generalised version of Shapiro-Wilk's test
\cite{2009-VillasenorAlva-generalizationShapiroWilk}. We have not
found any proposal of the use of the Hotelling's $T^2$ test to the
comparison of algorithms, what can be motivated by the mentioned
assumption required. Then, the null hypothesis is that the mean of the
$m$ measures is the same for the two matrices. This means that if the
null hypothesis is rejected, there is at least a measure where the
algorithms obtain different values.

However, this test has some drawbacks, as the prerequisite of the
normality of the results. Besides, in multiple objective optimisation
field, the researchers are interested in finding Pareto optimal
solutions, and it is not enough knowing that there are differences
between the compared algorithms.

\subsection{Non-Parametric Multiple Measures Test}
\label{sec:non-param-mult-meas-test}

A relevant proposal to multi-objective algorithms comparison comes
from de Campos and
Benavoli~\cite{2016-deCampos-JointAnalysisMultiple}. In this study,
they propose a new test for the comparison of the results of two
algorithms in multiple problems and measures. They make a
non-parametric proposal for the comparison of two algorithms and its
Bayesian version.

In this comparison, we have two matrices with the results of two
algorithms in $n$ rows representing the problems or benchmark
functions measured in $m$ quality criteria or objectives. Then, let
$M_1,\dots,M_m$ be a set of $m$ performance measures, and, in the
comparison of the algorithms $A$ and $B$, we call a \textit{dominance
  statement} $D^{(B,A)} = [\succ, \prec, \dots, \prec]$, where the
comparison $\succ$ in the $i$-th entry of $D^{(B,A)}$ means that the
algorithm $B$ is better than $A$ for $M_{i}$. Then we want to decide
which (from $2^n$ possibilities) $D^{(B,A)}$ is the most appropriate
vector given the results matrices from each algorithm.

We denote $\mathbf{\theta} = [\theta_0,\dots,\theta_{2^{m-1}}]$ the set
of probabilities for each possible dominance statement and $i^{*}$ the
index of the most observed configuration. Then, the null hypothesis
$\mathcal{H}_{0}$ is defined as
$\theta_{i^{*}} \leq \max(\mathbf{\theta} \setminus \theta_{i^{*}})$,
\ie rejecting the null hypothesis would mean rejecting the fact that
the probability assigned to the most observed configuration is less or
equal to the second greatest probability. The computation of the
statistic is detailed in \cite{2016-deCampos-JointAnalysisMultiple}.

\subsection{Bayesian Multiple Measures Test}
\label{sec:bayes-mult-meas}

The Bayesian version of the Multiple Measures Test follows a Bayesian
estimation approach and estimates the posterior probability of the
vector of probabilities $\mathbf{\theta}$. A Dirichlet distribution is
considered to be the prior distribution. Then the weights are updated
according to the observations. The posterior probabilities of the
\textit{dominance statements} are computed by Monte Carlo sampling
from the space of $\theta$ and counting the fraction of times for each
$i$ that $\theta_i$ is the maximum of $\mathbf{\theta}$.

\section{Experimental Framework}
\label{sec:exper-fram}

In this section we describe the framework that will be used in the
experiments in \autoref{sec:experiments-results}. This way the use of
the previously defined test in the scenario of a statistical
comparison of the CEC'17 Special Session and Competition on Single
Objective Real Parameter Numerical Optimisation can be illustrated
\cite{2016-Awad-ProblemDefinitionsEvaluation}. The results have been
obtained from the organiser's GitHub repository
\footnote{\url{https://github.com/P-N-Suganthan/CEC2017-BoundContrained}}. The
mean final results for the LSHADE-cnEpSin algorithm, whose results
have not been correctly included in the organisation data, have been
extracted from the original paper.

\subsection{Benchmarks Functions}
\label{sec:benchmarks-functions}

The competition goal is finding the minimum of the
test functions $f(\mathbf{x})$, where
$\mathbf{x} \in \mathbb{R}^{D},\ D \in \{10,30,50,100\}$. All the
benchmark functions are shifted to a global optimum $\mathbf{o}$ and
scalable and rotated according to $M_{i}$ a rotation matrix. The
search range is $[-100,100]^{D}$ for all functions. Below we have
included a simple summary of the test functions:

\begin{itemize}
\item Unimodal functions:
  \begin{itemize}
  \item Bent Cigar Function
  \item Sum of Different Power Function. The results of this function
    have been discarded in the experiments because they presented some
    unstable behavior for the same algorithm presented in different languages.
  \item Zakharov Function
  \end{itemize}
\item Simple Multimodal Functions:
  \begin{itemize}
  \item Disembark Function
  \item Expanded Scaffer's F6 Function
  \item Lunacek Bi-Rastrigin Function
  \item Non-Continuous Rastrigin's Function
  \item Levy Function
  \item Schwefel's Function
  \end{itemize}
\item Ten Hybrid Functions formed as the sum of different basic functions.
\item Ten Composition Functions formed as a weighted sum of basic
  functions plus a bias according to which component optimum is the
  global one.
\end{itemize}

\subsection{Contestant Algorithms}
\label{sec:cont-algor}

The contestant algorithms are briefly described in the following list,
and are ordered according to the ranking obtained in the
competition. A short-name has been assigned to each algorithm in order
to present clearer tables and plots.
	
\begin{enumerate}
\item EBOwithCMAR (EBO) \cite{2017-Kumar-Improvinglocalsearcha}: Effective
  Butterfly Optimiser with a new phase which uses Covariace Matrix
  (CMAR).
\item jSO \cite{2017-Brest-Singleobjectiverealparameter}: Improved
  variant of iL-SHADE algorithm based on a new weighted version of
  mutation strategy.
\item LSHADE-cnEpSin (LSCNE) \cite{2017-Awad-Ensemblesinusoidaldifferential}:
  New version of LSHADE-EpSin that uses an ensemble of sinusoidal
  approaches based on the current adaptation and a modification of the
  crossover operator with a covariance matrix.
\item LSHADE-SPACMA (LSSPA) \cite{2017-Mohamed-LSHADEsemiparameteradaptation}:
  Hybrid version of a proposed algorithm LSHADE-SPA and CMA-ES.
\item DES \cite{2017-Jagodzinski-differentialevolutionstrategy}: An
  evolutionary algorithm that generates new individuals using a
  non-elitist truncation selection and an enriched differential
  mutation.
\item MM-OED (MM)\cite{2017-Sallam-Multimethodbasedorthogonal}:
  Multi-method based evolutionary algorithm with orthogonal experiment
  design (OED) and factor analysis to select the best strategies and
  crossover operators.
\item IDE-bestNsize (IDEN)
  \cite{2017-Bujok-Enhancedindividualdependentdifferential}: Variant
  of individual-dependent differential evolution with a new mutation
  strategy in the last phase.
\item RB-IPOP-CMA-ES (RBI) \cite{2017-Biedrzycki-versionIPOPCMAESalgorithm}:
  New version of IPOP-CMA-ES with a restart trigger according to the
  midpoint fitness.
\item MOS (MOS11, MOS12, MOS13)
  \cite{2017-LaTorre-comparisonthreelargescale}: Three large-scale
  global optimiser used in these scenarios.
\item PPSO \cite{2017-Tangherloni-ProactiveParticlesSwarm}: Self
  tuning Particle Swarm Optimisation relying on Fuzzy Logic.
\item DYYPO \cite{2017-Maharana-DynamicYinYangPair}: Version of
  Yin-Yang Pair Optimisation that converts a static archive updating
  interval into a dynamic one.
\item TLBO-FL (TFL) \cite{2017-Kommadath-TeachingLearningBased}: Variant to
  the Teaching Learning Based Optimisation algorithm that includes
  focused learning of students.
\end{enumerate}

\section{Experiments and Results}
\label{sec:experiments-results}

In this section we perform the previously described tests on the
competition results in order to provide clear examples of their use.

\paragraph{Setup considerations}
\begin{itemize}
\item Here we use the self-developed \textit{shiny} application
  \texttt{shinytests}\footnote{\url{https://github.com/JacintoCC/shinytests}},
  which makes use of our \texttt{R} package \texttt{rNPBST} for the
  analysis of the results of the competition. The associated blocks of
  code and scripts are available as a vignette in the \texttt{rNPBST}
  package\footnote{\url{https://jacintocc.github.io/rNPBST/articles/StatisticalAnalysis.html}}.
\item We will mainly use two data sets, one with all the results (all
  iterations of the execution of all the algorithms in all benchmark
  functions for all dimensions), \texttt{cec17.extended.final}, and
  the mean data set (which aggregates the results among the
  runs), \texttt{cec17.final}.
\item In most of the pairwise comparisons we have involved EBOwithCMAR
  and jSO algorithms, as they are the
  best-classified algorithms in the competition.
\item \autoref{tab:data-mean} shows the mean among different runs of
  the results obtained at the end of all of the steps of each
  algorithm on each benchmark function for the 10
  dimension scenario.
\item The tables included in this section are obtained with the
  function \texttt{AdjustFormatTable} of the package used, which is
  helpful to highlight the rejected hypotheses by emboldening the
  associated $p$-values.
\end{itemize}

\begin{sidewaystable}
\begin{adjustbox}{width=\textheight}
  \centering
\begin{tabular}{rSSSSSSSSSSSSSS}
  \hline 
 Benchmark & {DES} & {DYYPO} & {EBO} & {IDEN} & {jSO} & {LSSPA} & {MM} & {MOS11} & {MOS12} & {MOS13} & {PPSO} & {RBI} & {TFL} & {LSCNE} \\  \hline
  \hline
  1 & 0.00 & 2855.62 & 0.00 & 0.00 & 0.00 & 0.00 & 0.00 & 691.55 & 2891.97 & 3916.21 & 239.25 & 0.00 & 2022.58 & 0.00 \\ 
  3 & 0.00 & 0.00 & 0.00 & 0.00 & 0.00 & 0.00 & 0.00 & 0.00 & 0.00 & 0.00 & 0.00 & 0.00 & 0.00 & 0.00 \\ 
  4 & 0.00 & 2.07 & 0.00 & 0.00 & 0.00 & 0.00 & 0.00 & 0.00 & 0.00 & 0.00 & 1.20 & 0.00 & 3.03 & 0.00 \\ 
  5 & 1.54 & 11.20 & 0.00 & 3.28 & 1.76 & 1.76 & 1.11 & 6.94 & 64.75 & 16.42 & 18.08 & 1.58 & 8.75 & 1.69 \\ 
  6 & 0.12 & 0.00 & 0.00 & 0.00 & 0.00 & 0.00 & 0.00 & 0.00 & 50.74 & 0.00 & 0.23 & 0.00 & 0.00 & 0.00 \\ 
  7 & 11.93 & 21.80 & 10.55 & 12.89 & 11.79 & 10.93 & 11.52 & 18.96 & 48.51 & 27.73 & 16.91 & 10.11 & 27.64 & 11.98 \\ 
  8 & 1.56 & 13.25 & 0.00 & 2.89 & 1.95 & 0.84 & 1.11 & 6.97 & 65.03 & 14.22 & 9.95 & 1.97 & 12.28 & 1.80 \\ 
  9 & 0.00 & 0.02 & 0.00 & 0.00 & 0.00 & 0.00 & 0.00 & 0.00 & 2622.08 & 0.00 & 0.00 & 0.00 & 0.01 & 0.00 \\ 
  10 & 5.66 & 367.03 & 37.21 & 190.42 & 35.90 & 21.83 & 17.89 & 360.26 & 1335.45 & 610.39 & 503.09 & 435.43 & 954.62 & 43.03 \\ 
  11 & 0.12 & 9.28 & 0.00 & 0.00 & 0.00 & 0.00 & 0.00 & 2.54 & 32.53 & 7.18 & 16.89 & 0.17 & 4.12 & 0.00 \\ 
  12 & 440.84 & 13492.01 & 90.15 & 2.44 & 2.66 & 119.44 & 101.60 & 11049.04 & 15052.37 & 12283.25 & 4551.42 & 110.46 & 65562.15 & 101.28 \\ 
  13 & 3.31 & 5079.25 & 2.17 & 0.84 & 2.96 & 4.37 & 4.19 & 3293.93 & 11951.20 & 4367.38 & 1391.14 & 4.17 & 2447.82 & 3.66 \\ 
  14 & 12.25 & 20.67 & 0.06 & 0.00 & 0.06 & 0.16 & 0.09 & 158.28 & 7602.91 & 275.82 & 37.31 & 15.93 & 67.28 & 0.08 \\ 
  15 & 3.25 & 43.64 & 0.11 & 0.01 & 0.22 & 0.41 & 0.07 & 89.50 & 5916.64 & 408.54 & 53.27 & 0.49 & 125.90 & 0.32 \\ 
  16 & 6.09 & 43.84 & 0.42 & 0.49 & 0.57 & 0.74 & 0.25 & 34.47 & 496.95 & 13.51 & 82.98 & 97.10 & 8.91 & 0.54 \\ 
  17 & 21.04 & 14.09 & 0.15 & 0.79 & 0.50 & 0.16 & 0.06 & 2.89 & 210.46 & 22.22 & 24.59 & 52.46 & 38.30 & 0.31 \\ 
  18 & 29.13 & 8757.31 & 0.70 & 0.05 & 0.31 & 4.35 & 0.97 & 1401.92 & 9489.18 & 6393.55 & 878.26 & 19.72 & 6150.51 & 3.86 \\ 
  19 & 2.52 & 92.67 & 0.02 & 0.01 & 0.01 & 0.23 & 0.00 & 9.45 & 5658.60 & 1093.15 & 22.48 & 1.82 & 60.59 & 0.04 \\ 
  20 & 12.17 & 8.01 & 0.15 & 0.00 & 0.34 & 0.31 & 0.07 & 0.03 & 281.40 & 7.70 & 27.82 & 106.05 & 14.61 & 0.26 \\ 
  21 & 201.58 & 100.37 & 114.02 & 149.23 & 132.38 & 100.71 & 104.07 & 173.06 & 278.59 & 147.27 & 104.33 & 137.35 & 142.40 & 146.36 \\ 
  22 & 100.00 & 97.47 & 98.46 & 96.08 & 100.00 & 100.04 & 100.02 & 87.96 & 1260.22 & 99.57 & 96.73 & 99.26 & 93.31 & 100.01 \\ 
  23 & 301.46 & 309.44 & 300.17 & 301.91 & 301.21 & 302.68 & 298.42 & 311.92 & 368.51 & 318.16 & 342.15 & 275.02 & 306.87 & 302.00 \\ 
  24 & 303.04 & 116.85 & 166.21 & 292.97 & 296.60 & 274.60 & 103.93 & 276.24 & 345.06 & 301.33 & 226.63 & 197.58 & 310.38 & 315.83 \\ 
  25 & 407.65 & 422.99 & 412.35 & 413.91 & 405.96 & 428.37 & 414.05 & 415.02 & 433.67 & 430.51 & 404.10 & 402.27 & 425.52 & 425.56 \\ 
  26 & 296.08 & 303.38 & 265.40 & 300.00 & 300.00 & 300.00 & 294.12 & 254.78 & 1149.08 & 256.02 & 266.67 & 272.69 & 300.89 & 300.00 \\ 
  27 & 395.72 & 396.49 & 391.57 & 392.80 & 389.39 & 389.65 & 389.50 & 392.67 & 452.08 & 396.18 & 426.52 & 394.86 & 392.51 & 389.50 \\ 
  28 & 526.07 & 300.98 & 307.14 & 322.81 & 339.08 & 317.24 & 336.81 & 368.57 & 515.80 & 366.63 & 294.12 & 402.26 & 446.98 & 384.88 \\ 
  29 & 236.08 & 260.31 & 231.19 & 236.65 & 234.20 & 231.44 & 235.68 & 269.45 & 559.67 & 280.45 & 277.87 & 265.83 & 274.50 & 228.41 \\ 
  30 & 154002.28 & 6701.91 & 406.68 & 403.79 & 394.52 & 430.18 & 56938.19 & 21723.41 & 457211.16 & 60963.14 & 2993.25 & 2045.99 & 278998.94 & 17618.43 \\ 
\hline
\end{tabular}
\end{adjustbox}
\caption{Mean final results across runs in 10 dimension scenario}
\label{tab:data-mean}
\end{sidewaystable}

\subsection{Parametric Analysis}\label{parametric-analysis}

As we have described before, traditionally the statistical tests
applied to the comparison of different algorithms belonged to the
parametric family of tests. We start the statistical analysis of the
results with these kinds of tests and the study of the prerequisites in
order to use them safely.


The traditional parametric test used in the context of a comparison of
multiple algorithms over multiple problems (benchmarks) is the ANOVA
test, as we have seen in \autoref{sec:param-stat-tests}. This test
makes some assumptions that should be checked before it is performed:

\begin{enumerate}
\item The distribution of the results for each algorithm among
  different benchmarks follows a Gaussian distribution.
\item The standard deviation of results among groups is equal.
\end{enumerate}

In \autoref{tab:shapiro} we gather the $p$-values associated with the
normality of each group of mean results for an algorithm in a
dimension scenario. All the null hypotheses are rejected because the
$p$-values are less than $0.05$, which means that we reject that the
distribution of the mean results for each benchmark function follow a
normal distribution. This conclusion could be expected because of the
different difficulty of the benchmark functions in higher dimension
scenarios. This is marked with boldface in subsequent
tables. 


\begin{table}[ht]
\centering
\begin{tabular}{rllll}
  \hline
  Algorithm & Dim. 10 & Dim. 30 & Dim. 50 & Dim. 100 \\ \hline \\[-2ex]
DES & \(\mathbf{1.40 \cdot 10^{-11}}\) & \(\mathbf{6.37 \cdot 10^{-08}}\) & \(\mathbf{1.36 \cdot 10^{-11}}\) & \(\mathbf{3.85 \cdot 10^{-08}}\) \\ 
  DYYPO & \(\mathbf{7.62 \cdot 10^{-09}}\) & \(\mathbf{2.35 \cdot 10^{-11}}\) & \(\mathbf{3.67 \cdot 10^{-11}}\) & \(\mathbf{1.68 \cdot 10^{-11}}\) \\ 
  EBO & \(\mathbf{2.88 \cdot 10^{-06}}\) & \(\mathbf{1.22 \cdot 10^{-07}}\) & \(\mathbf{1.39 \cdot 10^{-11}}\) & \(\mathbf{2.71 \cdot 10^{-08}}\) \\ 
  IDEN & \(\mathbf{3.15 \cdot 10^{-06}}\) & \(\mathbf{4.24 \cdot 10^{-08}}\) & \(\mathbf{1.59 \cdot 10^{-11}}\) & \(\mathbf{1.21 \cdot 10^{-10}}\) \\ 
  jSO & \(\mathbf{1.28 \cdot 10^{-06}}\) & \(\mathbf{1.42 \cdot 10^{-07}}\) & \(\mathbf{1.40 \cdot 10^{-11}}\) & \(\mathbf{9.00 \cdot 10^{-09}}\) \\ 
  LSSPA & \(\mathbf{2.84 \cdot 10^{-06}}\) & \(\mathbf{3.76 \cdot 10^{-07}}\) & \(\mathbf{1.39 \cdot 10^{-11}}\) & \(\mathbf{3.71 \cdot 10^{-08}}\) \\ 
  MM & \(\mathbf{1.51 \cdot 10^{-11}}\) & \(\mathbf{1.68 \cdot 10^{-07}}\) & \(\mathbf{1.40 \cdot 10^{-11}}\) & \(\mathbf{7.61 \cdot 10^{-07}}\) \\ 
  MOS11 & \(\mathbf{3.38 \cdot 10^{-10}}\) & \(\mathbf{3.75 \cdot 10^{-10}}\) & \(\mathbf{1.51 \cdot 10^{-10}}\) & \(\mathbf{1.89 \cdot 10^{-09}}\) \\ 
  MOS12 & \(\mathbf{2.16 \cdot 10^{-11}}\) & \(\mathbf{2.42 \cdot 10^{-08}}\) & \(\mathbf{3.64 \cdot 10^{-11}}\) & \(\mathbf{2.79 \cdot 10^{-10}}\) \\ 
  MOS13 & \(\mathbf{1.09 \cdot 10^{-10}}\) & \(\mathbf{2.82 \cdot 10^{-09}}\) & \(\mathbf{3.03 \cdot 10^{-11}}\) & \(\mathbf{9.56 \cdot 10^{-11}}\) \\ 
  PPSO & \(\mathbf{7.57 \cdot 10^{-09}}\) & \(\mathbf{1.77 \cdot 10^{-10}}\) & \(\mathbf{3.32 \cdot 10^{-10}}\) & \(\mathbf{4.49 \cdot 10^{-11}}\) \\ 
  RBI & \(\mathbf{4.75 \cdot 10^{-09}}\) & \(\mathbf{2.12 \cdot 10^{-07}}\) & \(\mathbf{5.75 \cdot 10^{-11}}\) & \(\mathbf{2.07 \cdot 10^{-10}}\) \\ 
  TFL & \(\mathbf{4.47 \cdot 10^{-11}}\) & \(\mathbf{7.50 \cdot 10^{-11}}\) & \(\mathbf{2.41 \cdot 10^{-09}}\) & \(\mathbf{1.58 \cdot 10^{-11}}\) \\ 
  LSCNE & \(\mathbf{2.17 \cdot 10^{-11}}\) & \(\mathbf{2.20 \cdot 10^{-07}}\) & \(\mathbf{1.39 \cdot 10^{-11}}\) & \(\mathbf{1.95 \cdot 10^{-08}}\) \\ 
  \hline
\end{tabular}
\caption{$p$-values for Shapiro tests for the normality of the mean
  results}
\label{tab:shapiro}
\end{table}

In some circumstances like the Multi-Objective Optimisation we need to
include different measures in the comparison. We will now consider the
results of the different dimensions as if they were different measures
in the same benchmark function. Then, in order to perform the
Hotelling's $T^2$ test we first check the normality of the population
with the multivariate generalisation of
Shapiro-Wilk's test. \autoref{tab:mv-shapiro} shows that the normality
hypothesis is rejected for every algorithm. Therefore, we stop the
parametric analysis of the results here because the assumptions of
parametric tests are not satisfied.


\begin{table}[ht]
\centering
\begin{tabular}{lr}
  \hline
Algorithm & p-value \\ \hline \\[-2ex]
DES & \(\mathbf{5.08 \cdot 10^{-21}}\) \\ 
  DYYPO & \(\mathbf{7.67 \cdot 10^{-31}}\) \\ 
  EBO & \(\mathbf{3.57 \cdot 10^{-23}}\) \\ 
  IDEN & \(\mathbf{6.54 \cdot 10^{-26}}\) \\ 
  jSO & \(\mathbf{3.82 \cdot 10^{-24}}\) \\ 
  LSSPA & \(\mathbf{4.72 \cdot 10^{-23}}\) \\ 
  MM & \(\mathbf{2.64 \cdot 10^{-21}}\) \\ 
  MOS11 & \(\mathbf{4.48 \cdot 10^{-27}}\) \\ 
  MOS12 & \(\mathbf{1.84 \cdot 10^{-26}}\) \\ 
  MOS13 & \(\mathbf{7.50 \cdot 10^{-30}}\) \\ 
  PPSO & \(\mathbf{1.38 \cdot 10^{-29}}\) \\ 
  RBI & \(\mathbf{2.21 \cdot 10^{-24}}\) \\ 
  TFL & \(\mathbf{5.03 \cdot 10^{-29}}\) \\ 
  LSCNE & \(\mathbf{3.61 \cdot 10^{-23}}\) \\  
   \hline
\end{tabular}
\caption{$p$-values for Multivariate Shapiro tests}
\label{tab:mv-shapiro}
\end{table}

\subsection{Non-parametric Tests}\label{non-parametric-tests}

In this subsection we perform the most popular tests in the field of the
comparison of optimisation algorithms. We continue using the aggregated
results across the different runs at the end of the iterations, except
for the Page test for the study of the convergence.

\subsubsection{Classic tests}\label{classic-tests}

\paragraph{Pairwise comparisons}

First, we perform the non-parametric pairwise comparisons with the
Sign, Wilcoxon and Wilcoxon Rank-Sum tests described in
\autoref{sec:pairwise-comparisons} for the 10 dimension scenario. The
hypotheses of the equality of the medians is only rejected by the
Wilcoxon Rank-Sum test, and we can see in \autoref{tab:np-pairwise}
how for example Wilcoxon's statistics $R+$ and $R-$ are both high
numbers, which means that there is no significant difference between
the ranking of the observations where one algorithm outperforms the
other. Then, the next step requires all the different algorithms in
the competition to be involved in the comparison.




\begin{table}[ht]
  \centering
\begin{tabular}{@{}lllllll@{}}
  \toprule \\[-2ex]
  Test & \multicolumn{2}{l}{Binomial Sign} & \multicolumn{2}{l}{Wilcoxon} & \multicolumn{2}{l}{Wilcoxon SR} \\
  \midrule
  p-value & \multicolumn{2}{c}{0.211} & \multicolumn{2}{c}{0.693} & \multicolumn{2}{c}{\textbf{0.00035}}     \\
  \multirow{2}{*}{Statistics} & K & 8 & R+  & 151.5 & WRank & 627 \\
                              & K2 & 15 & R- & 283.5 & &               \\ \bottomrule
\end{tabular}
\caption{Non-parametric pairwise comparison between EBO and jSO for 10
  dimensional problems}
\label{tab:np-pairwise}
\end{table}

\paragraph{Multiple comparisons}

We can see in \autoref{tab:np-tests} how the tests that involve
multiple algorithms, described in \autoref{sec:multiple-comparisons}
reject the null hypotheses, that is, the equivalence of the medians of
the results of the different benchmarks.  We must keep in mind that a
comparison between thirteen algorithms is not the recommended
procedure if we want to compare our proposal. We should only include
the state-of-the-art algorithms in the comparison, because the
inclusion of an algorithm with lower performance could lead to the
rejection of the null hypothesis, not due to the differences between
our algorithm and the comparison group, but because of the differences
between this dummy algorithm and the others.



\begin{table}[ht]
\centering
\begin{tabular}{rllll}
  \hline
 Test & Dim. 10 & Dim. 30 & Dim. 50 & Dim. 100 \\ 
  \hline \\[-2ex]
Friedman & \(\mathbf{9.91 \cdot 10^{-11}}\) & \(\mathbf{1.12 \cdot 10^{-10}}\) & \(\mathbf{1.08 \cdot 10^{-10}}\) & \(\mathbf{1.33 \cdot 10^{-10}}\) \\ 
  Friedman AR & \(\mathbf{0.00 \cdot 10^{+00}}\) & \(\mathbf{0.00 \cdot 10^{+00}}\) & \(\mathbf{0.00 \cdot 10^{+00}}\) & \(\mathbf{0.00 \cdot 10^{+00}}\) \\ 
  Iman-Davenport & \(\mathbf{0.00 \cdot 10^{+00}}\) & \(\mathbf{0.00 \cdot 10^{+00}}\) & \(\mathbf{0.00 \cdot 10^{+00}}\) & \(\mathbf{0.00 \cdot 10^{+00}}\) \\ 
  Quade & \(\mathbf{7.91 \cdot 10^{-37}}\) & \(\mathbf{6.87 \cdot 10^{-62}}\) & \(\mathbf{1.27 \cdot 10^{-57}}\) & \(\mathbf{2.52 \cdot 10^{-61}}\) \\ 
   \hline
\end{tabular}
\caption[$p$-values for Non-Parametric tests for the results in the different
scenarios]{$p$-values for Non-Parametric tests for the results in the
  different scenarios\protect\footnotemark}
\label{tab:np-tests}
\end{table}

\footnotetext{In this and following tables, the precision used to
  round a number to zero is $2.220446 \cdot 10^{-16}$.}

\subsubsection{Post-hoc tests}\label{post-hoc-tests}

Then, we proceed to perform the post-hoc tests (described in
\autoref{sec:post-hoc-procedures}) in order to determine the location
of the differences between these algorithms. We use the modification
of the classic non-parametric tests to obtain the $p$-value associated
with each hypothesis, although we should adjust these $p$-values with
a post-hoc procedure.

\paragraph{Control algorithm}\label{control-algorithm}

To illustrate this, we first suppose that we are in a One versus all
scenario where we are presenting our algorithm (we will use
EBOwithCMAR, the winner of the CEC'17 competition). The possible
approach here, as in the rest of the analysis is:

\begin{itemize}
\item Considering all the results in the different dimensions as if
  they were different function or benchmarks, we would only obtain a
  single $p$-value for the comparison between EBO-CMAR with each
  contestant algorithm. The adjusted $p$-values are shown in
  \autoref{tab:np-tests-ph-control} for the Friedman, Friedman
  Aligned-Rank and Quade test. Here we see that there is not much
  difference between the different tests and that differences are
  found in the comparison with DYYPO, IDEN, MOS, PPSO, RBI and TFL,
  although the sign of these differences need to be checked in the raw
  data.
\end{itemize}



\begin{table}
\centering
\begin{tabular}{rSSS}
  \hline
 Algorithm & {Friedman} & {FriedmanAR} & {Quade} \\ 
  \hline
DES & \(\mathbf{4.26 \cdot 10^{-03}}\) & \(\mathbf{4.26 \cdot 10^{-03}}\) & \(8.93 \cdot 10^{-01}\) \\ 
  DYYPO & \(\mathbf{0.00 \cdot 10^{+00}}\) & \(\mathbf{0.00 \cdot 10^{+00}}\) & \(\mathbf{2.94 \cdot 10^{-05}}\) \\ 
  IDEN & \(\mathbf{3.61 \cdot 10^{-07}}\) & \(\mathbf{3.61 \cdot 10^{-07}}\) & \(3.71 \cdot 10^{-01}\) \\ 
  jSO & \(2.85 \cdot 10^{-01}\) & \(2.85 \cdot 10^{-01}\) & \(9.83 \cdot 10^{-01}\) \\ 
  LSSPA & \(2.85 \cdot 10^{-01}\) & \(2.85 \cdot 10^{-01}\) & \(9.52 \cdot 10^{-01}\) \\ 
  MM & \(5.05 \cdot 10^{-02}\) & \(5.05 \cdot 10^{-02}\) & \(8.93 \cdot 10^{-01}\) \\ 
  MOS11 & \(\mathbf{0.00 \cdot 10^{+00}}\) & \(\mathbf{0.00 \cdot 10^{+00}}\) & \(\mathbf{1.62 \cdot 10^{-04}}\) \\ 
  MOS12 & \(\mathbf{0.00 \cdot 10^{+00}}\) & \(\mathbf{0.00 \cdot 10^{+00}}\) & \(\mathbf{1.45 \cdot 10^{-06}}\) \\ 
  MOS13 & \(\mathbf{0.00 \cdot 10^{+00}}\) & \(\mathbf{0.00 \cdot 10^{+00}}\) & \(\mathbf{8.00 \cdot 10^{-05}}\) \\ 
  PPSO & \(\mathbf{0.00 \cdot 10^{+00}}\) & \(\mathbf{0.00 \cdot 10^{+00}}\) & \(\mathbf{1.62 \cdot 10^{-04}}\) \\ 
  RBI & \(\mathbf{2.97 \cdot 10^{-05}}\) & \(\mathbf{2.97 \cdot 10^{-05}}\) & \(3.97 \cdot 10^{-01}\) \\ 
  TFL & \(\mathbf{0.00 \cdot 10^{+00}}\) & \(\mathbf{0.00 \cdot 10^{+00}}\) & \(\mathbf{4.04 \cdot 10^{-06}}\) \\ 
  LSCNE & \(2.85 \cdot 10^{-01}\) & \(2.85 \cdot 10^{-01}\) & \(9.83 \cdot 10^{-01}\) \\ 
   \hline
\end{tabular}
\caption{Post-Hoc Non-Parametric tests with control algorithm}
\label{tab:np-tests-ph-control}
\end{table}

\begin{itemize}
\item If we wanted to show that the differences between the algorithms
  also persist in each group of results obtained across the different
  dimensions, we should perform these tests repeatedly and apply the
  appropiate post-hoc procedure later. In
  \autoref{tab:grp-friedman-holland} we show the adjusted $p$-values.
\end{itemize}


\begin{table}[H]
\centering
\begin{tabular}{rllll}
  \hline
 & Dim 10 & Dim 30 & Dim 50 & Dim 100 \\ \hline
DES & \(\mathbf{8.35 \cdot 10^{-04}}\) & \(1.24 \cdot 10^{-01}\) & \(9.95 \cdot 10^{-01}\) & \(9.85 \cdot 10^{-01}\) \\ 
  DYYPO & \(\mathbf{3.67 \cdot 10^{-07}}\) & \(\mathbf{2.15 \cdot 10^{-12}}\) & \(\mathbf{1.44 \cdot 10^{-09}}\) & \(\mathbf{3.82 \cdot 10^{-09}}\) \\ 
  IDEN & \(9.67 \cdot 10^{-01}\) & \(1.19 \cdot 10^{-01}\) & \(\mathbf{1.49 \cdot 10^{-02}}\) & \(\mathbf{1.80 \cdot 10^{-02}}\) \\ 
  jSO & \(9.86 \cdot 10^{-01}\) & \(9.96 \cdot 10^{-01}\) & \(9.96 \cdot 10^{-01}\) & \(9.96 \cdot 10^{-01}\) \\ 
  LSSPA & \(6.35 \cdot 10^{-01}\) & \(8.96 \cdot 10^{-01}\) & \(9.96 \cdot 10^{-01}\) & \(9.96 \cdot 10^{-01}\) \\ 
  MM & \(9.96 \cdot 10^{-01}\) & \(8.61 \cdot 10^{-01}\) & \(6.35 \cdot 10^{-01}\) & \(9.96 \cdot 10^{-01}\) \\ 
  MOS11 & \(\mathbf{1.79 \cdot 10^{-04}}\) & \(\mathbf{4.94 \cdot 10^{-09}}\) & \(\mathbf{3.49 \cdot 10^{-08}}\) & \(\mathbf{2.70 \cdot 10^{-05}}\) \\ 
  MOS12 & \(\mathbf{0.00 \cdot 10^{+00}}\) & \(\mathbf{0.00 \cdot 10^{+00}}\) & \(\mathbf{4.44 \cdot 10^{-13}}\) & \(\mathbf{2.10 \cdot 10^{-09}}\) \\ 
  MOS13 & \(\mathbf{3.82 \cdot 10^{-09}}\) & \(\mathbf{6.17 \cdot 10^{-12}}\) & \(\mathbf{4.47 \cdot 10^{-11}}\) & \(\mathbf{1.80 \cdot 10^{-06}}\) \\ 
  PPSO & \(\mathbf{2.19 \cdot 10^{-05}}\) & \(\mathbf{4.53 \cdot 10^{-10}}\) & \(\mathbf{3.64 \cdot 10^{-11}}\) & \(\mathbf{1.47 \cdot 10^{-08}}\) \\ 
  RBI & \(\mathbf{3.74 \cdot 10^{-02}}\) & \(6.65 \cdot 10^{-02}\) & \(5.98 \cdot 10^{-01}\) & \(9.92 \cdot 10^{-01}\) \\ 
  TFL & \(\mathbf{1.44 \cdot 10^{-09}}\) & \(\mathbf{6.80 \cdot 10^{-12}}\) & \(\mathbf{2.96 \cdot 10^{-11}}\) & \(\mathbf{1.51 \cdot 10^{-11}}\) \\ 
  LSCNE & \(3.85 \cdot 10^{-01}\) & \(9.95 \cdot 10^{-01}\) & \(9.96 \cdot 10^{-01}\) & \(9.96 \cdot 10^{-01}\) \\ 
   \hline
\end{tabular}
\caption{Results grouped by dimension. Friedman test + Holland adjust
  using EBO-CMAR as control algorithm.}
\label{tab:grp-friedman-holland}
\end{table}

\paragraph{$n$ versus $n$ scenario}\label{n-versus-n-scenario}

In the scenario of the statistical analysis of the results obtained
during a competition, we do not focus on the comparison of the results
of a single algorithm, rather we would make all the posible pairs, and
therfore we would not use the control algorithm.


The results of the $n \times n$ comparison using a Friedman test and a
Post-Hoc Holland adjust of the $p$-values is shown in
Tables~\ref{tab:nvsn-dim10-friedman-holland} and
  \ref{tab:nvsn-dim10-friedman-holland-2}. In these tables we can see
that there is not a single algorithm whose equivalence with the rest
of the algorithms is discarded (in the 10 dimension
scenario). However, for a multiple comparison with a high number of
algorithms, like in the competition used as example, the adjustment
makes finding differences between the algorithms more difficult. If we
observe the results of the best classified algorithms in the
competition, like jSO and EBO, we see that there are significant
differences with algorithms like MOS, PPSO or RBI but this difference
is not significant for LSHADE variants or MM.

\begin{table}[H]
\begin{adjustbox}{width=\textwidth}
  \centering
\begin{tabular}{rSSSSSSS}
  \hline
       & {DES}                            & {DYYPO}                          & {EBO}                            & {IDEN}                           & {jSO}                            & {LSSPA}                          & {MM}                             \\ \hline
DES   &  & \(9.99 \cdot 10^{-01}\) & \(\mathbf{3.69 \cdot 10^{-03}}\) & \(2.94 \cdot 10^{-01}\) & \(1.77 \cdot 10^{-01}\) & \(8.44 \cdot 10^{-01}\) & \(5.48 \cdot 10^{-02}\) \\
DYYPO & \(9.99 \cdot 10^{-01}\) &  & \(\mathbf{1.78 \cdot 10^{-06}}\) & \(\mathbf{1.20 \cdot 10^{-03}}\) & \(\mathbf{5.12 \cdot 10^{-04}}\) & \(\mathbf{1.82 \cdot 10^{-02}}\) & \(\mathbf{7.67 \cdot 10^{-05}}\) \\
EBO   & \(\mathbf{3.69 \cdot 10^{-03}}\) & \(\mathbf{1.78 \cdot 10^{-06}}\) &  & \(1.00 \cdot 10^{+00}\) & \(1.00 \cdot 10^{+00}\) & \(9.70 \cdot 10^{-01}\) & \(1.00 \cdot 10^{+00}\) \\
IDEN  & \(2.94 \cdot 10^{-01}\) & \(\mathbf{1.20 \cdot 10^{-03}}\) & \(1.00 \cdot 10^{+00}\) &  & \(1.00 \cdot 10^{+00}\) & \(1.00 \cdot 10^{+00}\) & \(1.00 \cdot 10^{+00}\) \\
jSO   & \(1.77 \cdot 10^{-01}\) & \(\mathbf{5.12 \cdot 10^{-04}}\) & \(1.00 \cdot 10^{+00}\) & \(1.00 \cdot 10^{+00}\) &  & \(1.00 \cdot 10^{+00}\) & \(1.00 \cdot 10^{+00}\) \\
LSSPA & \(8.44 \cdot 10^{-01}\) & \(\mathbf{1.82 \cdot 10^{-02}}\) & \(9.70 \cdot 10^{-01}\) & \(1.00 \cdot 10^{+00}\) & \(1.00 \cdot 10^{+00}\) &  & \(1.00 \cdot 10^{+00}\) \\
MM    & \(5.48 \cdot 10^{-02}\) & \(\mathbf{7.67 \cdot 10^{-05}}\) & \(1.00 \cdot 10^{+00}\) & \(1.00 \cdot 10^{+00}\) & \(1.00 \cdot 10^{+00}\) & \(1.00 \cdot 10^{+00}\) &  \\
MOS11 & \(1.00 \cdot 10^{+00}\) & \(1.00 \cdot 10^{+00}\) & \(\mathbf{8.41 \cdot 10^{-04}}\) & \(1.12 \cdot 10^{-01}\) & \(6.16 \cdot 10^{-02}\) & \(5.73 \cdot 10^{-01}\) & \(\mathbf{1.53 \cdot 10^{-02}}\) \\
MOS12 & \(\mathbf{7.67 \cdot 10^{-05}}\) & \(5.48 \cdot 10^{-02}\) & \(\mathbf{0.00 \cdot 10^{+00}}\) & \(\mathbf{3.91 \cdot 10^{-13}}\) & \(\mathbf{7.90 \cdot 10^{-14}}\) & \(\mathbf{6.43 \cdot 10^{-11}}\) & \(\mathbf{0.00 \cdot 10^{+00}}\) \\
MOS13 & \(8.23 \cdot 10^{-01}\) & \(1.00 \cdot 10^{+00}\) & \(\mathbf{1.69 \cdot 10^{-08}}\) & \(\mathbf{3.12 \cdot 10^{-05}}\) & \(\mathbf{1.14 \cdot 10^{-05}}\) & \(\mathbf{7.92 \cdot 10^{-04}}\) & \(\mathbf{1.24 \cdot 10^{-06}}\) \\
PPSO  & \(1.00 \cdot 10^{+00}\) & \(1.00 \cdot 10^{+00}\) & \(\mathbf{1.03 \cdot 10^{-04}}\) & \(\mathbf{2.59 \cdot 10^{-02}}\) & \(\mathbf{1.28 \cdot 10^{-02}}\) & \(2.10 \cdot 10^{-01}\) & \(\mathbf{2.55 \cdot 10^{-03}}\) \\
RBI   & \(1.00 \cdot 10^{+00}\) & \(6.05 \cdot 10^{-01}\) & \(1.41 \cdot 10^{-01}\) & \(9.75 \cdot 10^{-01}\) & \(9.22 \cdot 10^{-01}\) & \(1.00 \cdot 10^{+00}\) & \(6.62 \cdot 10^{-01}\) \\
TFL   & \(6.85 \cdot 10^{-01}\) & \(1.00 \cdot 10^{+00}\) & \(\mathbf{5.99 \cdot 10^{-09}}\) & \(\mathbf{1.34 \cdot 10^{-05}}\) & \(\mathbf{4.78 \cdot 10^{-06}}\) & \(\mathbf{3.88 \cdot 10^{-04}}\) & \(\mathbf{4.90 \cdot 10^{-07}}\) \\
LSCNE & \(9.75 \cdot 10^{-01}\) & \(6.16 \cdot 10^{-02}\) & \(8.28 \cdot 10^{-01}\) & \(1.00 \cdot 10^{+00}\) & \(1.00 \cdot 10^{+00}\) & \(1.00 \cdot 10^{+00}\) & \(9.98 \cdot 10^{-01}\) \\
  \hline
\end{tabular}
\end{adjustbox}
\caption{Results $n$ vs $n$, dimension 10. Friedman test + Holland
  adjust.}
\label{tab:nvsn-dim10-friedman-holland}
\end{table}

\begin{table}[H]
\begin{adjustbox}{width=\textwidth}
  \centering
\begin{tabular}{rSSSSSSS}
  \hline
       & {MOS11}                          & {MOS12}                          & {MOS13}                          & {PPSO}                           & {RBI}                            & {TFL}                            & {LSCNE} \\                          
  \hline
DES   & \(1.00 \cdot 10^{+00}\) & \(\mathbf{7.67 \cdot 10^{-05}}\) & \(8.23 \cdot 10^{-01}\) & \(1.00 \cdot 10^{+00}\) & \(1.00 \cdot 10^{+00}\) & \(6.85 \cdot 10^{-01}\) & \(9.75 \cdot 10^{-01}\) \\
DYYPO & \(1.00 \cdot 10^{+00}\) & \(5.48 \cdot 10^{-02}\) & \(1.00 \cdot 10^{+00}\) & \(1.00 \cdot 10^{+00}\) & \(6.05 \cdot 10^{-01}\) & \(1.00 \cdot 10^{+00}\) & \(6.16 \cdot 10^{-02}\) \\
EBO   & \(\mathbf{8.41 \cdot 10^{-04}}\) & \(\mathbf{0.00 \cdot 10^{+00}}\) & \(\mathbf{1.69 \cdot 10^{-08}}\) & \(\mathbf{1.03 \cdot 10^{-04}}\) & \(1.41 \cdot 10^{-01}\) & \(\mathbf{5.99 \cdot 10^{-09}}\) & \(8.28 \cdot 10^{-01}\) \\
IDEN  & \(1.12 \cdot 10^{-01}\) & \(\mathbf{3.91 \cdot 10^{-13}}\) & \(\mathbf{3.12 \cdot 10^{-05}}\) & \(\mathbf{2.59 \cdot 10^{-02}}\) & \(9.75 \cdot 10^{-01}\) & \(\mathbf{1.34 \cdot 10^{-05}}\) & \(1.00 \cdot 10^{+00}\) \\
jSO   & \(6.16 \cdot 10^{-02}\) & \(\mathbf{7.90 \cdot 10^{-14}}\) & \(\mathbf{1.14 \cdot 10^{-05}}\) & \(\mathbf{1.28 \cdot 10^{-02}}\) & \(9.22 \cdot 10^{-01}\) & \(\mathbf{4.78 \cdot 10^{-06}}\) & \(1.00 \cdot 10^{+00}\) \\
LSSPA & \(5.73 \cdot 10^{-01}\) & \(\mathbf{6.43 \cdot 10^{-11}}\) & \(\mathbf{7.92 \cdot 10^{-04}}\) & \(2.10 \cdot 10^{-01}\) & \(1.00 \cdot 10^{+00}\) & \(\mathbf{3.88 \cdot 10^{-04}}\) & \(1.00 \cdot 10^{+00}\) \\
MM    & \(\mathbf{1.53 \cdot 10^{-02}}\) & \(\mathbf{0.00 \cdot 10^{+00}}\) & \(\mathbf{1.24 \cdot 10^{-06}}\) & \(\mathbf{2.55 \cdot 10^{-03}}\) & \(6.62 \cdot 10^{-01}\) & \(\mathbf{4.90 \cdot 10^{-07}}\) & \(9.98 \cdot 10^{-01}\) \\
MOS11 &  & \(\mathbf{4.14 \cdot 10^{-04}}\) & \(9.70 \cdot 10^{-01}\) & \(1.00 \cdot 10^{+00}\) & \(1.00 \cdot 10^{+00}\) & \(9.20 \cdot 10^{-01}\) & \(8.44 \cdot 10^{-01}\) \\
MOS12 & \(\mathbf{4.14 \cdot 10^{-04}}\) &  & \(4.42 \cdot 10^{-01}\) & \(\mathbf{2.89 \cdot 10^{-03}}\) & \(\mathbf{2.78 \cdot 10^{-07}}\) & \(5.81 \cdot 10^{-01}\) & \(\mathbf{7.71 \cdot 10^{-10}}\) \\
MOS13 & \(9.70 \cdot 10^{-01}\) & \(4.42 \cdot 10^{-01}\) &  & \(9.99 \cdot 10^{-01}\) & \(9.71 \cdot 10^{-02}\) & \(1.00 \cdot 10^{+00}\) & \(\mathbf{3.50 \cdot 10^{-03}}\) \\
PPSO  & \(1.00 \cdot 10^{+00}\) & \(\mathbf{2.89 \cdot 10^{-03}}\) & \(9.99 \cdot 10^{-01}\) &  & \(9.88 \cdot 10^{-01}\) & \(9.97 \cdot 10^{-01}\) & \(4.82 \cdot 10^{-01}\) \\
RBI   & \(1.00 \cdot 10^{+00}\) & \(\mathbf{2.78 \cdot 10^{-07}}\) & \(9.71 \cdot 10^{-02}\) & \(9.88 \cdot 10^{-01}\) &  & \(5.92 \cdot 10^{-02}\) & \(1.00 \cdot 10^{+00}\) \\
TFL   & \(9.20 \cdot 10^{-01}\) & \(5.81 \cdot 10^{-01}\) & \(1.00 \cdot 10^{+00}\) & \(9.97 \cdot 10^{-01}\) & \(5.92 \cdot 10^{-02}\) &  & \(\mathbf{1.82 \cdot 10^{-03}}\) \\
LSCNE & \(8.44 \cdot 10^{-01}\) & \(\mathbf{7.71 \cdot 10^{-10}}\) & \(\mathbf{3.50 \cdot 10^{-03}}\) & \(4.82 \cdot 10^{-01}\) & \(1.00 \cdot 10^{+00}\) & \(\mathbf{1.82 \cdot 10^{-03}}\) &  \\ 
  \hline
  \end{tabular}
\end{adjustbox}
\caption{Results $n$ vs $n$, dimension 10. Friedman test + Holland
  adjust (cont.).}
\label{tab:nvsn-dim10-friedman-holland-2}
\end{table}

\begin{figure}[h]
    \centering
  \includegraphics[trim={1.5cm 2.5cm 1cm 2cm}, clip, width=.6\textwidth]{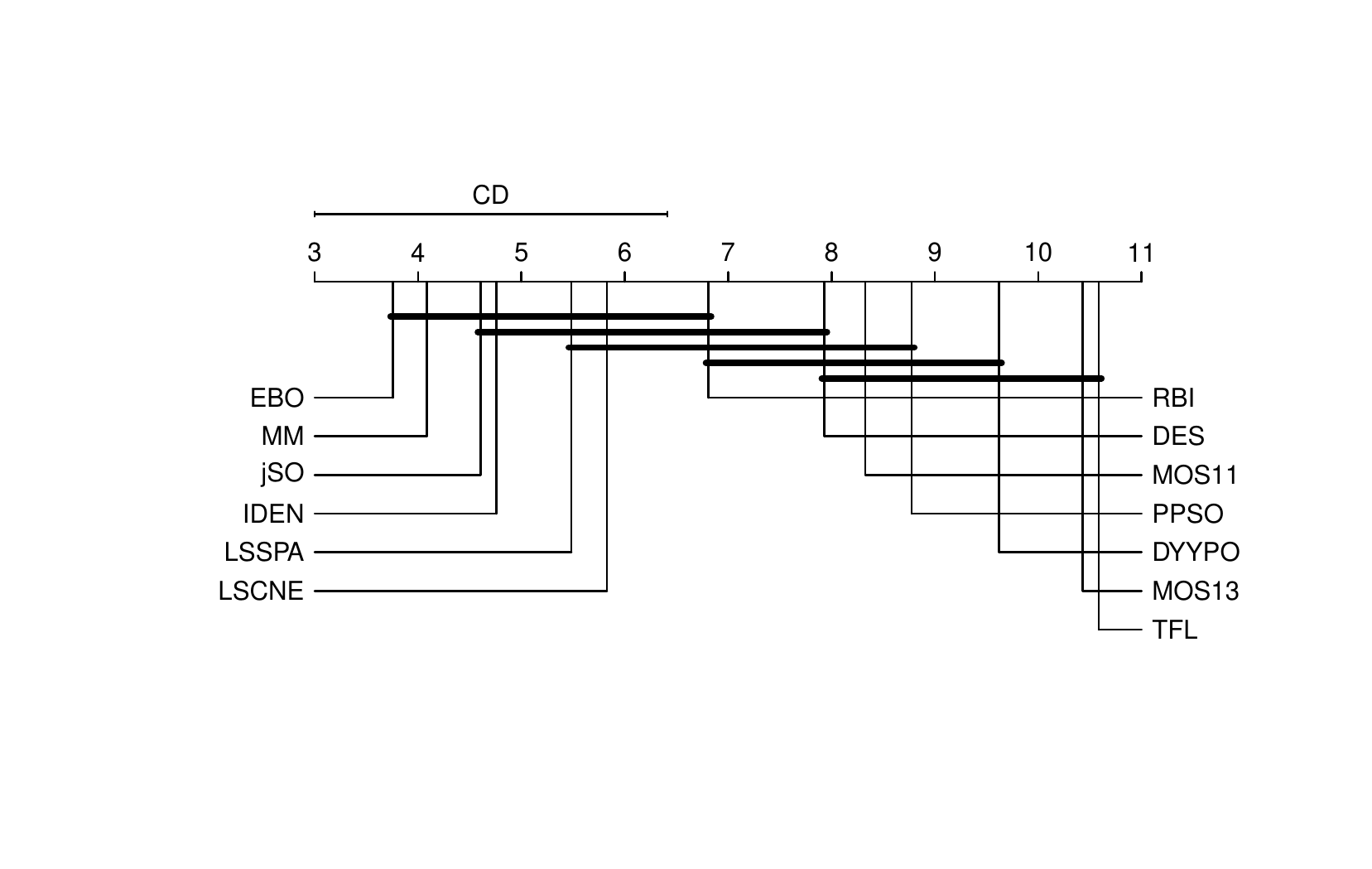}
  \caption{CD plot for 10 dimensional CEC 2017 functions}
\label{fig:cd-example}
\end{figure}

The CD plot associated with the scenario of an $n$ vs. $n$ comparison
is described in \autoref{sec:post-hoc-procedures} and performed with
the Nemenyi test provides an interesting visualisation fo the
significance of the observed paired differences. In
\autoref{fig:cd-example}, we show the results of this comparison,
where the differences between the group of the first classified
algorithms whose equivalence cannot be discarded includes up to the
RBI algorithm (7th classified in the 10 Dimension scenario). This
plot, with several overlapped groups that contain many algorithms,
shows that the differences are hard to identify in algorithms that
perform similarly. 



\subsubsection{Convergence test}
\label{convergence-test}

For the comparison of the convergence of two algorithms, we use the
Page test described in \autoref{sec:comp-conv-perf} with the mean
result across the different runs for each benchmark function of two
algorithms. These results could be equally extended using the
convenient adjustments. It is relevant to note that the LSCNE
algorithm (third classified) does not provide its partial results.

The results in \autoref{tab:page-1} show the comparison of the
convergence of the jSO and the LSSPA algorithms in the competition for
the 10 and the comparison of the convergence of jSO and DYYPO
algorithms for 100 dimension scenario. Here, the null hypothesis of
the difference between LSSPA and jSO getting a positive trend cannot
be rejected in the 10 dimension scenario. In the 100 dimension
scenario, the test detects an increasing trend in the ranks of the
difference between jSO and DYYPO as the null hypothesis is rejected.



\begin{table}[] 
\centering
\begin{tabular}{llll} 
  \hline
  \multicolumn{4}{c}{Page test} \\ \hline
  \multicolumn{4}{c}{10 dimensions scenario} \\ \hline
  \multirow{1}{*}{Comparison} & &  LSSPA-jSO & jSO-LSSPA \\
  \multirow{1}{*}{statistic} & L & 22902  & 22773 \\ 
  \multirow{1}{*}{p.value} &  & 0.4253 & 0.5759  \\ \hline
  \multicolumn{4}{c}{100 dimensions scenario} \\ \hline
  \multirow{1}{*}{Comparison} &  & jSO-DYYPO (100) & DYYPO-jSO (100) \\ 
  \multirow{1}{*}{statistic} & L & 25730 & 19945 \\
  \multirow{1}{*}{p.value} &  & \textbf{0} & 1  \\ \hline
 \end{tabular}
\caption{Page test}
\label{tab:page-1}
\end{table}

\subsubsection{Confidence Intervals and Confidence Curves}
\label{confidence-intervals-and-confidence-curves}

In this subsection, we show the use of confidence
intervals and confidence curves in the comparison of optimisation
results as mentioned in \autoref{sec:conf-interv-beyond}. First, we
must advise that these comparisons only take care of two algorithms at
a time, and a post-hoc correction would be needed if the comparison
involved a greater number of algorithms, as we will see in the
following examples.

We perform the comparison of the final results of PPSO and jSO
algorithms for the 10 dimension scenario. Results are shown in
\autoref{tab:wilcoxon-np-ci}.

\begin{figure}
  \centering
  \includegraphics[width=.8\textwidth]{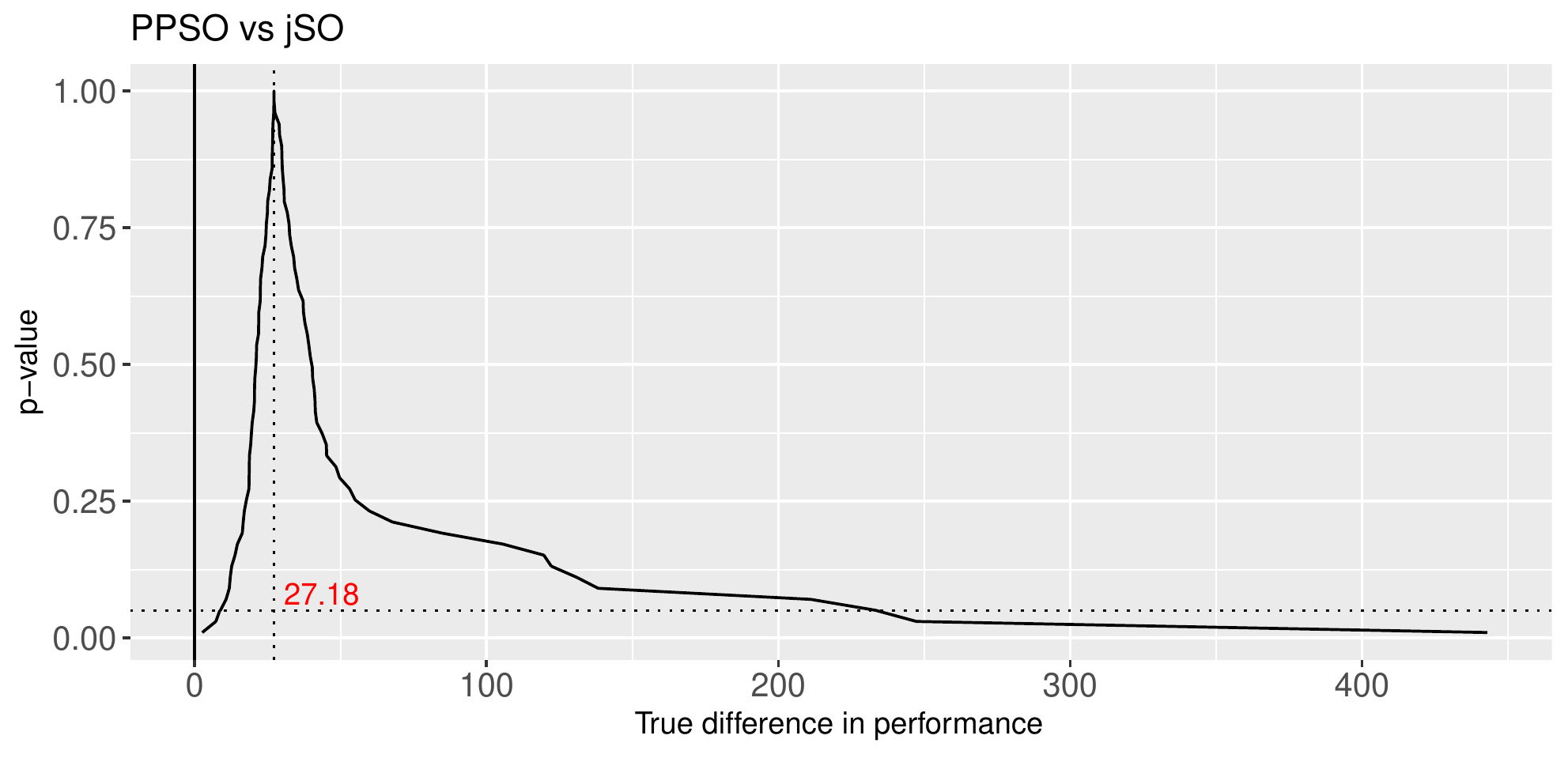}
  \caption{Confidence curve}
  \label{fig:conf-curve}
\end{figure}

As the 0 effect is not included in the non-parametric confidence
interval, the null hypothesis is then rejected. The interval is very
wide, so we have not much certainty the true location of the
parameter. If we only had done the Wilcoxon test, we would have
obtained the associated $p$-value, and the null hypothesis would also
be rejected, so the difference between the medians are detected with
both methods. If we look at the confidence curve, we can reject the
classic null hypothesis if the interval bounded by the intersections
of the horizontal line at the $\alpha$ value and the curve does not
contain 0. The confidence curve associated with the previous test is
plotted in \autoref{fig:conf-curve}, where we check that the null
hypothesis can be rejected. The estimated difference is represented
with the dotted vertical line. The width of the intersection between
the curve and the dotted horizontal line indicates that there is not
much certainty about the true location of the parameter,
primarily in the upper bound.



\begin{table}[] 
\centering 
\begin{tabular}{lll} 
\hline
\multicolumn{3}{c}{Wilcoxon test} \\ \hline

  \multirow{1}{*}{Comparison}
  & 		 & 		 \\ \hline
  \multirow{2}{*}{statistic}
  & 	R+	 & 	351.50	\\
  & 	R-	 & 	 83.50	 \\ \hline
  \multirow{2}{*}{p-value}
  & 	Exact Double pvalue	 & 	\textbf{0.0013990}	\\
  & 	Asymptotic Double Tail	 & 	\textbf{0.0000994}	 \\ \hline
  \multirow{2}{*}{Associated NP Confidence Interval}
  & 	Lower Bound	 & 	8.557851	\\
  & 	Upper Bound	 & 	233.709039	 \\ \hline
\end{tabular}
\caption{Wilcoxon test and Non-Parametric Confidence Interval}
\label{tab:wilcoxon-np-ci}
\end{table}

\subsection{Bayesian Tests}
\label{bayesian-tests}

In this subsection we illustrate the use of the described Bayesian tests.
The considerations are analogous to the ones made in the frequentist
case, as the described tests use the aggregations of the runs to compare
the results of the different benchmark functions, or use these runs with
the drawback of obtaining a restrained statement about the results in
one single problem.

\subsubsection{Bayesian Friedman test}
\label{tutorial-bayesian-friedman-test}

We start with the Bayesian version of the Friedman test, mentioned in
\autoref{sec:bayes-friedm-test}. In this test we do not obtain a
single $p$-value, but the accepted hypothesis. Due to the high number
of contestant algorithms and the memory needed to allocate the
covariance matrix of all the possible permutations, here we will
perform the imprecise version of the test. The null hypothesis of the
equivalence of the mean ranks for the 10 dimension scenario is
rejected. The mean ranks of the algorithms are shown in
\autoref{tab:friedman-mean-rank}.


\begin{table}[ht]
\centering
\begin{tabular}{rr}
  \hline
  Algorithm & Mean Rank \\ 
  \hline
  DES &  7.98 \\ 
  DYYPO & 9.62 \\ 
  EBO & 3.55 \\ 
  IDEN & 4.9 \\ 
  jSO & 4.7 \\ 
  LSSPA & 5.6 \\ 
  MM & 4.28 \\ 
  MOS-11 & 8.35 \\ 
  MOS-12 & 13.32 \\ 
  MOS-13 & 10.41 \\ 
  PPSO & 8.81 \\ 
  RBI & 6.01 \\ 
  TFL & 10.58 \\ 
  LSCNE & 5.9 \\ 
  \hline
\end{tabular}
\caption{Mean Rank of Bayesian Friedman Test}
\label{tab:friedman-mean-rank}
\end{table}

\subsubsection{Bayesian Sign and Signed-Rank test}
\label{bayesian-sign-and-signed-rank-test}

The original proposal of the use of the Bayesian Sign and Signed-Rank
tests included in \autoref{sec:bayesian-sign-signed} is the comparison
of classification algorithms and the proposed \textit{rope} is
$[-0.01,0.01]$ for a measure in the range $[0,1]$. In the scenario of
optimisation problems, we should be concerned that the possible
outcomes are lower-bounded by 0 but in many functions, there is not an
upper bound or the maximum is very high, so we must follow another
approach. As the difference in the 100 dimension comparison is between
0 and 15000, we state that the region of practical equivalence is
$[-10,10]$.

The tests compute the probability of the true location of
$\textrm{EBO-CMAR} - \textrm{jSO}$ with respect to $0$, so both tests'
results shows that there is a similar probability for the three
hypotheses. In the Bayesian Sign test results, the hypothesis with a
greater probability is the left region (\ie the true location is less
than 0 and then jSO obtain worse results), although the results are
not significant enough to state that EBO is the winner
algorithm. Following the results of the Bayesian Signed-Rank test,
left region is also the most probable, although this is not
significant. Rope probability is also high, so the equivalence cannot
be discarded, which means that there are several ties in the benchmark
results.

\begin{figure}[h]
  \centering
  \includegraphics[width=.4\textwidth]{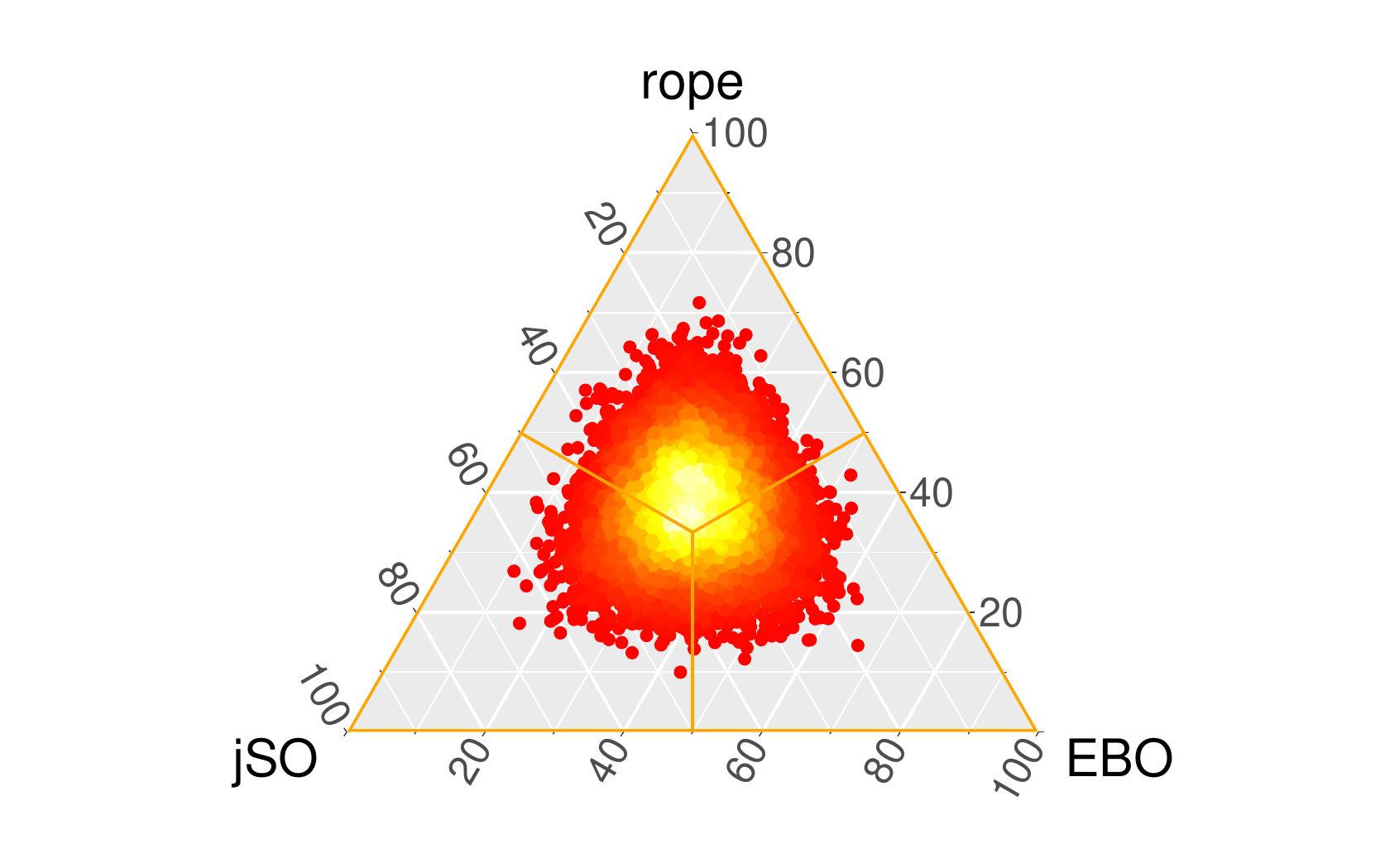}
  \includegraphics[width=.4\textwidth]{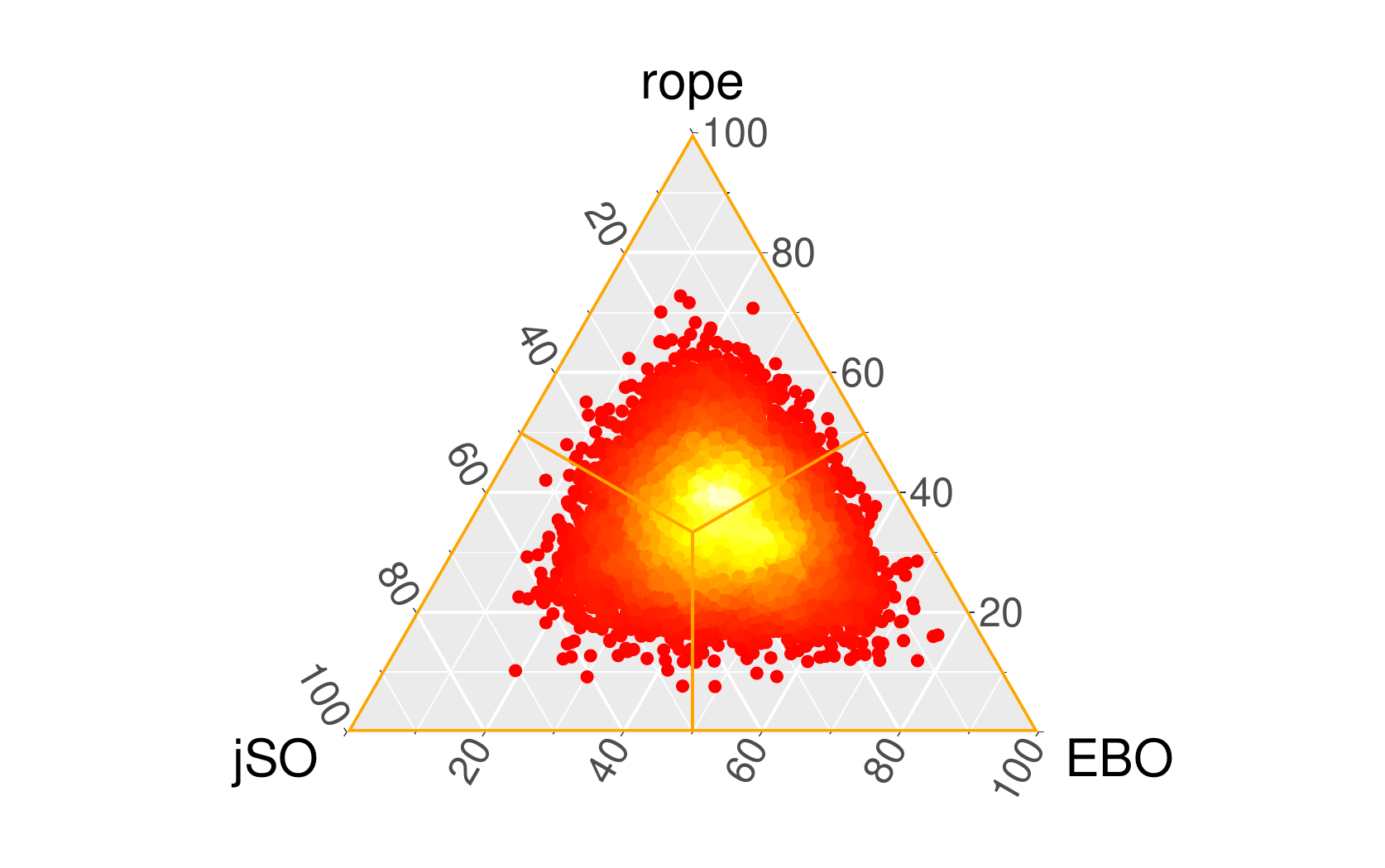}
  \caption{Bayesian Sign and Bayesian Signed-Rank tests}
  \label{fig:bayes-s-sr}
\end{figure}

We can see the posterior probability density of the parameter in
\autoref{fig:bayes-s-sr}, where each point represents an estimation of
the probability of the parameter which belongs to each region of
interest. The vertexes of the triangle represent the points where
there is probability 1 of the true location being in this region. The
proportion of the location of the points is shown in
\autoref{tab:wilcoxon-signed-rank}. This means that we have repeatedly
obtained the triplets of the probability of each region to be the true
location of the difference between the two samples, and then we have
plotted these triplets to obtain the posterior distribution of the
parameter. If we compare these results with a paired Wilcoxon test, we
see that the null hypothesis of the equivalence of the means is
rejected, although there is no information about if one algorithm
outperforms the other. However, using the Bayesian paradigm we can see
that this is not the situation, as we cannot establish the dominance
of one algorithm over the other either. Following the frequentist
paradigm we could be tempted to (erroneously) establish that EBO is
better, according to a single statistic and the result of the Wilcoxon
test.



\begin{table}[ht] 
\centering 
\begin{tabular}{lll} 
  \hline
  \multirow{1}{*}{Wilcoxon Signed Ranks} \\ \hline
  Statistic & V & 655. \\ 
  p-value   &   & \textbf{0.002} \\ \hline

  \multirow{1}{*}{Bayesian Sign} \\ \hline 
            & jSO & 0.2777823  \\ 
  Posterior probability & rope & 0.3112033 \\ 
            & EBO & 0.4110144 \\ \hline

  \multirow{1}{*}{Bayesian Signed-Rank} \\ \hline
            & jSO & 0.2905800  \\ 
  Posterior probability & rope & 0.3409561   \\ 
            & EBO & 0.3684639  \\ \hline
\end{tabular}
\caption{Bayesian Sign and Signed Ranks tests}
\label{tab:wilcoxon-signed-rank}
\end{table}

\subsubsection{Imprecise Dirichlet Process Test}
\label{imprecise-dirichlet-process-test}

The Imprecise Dirichlet Process, as we have seen in
\autoref{sec:impr-dirichl-proc}, consists of a more complex idea of
the previous tests although the implications of the use of the
Bayesian Tests could be clearer. In this test, we try to not introduce
any prior distribution, not even the prior distribution where both
algorithms have the same performance, but all the possible probability
measures $\alpha$, and then obtain an upper and a lower bounds for the
probability in which we are interested. The input consists of the
aggregated results among the different runs for all the benchmark
functions for a single dimension. The other parameters of the function
are the strong parameter $s$ and the pseudo-observation $c$. With
these data, we obtain two bounds for the posterior probability of the
first algorithm outperforming the second one, i.e.~the probability of
$P(X \leq Y) \geq 0.5$. These are the possible scenarios:

\begin{itemize}
 
\item Both bounds are greater than 0.95: Then we can say that the
  first algorithm outperforms the second algorithm with 95\%
  probability.
\item Both bounds are lower than 0.05: This is the inverse case. In
  this situation, the second algorithm outperforms the first algorithm
  with 95\% probability.
\item Both bounds are between 0.05 and 0.95: Then we can say that the
  probability of one algorithm outperforming the other is lower than
  the desired probability of 0.95.
\item Finally, if only one of the bounds is greater than 0.95 or lower
  than 0.05, the situation is undetermined and we cannot decide.
\end{itemize}


\begin{table}[ht] 
\centering 
\begin{tabular}{lll} 
\hline
\multicolumn{3}{c}{IDP - Wilcoxon test} \\ \hline
  \multirow{2}{*}{Posterior Distribution}
  & Upper Bound & 0.591 \\ 
  & Lower Bound & 0.452 \\ \hline
\end{tabular}
\caption{Imprecise Dirichlet Process of Wilcoxon test}
\label{tab:idp}
\end{table}

\begin{figure}[H]
  \centering
  \includegraphics[width=.9\textwidth]{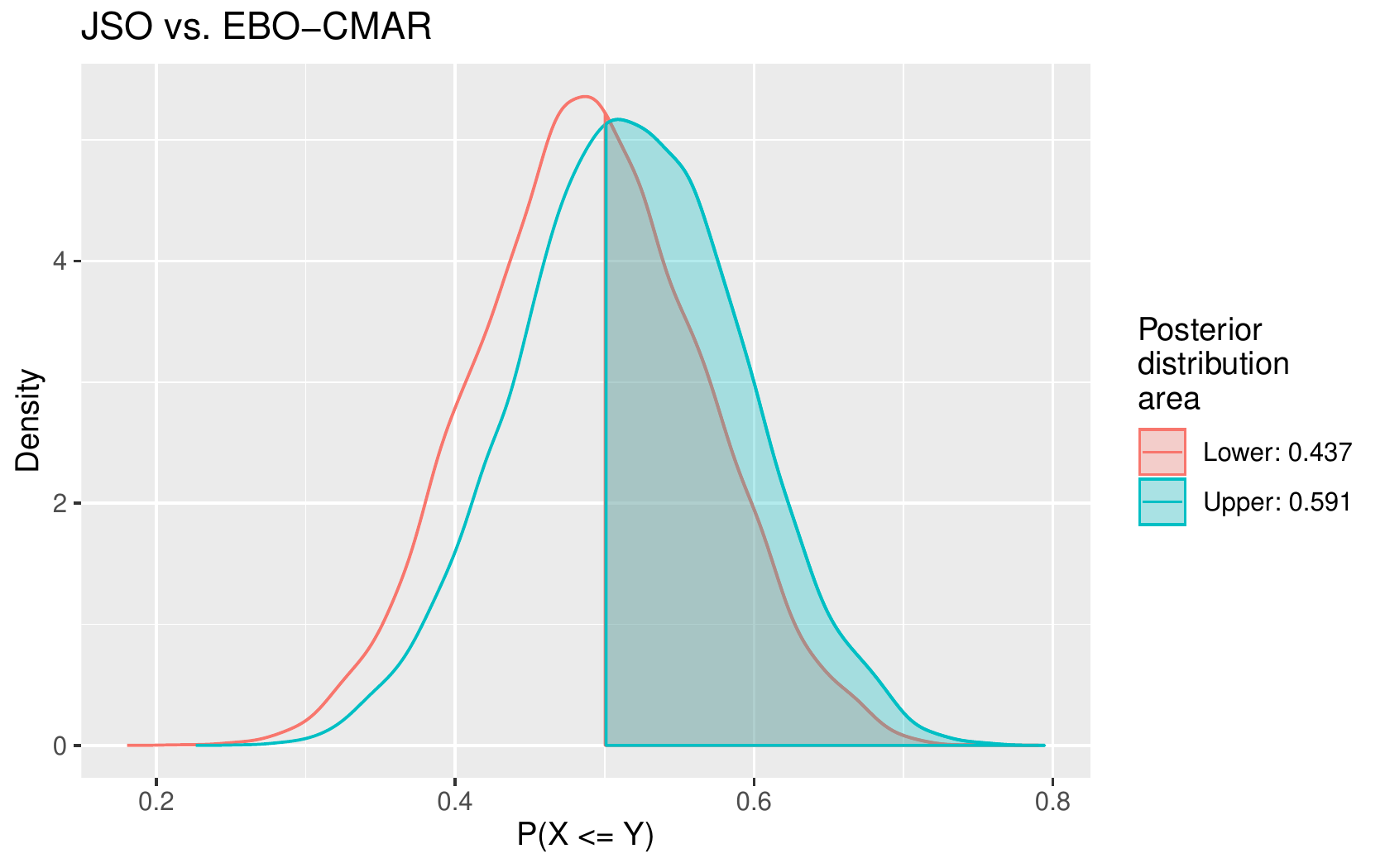}
  \caption{Imprecise Dirichlet Process - Wilcoxon test}
  \label{fig:idp}
\end{figure}

According to the results of the Bayesian Imprecise Dirichlet Process
(see \autoref{tab:idp}), the probability under the Dirichlet Process
of $P(EBO \leq jSO) \geq 0.5$, that is the probability of EBO-CMAR
outperforming jSO, is between 0.59 and 0.45, so there is not a
probability greater than 0.95 of EBO-CMAR outperforming jSO. These
numbers represent the area under the curve of the upper and lower
distributions when $P(X \leq Y) \geq 0.5$. In \autoref{fig:idp} we can
see both posterior distributions.

\subsection{Multi-Objective Comparison}
\label{sec:multi-object-comp}

In some circumstances like the Multi-Objective Optimisation, we need
to include different measures in the comparison. We include in this
section the illustration of the use of the tests presented in
\autoref{sec:mult-meas-tests}.

\subsubsection{Multiple measures test - GLRT}
\label{multiple-measures-test---glrt}

As we have mentioned in \autoref{parametric-analysis} concerning the
Hotelling's $T^2$ test, we can be interested in the simultaneous
comparison of multiple measures. This is the scenario of application
of the Non-Parametric Multiple Measures test, described in
\autoref{multiple-measures-test---glrt}. We select the means of the
executions of the two best algorithms and reshape them into a matrix
with the results of each benchmark in the rows and the different
dimensions in the columns. Then we use the test to see which
hypothesis of dominance is the most probable and if we can state that
the probability of this dominance statement is significant. According
to the results shown in \autoref{tab:glrt}, we obtain that the most
observed dominance statement is the configuration $[<,<,<,<]$, it is,
EBO-CMAR obtains a better result for all the dimensions. However, the
associated null hypothesis which states that the mentioned
configuration is no more probable than the following one, obtain a
$p$-value of 0.39, showed in \autoref{tab:glrt}, so this hypothesis
cannot be rejected. The second most probable configuration is
$[>,>,>,>]$ which means that EBO-CMAR obtains worse results in all the
dimensions, so we cannot be certain which is the most probable
situation in any scenario. It is relevant to note that the number of
observations can be a real value as the weight of an observation is
divided between the possible configuration when there is a tie for any
measure.


\begin{table}[h]
    \centering
    \begin{tabular}{ll}
      \\ \hline
      \multicolumn{2}{c}{GLRT Multiple Measures} \\ \hline
       $\lambda$ & 0.6917945 \\ \hline
        Configuration & Number of observations \\ \hline
  $<$ $<$ $<$ $<$ & 8.25 \\ 
  $<$ $<$ $<$ $>$ & 2.62 \\ 
  $<$ $<$ $>$ $<$ & 1.25 \\ 
  $<$ $<$ $>$ $>$ & 0.12 \\ 
  $<$ $>$ $<$ $<$ & 1.25 \\ 
  $<$ $>$ $<$ $>$ & 0.12 \\ 
  $<$ $>$ $>$ $<$ & 1.25 \\ 
  $<$ $>$ $>$ $>$ & 3.12 \\ 
  $>$ $<$ $<$ $<$ & 0.75 \\ 
  $>$ $<$ $<$ $>$ & 1.62 \\ 
  $>$ $<$ $>$ $<$ & 1.25 \\ 
  $>$ $<$ $>$ $>$ & 0.12 \\ 
  $>$ $>$ $<$ $<$ & 0.75 \\ 
  $>$ $>$ $<$ $>$ & 0.12 \\ 
  $>$ $>$ $>$ $<$ & 1.25 \\ 
  $>$ $>$ $>$ $>$ & 5.12 \\  \hline
        $p$-value & 0.3906 \\ \hline
    \end{tabular}
    \caption{Posterior configuration probability}
    \label{tab:glrt}
\end{table}

\subsubsection{Bayesian Multiple Measures Test}
\label{bayesian-multiple-measures-test}

In the Bayesian version of the Multiple Measures test the results are
analogous to the frequentist version shown in the previous section. We
use the same matrices with the results of each algorithm arranged by
dimensions. Here we obtain that the most probable configuration is
also the dominance of EBO in all the dimensions according to this
test, but the posterior probability is 0.75, as is shown in
\autoref{tab:bay-mmt}, so we cannot say that the difference with
respect to the remaining configurations is determinant.


\begin{table}[ht]
    \centering
    \begin{tabular}{lS}
      \hline
      Configuration & {Probability} \\
      \hline
  $<$ $<$ $<$ $<$ & 0.75 \\ 
  $<$ $<$ $<$ $>$ & 0.02 \\ 
  $<$ $<$ $>$ $<$ & 0.00 \\ 
  $<$ $<$ $>$ $>$ & 0.00 \\ 
  $<$ $>$ $<$ $<$ & 0.00 \\ 
  $<$ $>$ $<$ $>$ & 0.00 \\ 
  $<$ $>$ $>$ $<$ & 0.00 \\ 
  $<$ $>$ $>$ $>$ & 0.03 \\ 
  $>$ $<$ $<$ $<$ & 0.00 \\ 
  $>$ $<$ $<$ $>$ & 0.01 \\ 
  $>$ $<$ $>$ $<$ & 0.00 \\ 
  $>$ $<$ $>$ $>$ & 0.00 \\ 
  $>$ $>$ $<$ $<$ & 0.00 \\ 
  $>$ $>$ $<$ $>$ & 0.00 \\ 
  $>$ $>$ $>$ $<$ & 0.00 \\ 
  $>$ $>$ $>$ $>$ & 0.17 \\ 
      \hline
    \end{tabular}
    \caption{Posterior configuration probability}
    \label{tab:bay-mmt}
\end{table}

\section{Summary of results in CEC'2017}
\label{sec:summ-results-cec2017}

In this section, we include a summary of the
statistical analysis performed within the context of the CEC'2017.

\autoref{tab:CEC17scores} shows the official results of the algorithms
in their scoring system and the scores computed following the
indications of the report of the problem definition for the
competition \cite{2016-Awad-ProblemDefinitionsEvaluation} with the
available raw results of the algorithms. The $Score1$ is defined using
a weighted sum of the errors of the algorithms in all benchmark on
different dimensions. The weights are $0.1, 0.2, 0.3, 0.4$, with the
higher weights corresponding with the higher dimension
scenarios. Then, if we call $SE$ the summed error for an algorithm and
$SE_{\operatorname{min}}$ the minimum sum for an algorithm, $Score1$
is defined as
$Score1 = 0.5 * (1 - \frac{SE -
  SE_{\operatorname{min}}}{SE})$. Analogously, $Score2$ is defined as
a weighted sum of the ranks of the algorithms instead of using the
error: $Score2 = 0.5 * (1 - \frac{SR - SR_{\operatorname{min}}}{SR})$,
where $SR$ is the weighted sum of the ranks. The difference between
the official score and our computation may reside in the aggregation
method for the results of the different runs or the programming
language used for the computation of the scores. Different versions of
the results have been used without major impact in the final ranking,
which proves the robustness of the ranking and the
algorithms. However, these differences do not affect the final ranking
of the first classified algorithms. In the official CEC'17 summary,
the results of the MOS12 algorithm are not included, so we have
excluded them in the analyses of this section.

\begin{table}[ht]
\centering
\begin{tabular}{|lrrr|lrrr|}
  \hline
  \multicolumn{4}{|c|}{Official CEC'17 Results} & \multicolumn{4}{c|}{Score Computation}\\
  \hline
  Algorithm & Score1 & Score2 & Score & Algorithm & Score1 & Score2 & Score \\ 
  \hline
  EBO & \textbf{50.00} & 48.01 & \textbf{98.01} & EBO & \textbf{50.00} & \textbf{50.00} & \textbf{100.00} \\ 
  jSO & 49.69 & 47.08 & 96.77 &                   jSO & 49.69 & 43.01 & 92.70 \\  
  LSCNE & 46.82 & 49.74 & 96.56 &                 LSCNE & 46.82 & 44.75 & 91.56 \\
  LSSPA & 46.44 & \textbf{50.00} & 96.44 &        LSSPA & 46.44 & 44.73 & 91.17 \\
  DES & 45.94 & 43.20 & 89.14 &                   DES & 45.94 & 40.65 & 86.59 \\  
  MM & 45.96 & 40.12 & 86.07 &                    MM & 45.96 & 36.16 & 82.12 \\   
  IDEN & 29.85 & 27.68 & 57.53 &                  IDEN & 29.85 & 26.15 & 56.00 \\ 
  RBI & 3.79 & 33.61 & 37.40 &                    MOS13 & 18.94 & 17.33 & 36.27 \\
  MOS13 & 18.94 & 17.34 & 36.28 &                 RBI & 3.79 & 32.00 & 35.79 \\   
  MOS11 & 11.09 & 19.30 & 30.39 &                 MOS11 & 11.09 & 19.17 & 30.25 \\
  PPSO & 3.93 & 17.36 & 21.28 &                   PPSO & 3.93 & 17.26 & 21.19 \\  
  DYYPO & 0.59 & 17.03 & 17.62 &                  DYYPO & 0.59 & 17.06 & 17.65 \\ 
  TFL & 0.03 & 16.25 & 16.27 &                    TFL & 0.03 & 16.31 & 16.34 \\   
  \hline
\end{tabular}
\caption{CEC17 Results Scores with mean results} 
\label{tab:CEC17scores}
\end{table}

The classic statistical analysis that should be made in the context of
a competition would involve a non-parametric test with post-hoc test
for a $n$ versus $n$ scenario, as we do not have a preference for the
results of comparison of any specific algorithm. In order to preserve
the relative importance of the results in the different dimension
scenarios, we show the plots of the critical difference explained in
\autoref{sec:post-hoc-procedures} for the four scenarios.

\begin{figure}[H]
  \begin{subfigure}{.5\textwidth}
    \centering
  \includegraphics[trim={1cm 2.5cm 1cm 2cm}, clip, width=\linewidth]{cd-plot-dim10.pdf}
  \caption{CD plot Dim10}
  \label{fig:cd-plot-dim10}
  \end{subfigure}
  \begin{subfigure}{.5\textwidth}
    \centering
  \includegraphics[trim={1cm 2.5cm 1cm 2cm}, clip, width=\linewidth]{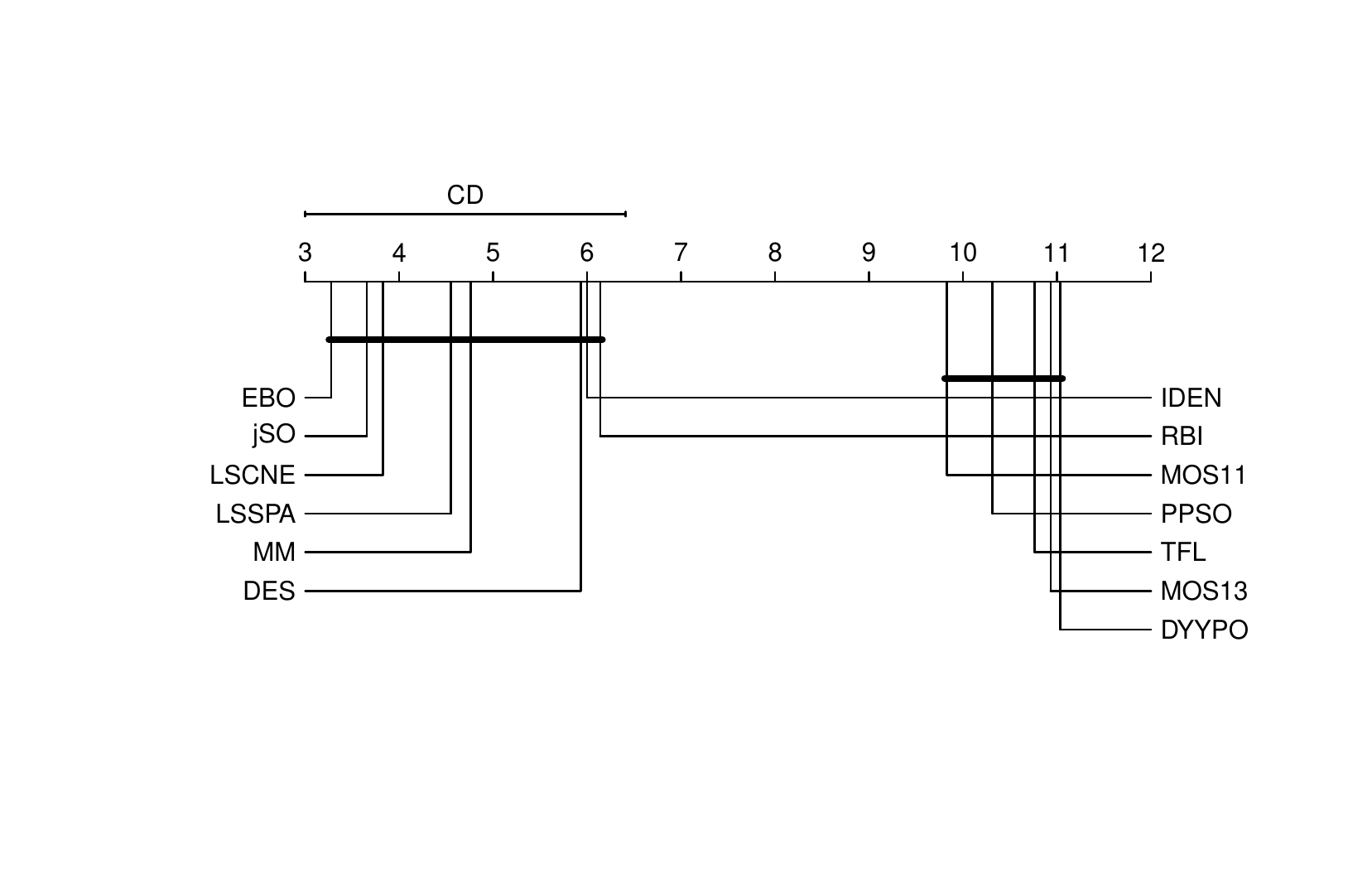}
  \caption{CD plot Dim30}
  \label{fig:cd-plot-dim30}
\end{subfigure}
  \begin{subfigure}{.5\textwidth}
    \centering
  \includegraphics[trim={1cm 2.5cm 1cm 2cm}, clip, width=\linewidth]{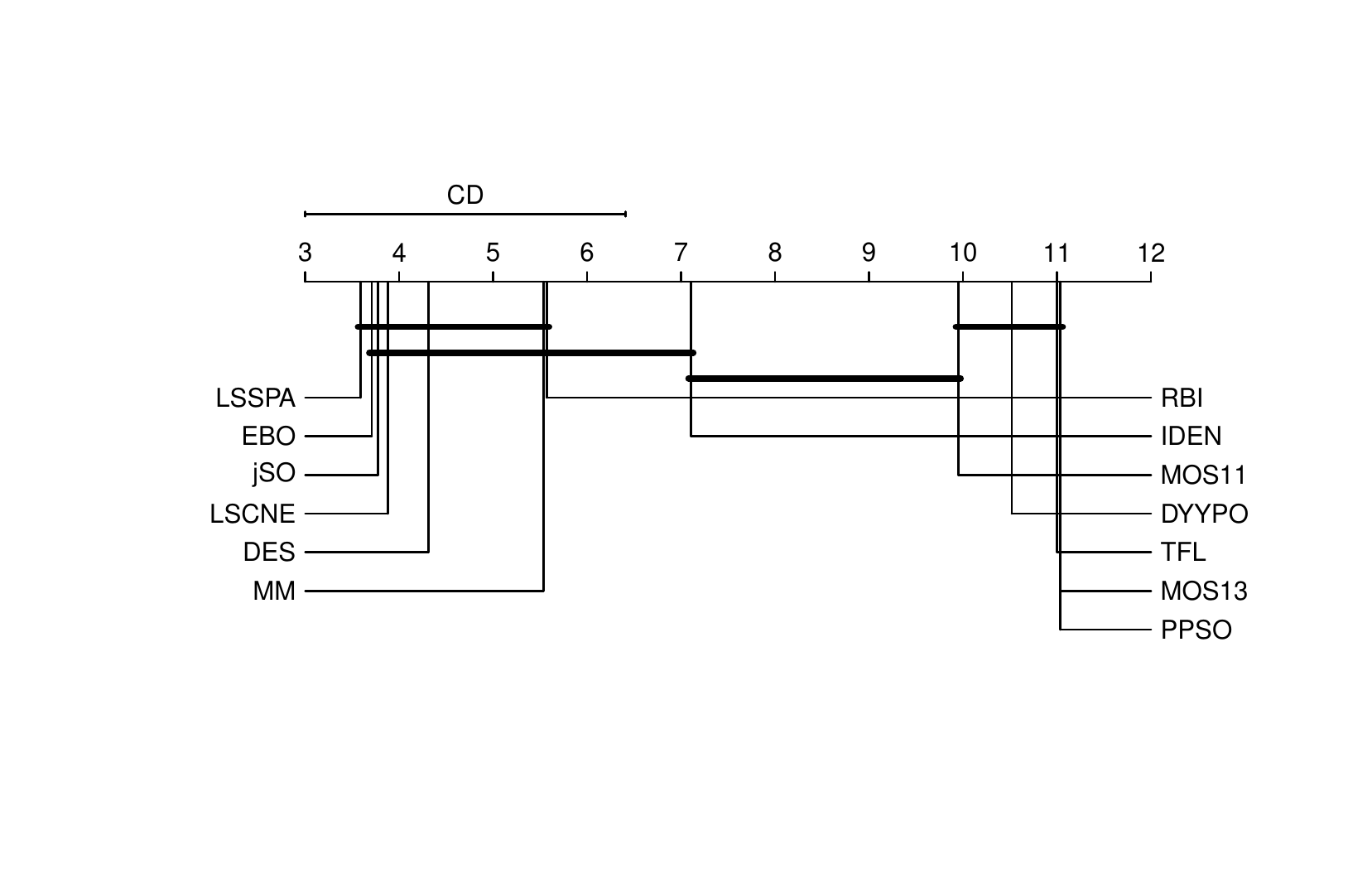}
  \caption{CD plot Dim50}
  \label{fig:cd-plot-dim50}
  \end{subfigure}
  \begin{subfigure}{.5\textwidth}
    \centering
  \includegraphics[trim={1cm 2.5cm 1cm 2cm}, clip, width=\linewidth]{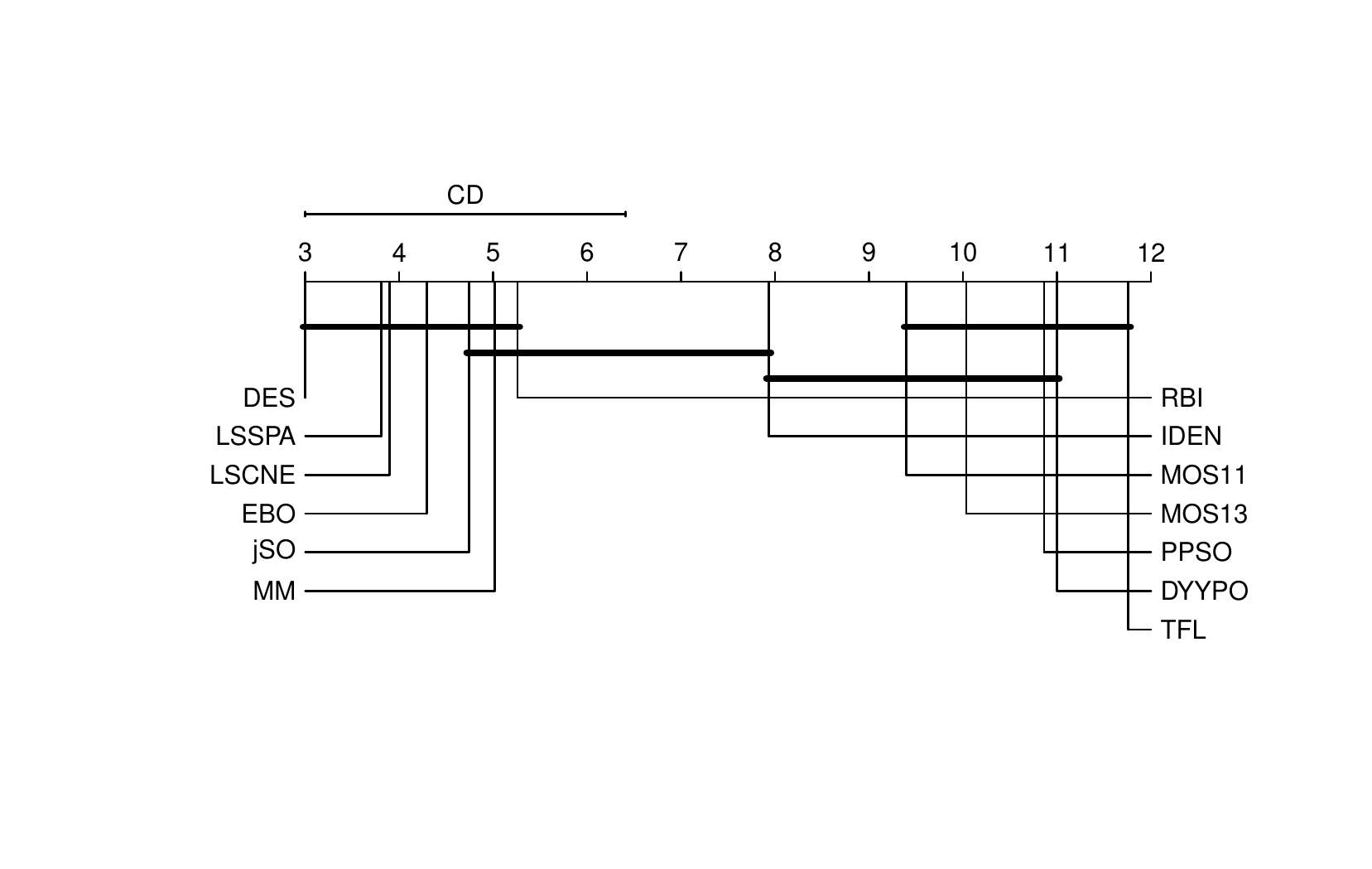}
  \caption{CD plot Dim100}
  \label{fig:cd-plot-dim100}
\end{subfigure}
\caption{Plots of Critical Differences}
\label{fig:cd-plots}
\end{figure}

As we can see in \autoref{fig:cd-plots}, summary scores presented in
\autoref{tab:CEC17scores} are consistent with the Critical Differences
plots made with the mean values, as the first classified algorithms
are also in the first positions of the graphical
representation. However, the statistical tests make it possible to
address the fact that there is a group of algorithms in the lead group
in every dimension scenario whose associated hypothesis of equivalence
cannot be discarded. These algorithms are LSSPA, DES, LSCNE, MM, jSO
and EBO. Moreover, there is not a single scenario where the winner
equivalences with this lead group algorithms can be discarded.

In the Bayesian paradigm, after having rejected the equivalence of all
the mean ranks of the algorithms with the Friedman tests, we
repeatedly perform the Bayesian Signed-Rank for every pair of
algorithms in each dimension scenario. The results are summarised in
Figures~7-10. Especially in lower dimensions, there
is less certainty than there was in the non-parametric analysis
concerning the dominance of an algorithm over the other in each
comparison, although from the Bayesian perspective we can state where
there is a tie between a pair of algorithms and the direction of the
difference, while the equivalence with NHST cannot be assured. The
tiles for each column and row represent the comparison between the two
algorithms. The colour depends on the result of the comparison,
indicating if the greater probability belongs to the region of an
algorithm or the rope. The probability of this hypothesis is written
in the tile as well as represented in the opacity of the colour, to
highlight the greater probabilities.

\begin{figure}[H]
    \centering
  \includegraphics[width=.8\textwidth]{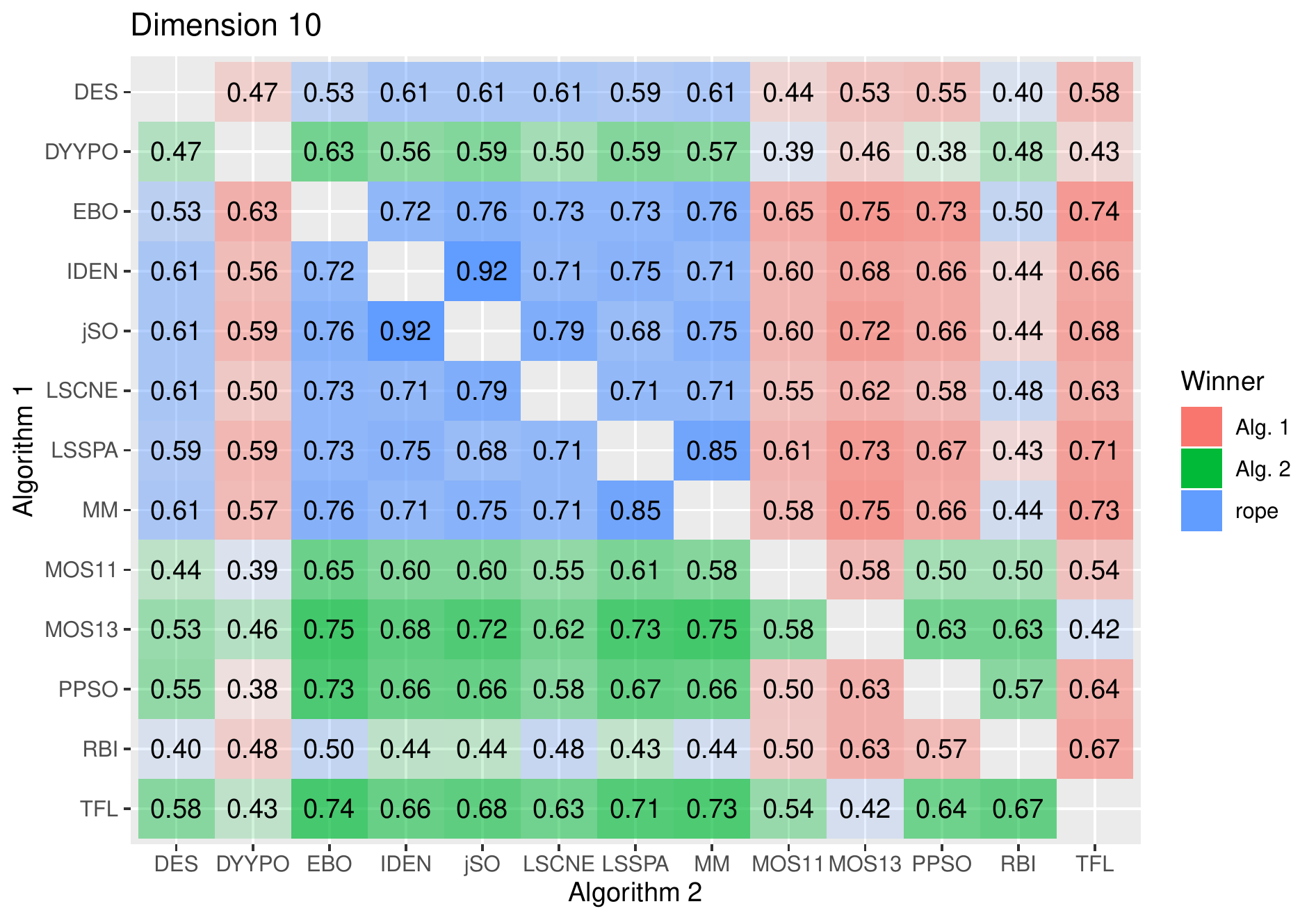}
  \caption{Bayesian Signed-Rank Dim10}
  \label{fig:bayesian-sr-dim10}
\end{figure}

\begin{figure}[H]
    \centering
  \includegraphics[width=.8\textwidth]{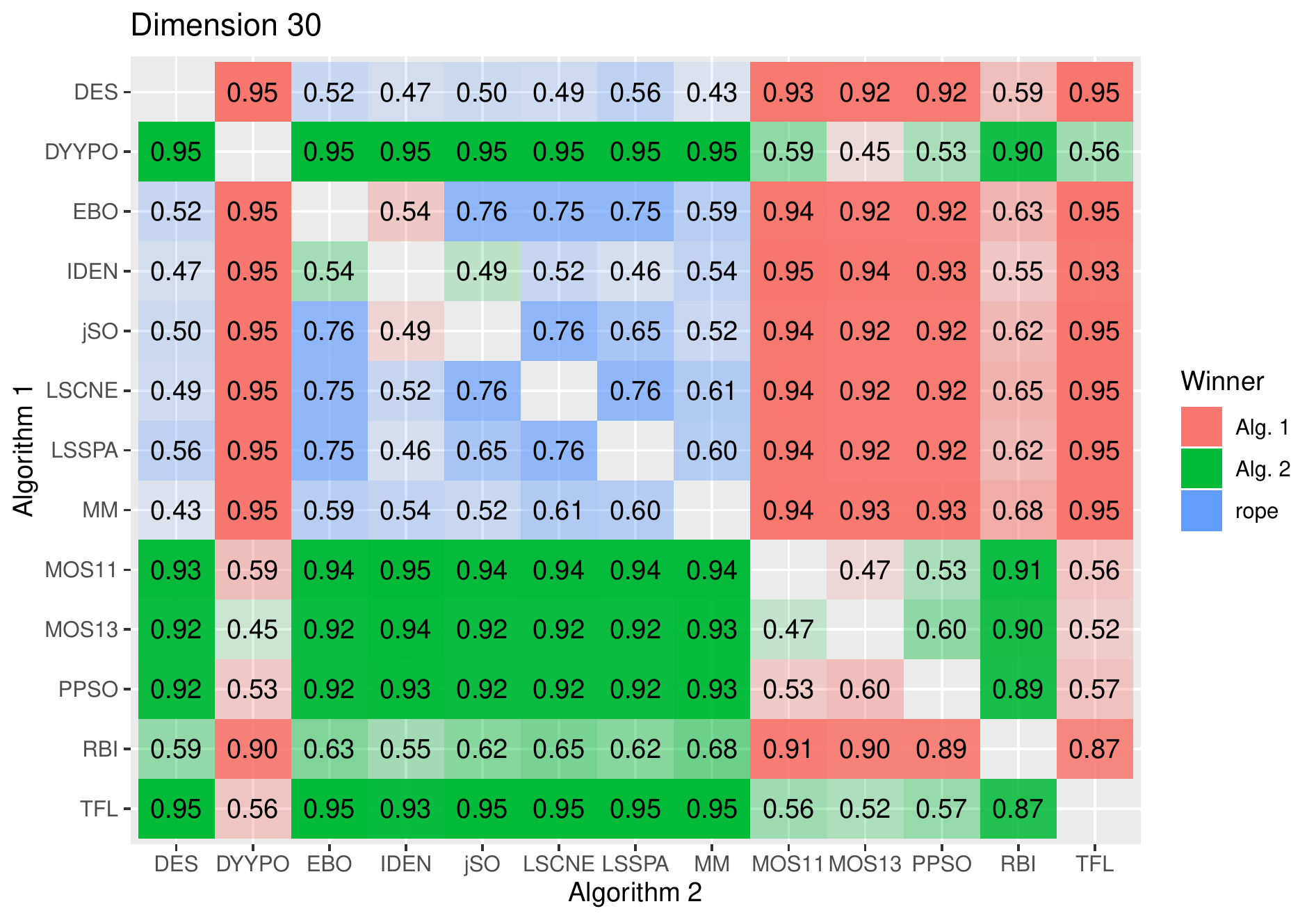}
  \caption{Bayesian Signed-Rank Dim30}
  \label{fig:bayesian-sr-dim30}
\end{figure}

As the error obtained in the competition increases, more comparisons
are marked as significant and less ties between algorithms are
detected. In \autoref{fig:bayesian-sr-dim10} we see how in the
comparison of the first classified algorithms the most probable
situation is a tie. This group starts to win with a greater
probability in the 30 Dimension scenario
(\autoref{fig:bayesian-sr-dim30}), while the ties persist within the
lead group.

\begin{figure}[H]
    \centering
  \includegraphics[width=.8\textwidth]{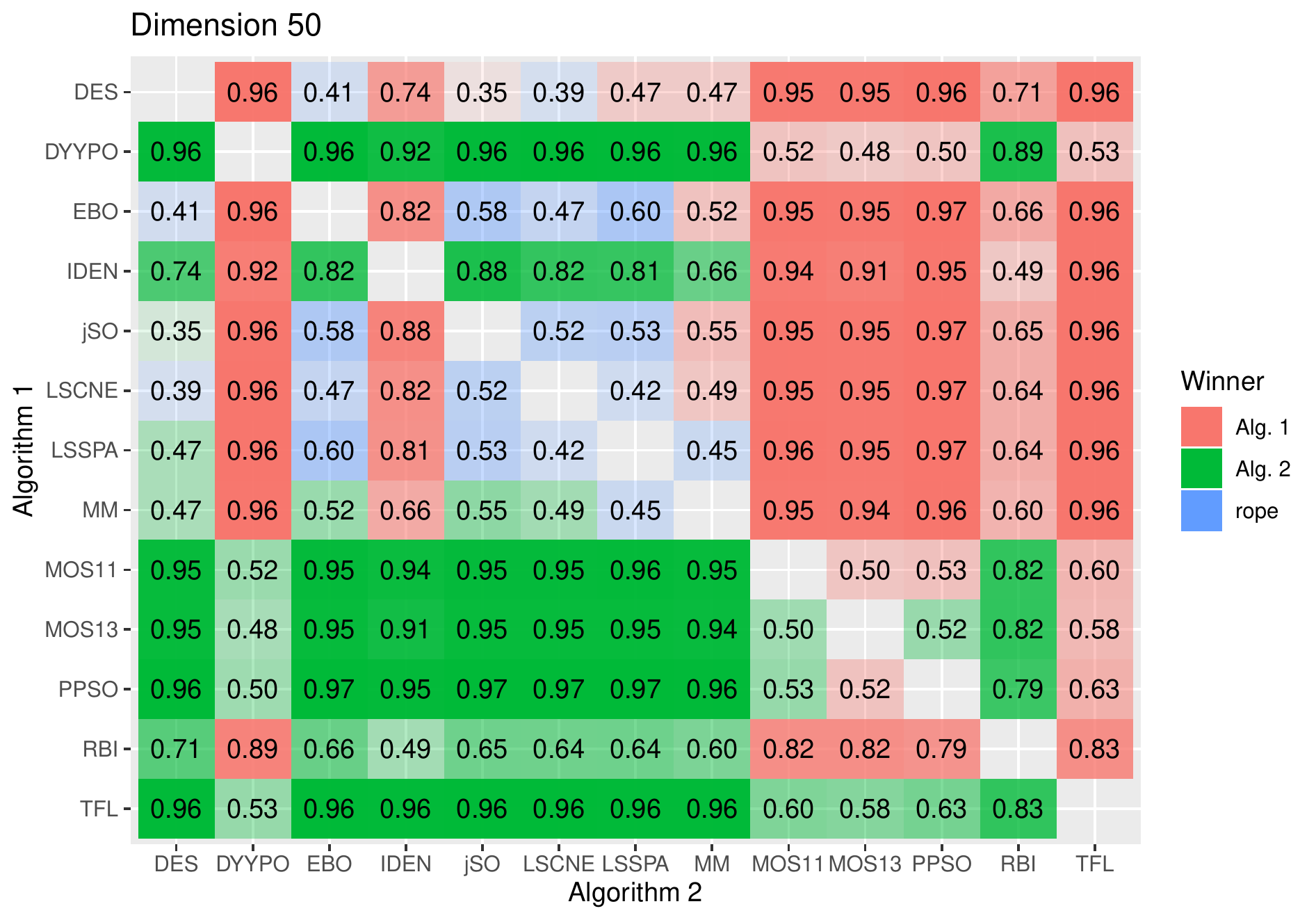}
  \caption{Bayesian Signed-Rank Dim50}
  \label{fig:bayesian-sr-dim50}
\end{figure}

The results of the 50 dimension scenario, shown in
\autoref{fig:bayesian-sr-dim50}, coincide with the conclusions
obtained in the non-parametric analysis. In this scenario the lead
group is reduced to EBO, jSO, LSCNE and DES, and the probabilities of
the ties are lower than in previous scenarios. Finally in 100
dimension scenario, DES wins in the comparisons with all the remaining
algorithms, with probabilities between 0.59 in the comparison versus
LSCNE and 0.97 versus PPSO.

\begin{figure}[H]
    \centering
  \includegraphics[width=.8\textwidth]{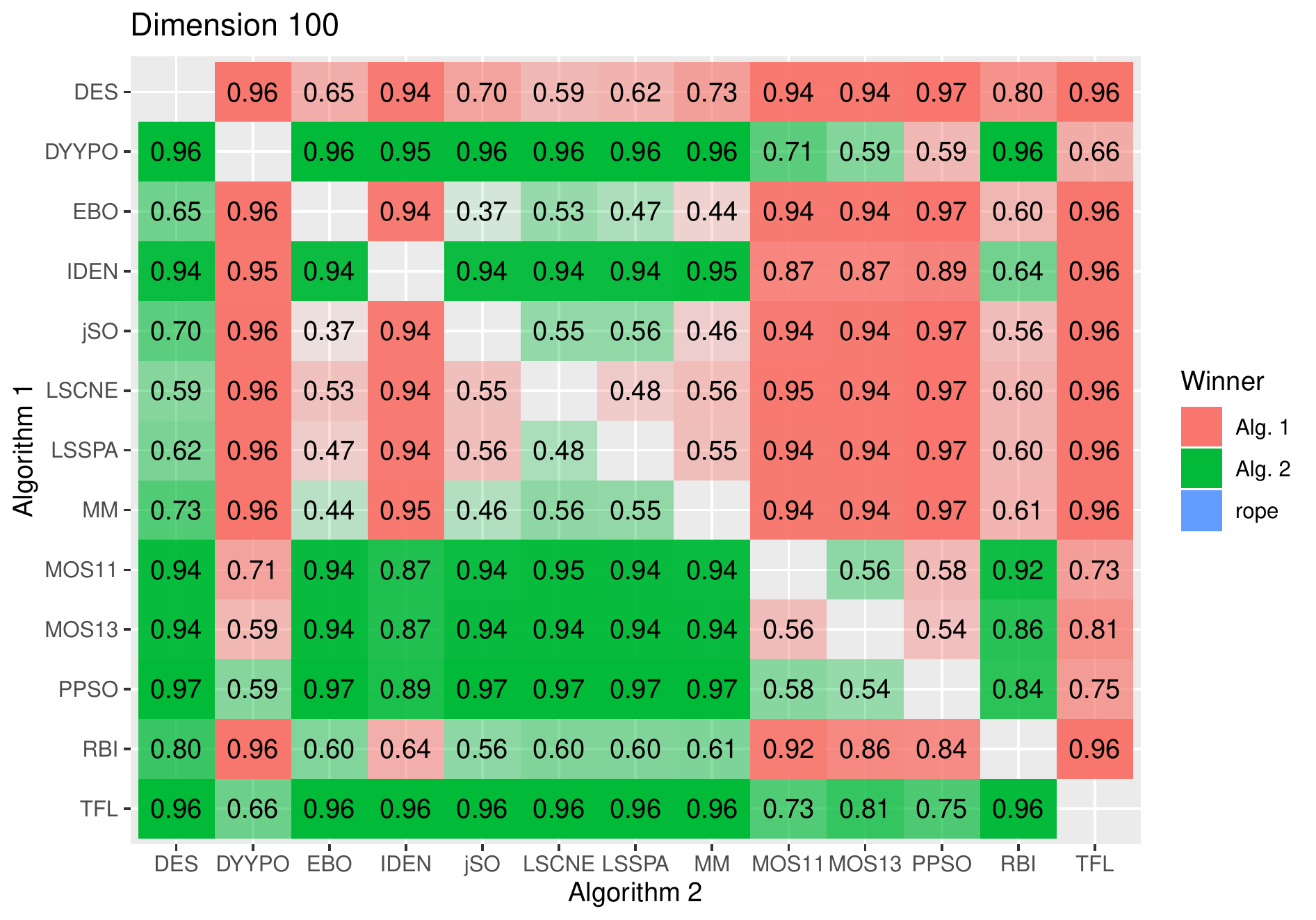}
  \caption{Bayesian Signed-Rank Dim100}
  \label{fig:bayesian-sr-dim100}
\end{figure}

\section{Discussion and Lessons Learnt}
\label{sec:misc-lessons-learnt}

In this section we include some considerations about the use of the
tests described in previous sections and other proposed tests.

\paragraph{Criticisms on the $p$-value}

The criticisms made regarding the $p$-value and NHST are not limited
to the Evolutionary Optimisation field or even Computer Science
\cite{2017-Berrar-JeffreysLindleyParadoxLooming}, but occur more
frequently in other research fields, like psychology
\cite{1998-Chow-Precisstatisticalsignificance} or neuroscience 
\cite{2016-Melinscak-pvaluesevaluationbrain}. 

A recent Nature paper \cite{2019-Amrhein-Scientistsrisestatistical}
warns about the common mistakes in the interpretation about the
meaning of the $p$-value, especially the statements
about the alleged ``no difference'' between the groups when the null
hypothesis is not rejected. This is one of the most direct and
powerful arguments for the promotion of the use of the Bayesian tests,
as the posterior distribution reflects the behaviour of the parameter
of interest, and it is more difficult to conclude that the algorithms'
performance is the same if it is not reflected in the plots of the
distribution. The other main practical criticism is related to the
researcher's intention and the effect size, as an elevated number of
samples could derive in the rejection of the null hypothesis even with
a tiny effect size. The Bayesian approach is not affected in the same
way, as increasing the number of samples should be reflected in a
posterior distribution closer to the underlying one.

Another controversial aspect of the NHST is their
performance in a dichotomous way in order to determine the result of
the experiment. However, these criticisms are not restricted to the
frequentist paradigm and also affects the Bayesian
perspective. Similar opinions appeared in ASA's statement
\cite{2016-Wasserstein-ASAStatementpValues}, which represents a major
setback to the use of NHST. They do recognise that
to obtain reliable and repeatable results, diverse
summaries of the data must be offered and the underlying process must
be clearly understood and none of them can be substituted with a
single index. This association now makes some proposals and gives some
suggestions to avoid these pitfalls
\cite{2019-Wasserstein-MovingWorld05}, like not asserting
``statistically significant'', accepting uncertainty or being more
open to sharing the data and procedures.

\paragraph{Criticisms on the Bayesian paradigm} Bayesian paradigm has
a clear drawback of more complex tests and results,
which causes this kind of test to be used less in
experimentation. Moreover, some criticisms made regarding NHST and the
decisions that depend on the researcher's intention, like the $\alpha$
value or the number of samples, remain in the Bayesian paradigm.

These decisions are the family and parameters from the prior
distribution. The Bayesian tests described in this paper also contain
the \textit{rope} parameter for the equivalence of two algorithms,
whose bounds are set to $0.01$ in the original proposal for a
classification scenario. Opting to adjust the results using Bayesian
inference, selecting the prior distribution and making statements
using this adjustment would lead to a model that could have a low
representation of the data. Then a special consideration should be
made in order to select a representative prior, with less information
concerning the parameter of interest.

\paragraph{Frequentist and Bayesian Tests Relation} Although there is
a clear difference between the frequentist and the Bayesian paradigms,
we could still hope to see a relationship between the results of both
kinds of tests
\cite{2018-Silva-correspondencefrequentistBayesian}. In this line of
argument, Couso \etal \cite{2017-Couso-ReconcilingBayesianFrequentist}
built a frequentist test that simulates the behavior of the IDP test
presented in \autoref{sec:bayesian-sign-signed} for when the
underlying distribution satisfies some properties. This study shows
that in some circumstances the imprecise scenario using a set of prior
distributions suggested from the Bayesian perspective is reflected in
the distribution of $p$-values in the frequentist version.

Therefore, there is not a single way of obtaining conclusions, and
they are not exempted from possible misuses or misunderstandings. The
proliferation of the statistical tests and their application in
erroneous circumstances can lead to spurious conclusions. Thus, the
use of different tests can help to put the results in context. These
considerations are independent of the guidelines for
the experimentation in the field of evolutionary optimisation. The
algorithms involved in the comparison should represent the
state-of-the-art and the used benchmark should be relevant in the
field of study.

\section{Conclusions}
\label{sec:conclusions}

This work contains an exhaustive set of statistical procedures and
examples of their use in the comparison of results from experimental
studies of optimisation algorithms, specifically in the scenario of a
competition. In this paper we have described a broad set of methods,
from basic non-parametric tests such as the Binomial Sign test to
recently proposed Bayesian techniques such as Imprecise Dirichlet
Process, with tools that can help researchers to complete their
experimental study.

In this paper, we undertake an extensive statistical analysis of the
results of the CEC '17 Special Session and Competition on
Single-Objective Real Parameter Optimisation. The statistical analysis
of the competition results provides different results according to the
test used in different circumstances. Then, we have used this case of
study to make some recommendations about which test is the most
appropriate to each situation and how to proceed with the information
that we get from them. Although the used paradigm depends on the
intention of the researchers, they should consider using parametric
tests only if the normality and homoscedasticity prerequisites are
fulfilled. We have described some tests for the comparison of two
algorithms, and the procedure when more algorithms are involved in the
comparison, with specific tests for this purpose and some post-hoc
procedures that indicate where the differences are.

Finally, we encourage the joint use of non-parametric and Bayesian
tests in order to obtain a complete perspective of the comparison of
the algorithms' results. While non-parametric tests can provide
significant results when there is a difference between the compared
algorithms, in some circumstances these tests do not provide any
valuable information and Bayesian tests can help to elucidate the real
difference between them. Furthermore, the rNPBST package and the
\texttt{shinytests} application implement both perspectives and the
procedures described in this tutorial.

\section*{Acknowledgements}

This work is supported by the Spanish Ministry of Economy, Industry
and Competitiveness under the Spanish National Research Project
TIN2017-89517-P. J. Carrasco was supported by the Spanish Ministry of
Science under the FPU Programme 998758-2016 and by a scholarship of
initiation to research granted by the University of Granada


\begin{thebibliography}{10}
\expandafter\ifx\csname url\endcsname\relax
  \def\url#1{\texttt{#1}}\fi
\expandafter\ifx\csname urlprefix\endcsname\relax\def\urlprefix{URL }\fi
\expandafter\ifx\csname href\endcsname\relax
  \def\href#1#2{#2} \def\path#1{#1}\fi

\bibitem{2019-Hellwig-Benchmarkingevolutionaryalgorithms}
M.~Hellwig, H.-G. Beyer, Benchmarking evolutionary algorithms for single
  objective real-valued constrained optimization \textendash{} {{A}} critical
  review, Swarm and Evolutionary Computation 44 (2019) 927--944.
\newblock \href {https://doi.org/10.1016/j.swevo.2018.10.002}
  {\path{doi:10.1016/j.swevo.2018.10.002}}.

\bibitem{2008-Demsar-appropriatenessstatisticaltests}
J.~Dem{\v s}ar, On the appropriateness of statistical tests in machine
  learning, in: Workshop on {{Evaluation Methods}} for {{Machine Learning}} in
  Conjunction with {{ICML}}, 2008, p.~65.

\bibitem{2003-Sheskin-Handbookparametricnonparametric}
D.~J. Sheskin, Handbook of Parametric and Nonparametric Statistical Procedures,
  {crc Press}, 2003.

\bibitem{2009-Garcia-studyusenonparametric}
S.~Garc{\'i}a, D.~Molina, M.~Lozano, F.~Herrera, A study on the use of
  non-parametric tests for analyzing the evolutionary algorithms' behaviour: A
  case study on the {{CEC}}'2005 {{Special Session}} on {{Real Parameter
  Optimization}}, J. Heuristics 15~(6) (2009) 617--644.
\newblock \href {https://doi.org/10.1007/s10732-008-9080-4}
  {\path{doi:10.1007/s10732-008-9080-4}}.

\bibitem{2014-Derrac-Analyzingconvergenceperformance}
J.~Derrac, S.~Garc{\'i}a, S.~Hui, P.~N. Suganthan, F.~Herrera, Analyzing
  convergence performance of evolutionary algorithms: {{A}} statistical
  approach, Information Sciences 289 (2014) 41--58.

\bibitem{2017-Berrar-Confidencecurvesalternative}
D.~Berrar, Confidence curves: An alternative to null hypothesis significance
  testing for the comparison of classifiers, Machine Learning 106~(6) (2017)
  911--949.

\bibitem{2003-Gelman-BayesianDataAnalysis}
A.~Gelman, J.~B. Carlin, H.~S. Stern, D.~B. Rubin, Bayesian {{Data Analysis}},
  {{Second Edition}} ({{Chapman}} \& {{Hall}}/{{CRC Texts}} in {{Statistical
  Science}}), 2nd Edition, {Chapman and Hall/CRC}, 2003, published: Hardcover.

\bibitem{2017-Benavoli-TimeChangeTutorial}
A.~Benavoli, G.~Corani, J.~Dem{\v s}ar, M.~Zaffalon, Time for a {{Change}}: A
  {{Tutorial}} for {{Comparing Multiple Classifiers Through Bayesian
  Analysis}}, Journal of Machine Learning Research 18~(77) (2017) 1--36.

\bibitem{2011-Japkowicz-EvaluatingLearningAlgorithms}
N.~Japkowicz, M.~Shah (Eds.), Evaluating {{Learning Algorithms}}: {{A
  Classification Perspective}}, {Cambridge University Press}, 2011.

\bibitem{2013-Odile-StatisticalTestsNonparametric}
P.~Odile, Statistical {{Tests}} of {{Nonparametric Hypotheses}}: {{Asymptotic
  Theory}}, {World Scientific}, 2013.

\bibitem{2010-Gibbons-NonparametricStatisticalInference}
J.~D. Gibbons, S.~Chakraborti, Nonparametric {{Statistical Inference}}, {CRC
  Press}, 2010.

\bibitem{1988-Campbell-StatisticsMedicineCalculating}
M.~J. Campbell, M.~J. Gardner, Statistics in {{Medicine}}: {{Calculating}}
  confidence intervals for some non-parametric analyses, British medical
  journal (Clinical research ed.) 296~(6634) (1988) 1454.

\bibitem{2016-Awad-ProblemDefinitionsEvaluation}
N.~Awad, M.~Ali, J.~Liang, B.~Qu, P.~Suganthan, Problem {{Definitions}} and
  {{Evaluation Criteria}} for the {{CEC}} 2017 {{Special Session}} and
  {{Competition}} on {{Single Objective Real}} - {{Parameter Numerical
  Optimization}}, Tech. rep., {NTU, Singapore} (2016).

\bibitem{2017-Carrasco-rNPBSTPackageCovering}
J.~Carrasco, S.~Garc{\'i}a, M.~{del Mar Rueda}, F.~Herrera, {{rNPBST}}: {{An R
  Package Covering Non}}-parametric and {{Bayesian Statistical Tests}}, in:
  F.~J. {Mart{\'i}nez de Pis{\'o}n}, R.~Urraca, H.~Quinti{\'a}n, E.~Corchado
  (Eds.), Hybrid {{Artificial Intelligent Systems}}: 12th {{International
  Conference}}, {{HAIS}} 2017, {{La Rioja}}, {{Spain}}, {{June}} 21-23, 2017,
  {{Proceedings}}, {Springer International Publishing}, {Cham}, 2017, pp.
  281--292.
\newblock \href {https://doi.org/10.1007/978-3-319-59650-1_24}
  {\path{doi:10.1007/978-3-319-59650-1_24}}.

\bibitem{2006-Demsar-Statisticalcomparisonsclassifiers}
J.~Dem{\v s}ar, Statistical comparisons of classifiers over multiple data sets,
  Journal of Machine learning research 7~(Jan) (2006) 1--30.

\bibitem{1988-Looney-statisticaltechniquecomparing}
S.~W. Looney, A statistical technique for comparing the accuracies of several
  classifiers, Pattern Recognition Letters 8~(1) (1988) 5--9.
\newblock \href {https://doi.org/10.1016/0167-8655(88)90016-5}
  {\path{doi:10.1016/0167-8655(88)90016-5}}.

\bibitem{1998-Dietterich-Approximatestatisticaltests}
T.~G. Dietterich, Approximate statistical tests for comparing supervised
  classification learning algorithms, Neural computation 10~(7) (1998)
  1895--1923.

\bibitem{1999-Alpaydin-CombinedcvTest}
E.~Alpaydin, Combined 5 x 2 cv {{F Test}} for {{Comparing Supervised
  Classification Learning Algorithms}}, Neural Computation 11~(8) (1999)
  1885--1892.
\newblock \href {https://doi.org/10.1162/089976699300016007}
  {\path{doi:10.1162/089976699300016007}}.

\bibitem{2002-Castillo-Valdivieso-Statisticalanalysisparameters}
P.~{Castillo-Valdivieso}, J.~Merelo, A.~Prieto, I.~Rojas, G.~Romero,
  Statistical analysis of the parameters of a neuro-genetic algorithm, IEEE
  Transactions on Neural Networks 13~(6) (2002) 1374--1394.
\newblock \href {https://doi.org/10.1109/tnn.2002.804281}
  {\path{doi:10.1109/tnn.2002.804281}}.

\bibitem{2002-Pizarro-Multiplecomparisonprocedures}
J.~Pizarro, E.~Guerrero, P.~L. Galindo, Multiple comparison procedures applied
  to model selection, Neurocomputing 48~(1) (2002) 155--173.

\bibitem{2003-Nadeau-Inferencegeneralizationerror}
C.~Nadeau, Y.~Bengio, Inference for the generalization error, Machine Learning
  52~(3) (2003) 239--281.

\bibitem{2003-Chen-StatisticalComparisonsMultiple}
D.~Chen, X.~Cheng, Statistical {{Comparisons}} of {{Multiple Classifiers}}, in:
  Proceedings of the {{International Conference}} on {{Machine Learning}};
  {{Models}}, {{Technologies}} and {{Applications}}. {{MLMTA}}'03, {{June}} 23
  - 26, 2003, {{Las Vegas}}, {{Nevada}}, {{USA}}, 2003, pp. 97--101.

\bibitem{2004-Czarn-Statisticalexploratoryanalysis}
A.~Czarn, C.~MacNish, K.~Vijayan, B.~Turlach, R.~Gupta, Statistical exploratory
  analysis of genetic algorithms, IEEE Transactions on Evolutionary Computation
  8~(4) (2004) 405--421.

\bibitem{2006-Moskowitz-Comparingpredictivevalues}
C.~S. Moskowitz, M.~S. Pepe, Comparing the predictive values of diagnostic
  tests: Sample size and analysis for paired study designs, Clinical Trials
  3~(3) (2006) 272--279.

\bibitem{2006-Yildiz-Orderingfindingbesta}
O.~T. Yildiz, E.~Alpaydin, Ordering and finding the best of {{K}} gt; 2
  supervised learning algorithms, IEEE Transactions on Pattern Analysis and
  Machine Intelligence 28~(3) (2006) 392--402.
\newblock \href {https://doi.org/10.1109/TPAMI.2006.61}
  {\path{doi:10.1109/TPAMI.2006.61}}.

\bibitem{2007-Smucker-comparisonstatisticalsignificance}
M.~D. Smucker, J.~Allan, B.~Carterette, A comparison of statistical
  significance tests for information retrieval evaluation, in: Proceedings of
  the Sixteenth {{ACM}} Conference on {{Conference}} on Information and
  Knowledge Management, {ACM}, 2007, pp. 623--632.

\bibitem{2008-Garcia-ExtensionStatisticalComparisons}
S.~Garcia, F.~Herrera, An {{Extension}} on``{{Statistical Comparisons}} of
  {{Classifiers}} over {{Multiple Data Sets}}''for all {{Pairwise
  Comparisons}}, Journal of Machine Learning Research 9~(Dec) (2008)
  2677--2694.

\bibitem{2009-Aslan-Statisticalcomparisonclassifiers}
O.~Aslan, O.~T. Y\i{}ld\i{}z, E.~Alpayd\i{}n, Statistical comparison of
  classifiers using area under the roc curve (2009).

\bibitem{2009-Garcia-studystatisticaltechniques}
S.~Garc{\'i}a, A.~Fern{\'a}ndez, J.~Luengo, F.~Herrera, A study of statistical
  techniques and performance measures for genetics-based machine learning:
  Accuracy and interpretability, Soft Computing 13~(10) (2009) 959.

\bibitem{2009-Luengo-studyusestatistical}
J.~Luengo, S.~Garc{\'i}a, F.~Herrera, A study on the use of statistical tests
  for experimentation with neural networks: {{Analysis}} of parametric test
  conditions and non-parametric tests, Expert Systems with Applications 36~(4)
  (2009) 7798--7808.

\bibitem{2010-Garcia-Advancednonparametrictests}
S.~Garc{\'i}a, A.~Fern{\'a}ndez, J.~Luengo, F.~Herrera, Advanced nonparametric
  tests for multiple comparisons in the design of experiments in computational
  intelligence and data mining: {{Experimental}} analysis of power, Inf. Sci.
  180~(10) (2010) 2044--2064.
\newblock \href {https://doi.org/10.1016/j.ins.2009.12.010}
  {\path{doi:10.1016/j.ins.2009.12.010}}.

\bibitem{2010-Westfall-MultipleMcNemarTests}
P.~H. Westfall, J.~F. Troendle, G.~Pennello, Multiple {{McNemar Tests}},
  Biometrics 66~(4) (2010) 1185--1191.
\newblock \href {https://doi.org/10.1111/j.1541-0420.2010.01408.x}
  {\path{doi:10.1111/j.1541-0420.2010.01408.x}}.

\bibitem{2010-Rodriguez-SensitivityAnalysiskFold}
J.~Rodriguez, A.~Perez, J.~Lozano, Sensitivity {{Analysis}} of k-{{Fold Cross
  Validation}} in {{Prediction Error Estimation}}, IEEE Transactions on Pattern
  Analysis and Machine Intelligence 32~(3) (2010) 569--575.
\newblock \href {https://doi.org/10.1109/TPAMI.2009.187}
  {\path{doi:10.1109/TPAMI.2009.187}}.

\bibitem{2010-Ojala-Permutationtestsstudying}
M.~Ojala, G.~C. Garriga, Permutation tests for studying classifier performance,
  Journal of Machine Learning Research 11~(Jun) (2010) 1833--1863.

\bibitem{2011-Carrano-MulticriteriaStatisticalBased}
E.~G. Carrano, E.~F. Wanner, R.~H.~C. Takahashi, A {{Multicriteria Statistical
  Based Comparison Methodology}} for {{Evaluating Evolutionary Algorithms}},
  IEEE Transactions on Evolutionary Computation 15~(6) (2011) 848--870.
\newblock \href {https://doi.org/10.1109/TEVC.2010.2069567}
  {\path{doi:10.1109/TEVC.2010.2069567}}.

\bibitem{2011-Derrac-practicaltutorialuse}
J.~Derrac, S.~Garc{\'i}a, D.~Molina, F.~Herrera, A practical tutorial on the
  use of nonparametric statistical tests as a methodology for comparing
  evolutionary and swarm intelligence algorithms, Swarm and Evolutionary
  Computation 1~(1) (2011) 3 -- 18.

\bibitem{2012-Trawinski-Nonparametricstatisticalanalysis}
B.~Trawi{\'n}ski, M.~Sm{\k{e}}tek, Z.~Telec, T.~Lasota, Nonparametric
  statistical analysis for multiple comparison of machine learning regression
  algorithms, International Journal of Applied Mathematics and Computer Science
  22~(4) (Jan. 2012).
\newblock \href {https://doi.org/10.2478/v10006-012-0064-z}
  {\path{doi:10.2478/v10006-012-0064-z}}.

\bibitem{2012-Ulas-Costconsciouscomparisonsupervised}
A.~Ula{\c s}, O.~T. Y\i{}ld\i{}z, E.~Alpayd\i{}n, Cost-conscious comparison of
  supervised learning algorithms over multiple data sets, Pattern Recognition
  45~(4) (2012) 1772--1781.
\newblock \href {https://doi.org/10.1016/j.patcog.2011.10.005}
  {\path{doi:10.1016/j.patcog.2011.10.005}}.

\bibitem{2012-Irsoy-DesignAnalysisClassifier}
O.~Irsoy, O.~T. Yildiz, E.~Alpaydin, Design and {{Analysis}} of {{Classifier
  Learning Experiments}} in {{Bioinformatics}}: {{Survey}} and {{Case
  Studies}}, IEEE/ACM Transactions on Computational Biology and Bioinformatics
  9~(6) (2012) 1663--1675.
\newblock \href {https://doi.org/10.1109/TCBB.2012.117}
  {\path{doi:10.1109/TCBB.2012.117}}.

\bibitem{2012-Lacoste-Bayesiancomparisonmachine}
A.~Lacoste, F.~Laviolette, M.~Marchand, Bayesian comparison of machine learning
  algorithms on single and multiple datasets, in: Artificial {{Intelligence}}
  and {{Statistics}}, 2012, pp. 665--675.

\bibitem{2012-Brodersen-Bayesianmixedeffectsinference}
K.~H. Brodersen, C.~Mathys, J.~R. Chumbley, J.~Daunizeau, C.~S. Ong, J.~M.
  Buhmann, K.~E. Stephan, Bayesian mixed-effects inference on classification
  performance in hierarchical data sets, Journal of Machine Learning Research
  13~(Nov) (2012) 3133--3176.

\bibitem{2013-Yildiz-StatisticalTestsUsing}
O.~T. Y\i{}ld\i{}z, E.~Alpayd\i{}n, Statistical {{Tests Using
  Hinge}}/{$\epsilon$}-{{Sensitive Loss}}, in: E.~Gelenbe, R.~Lent (Eds.),
  Computer and {{Information Sciences III}}, {Springer London}, {London}, 2013,
  pp. 153--160.
\newblock \href {https://doi.org/10.1007/978-1-4471-4594-3_16}
  {\path{doi:10.1007/978-1-4471-4594-3_16}}.

\bibitem{2013-Bostanci-EvaluationClassificationAlgorithms}
B.~Bostanci, E.~Bostanci, An {{Evaluation}} of {{Classification Algorithms
  Using Mc Nemar}}'s {{Test}}, in: J.~C. Bansal, P.~K. Singh, K.~Deep, M.~Pant,
  A.~K. Nagar (Eds.), Proceedings of {{Seventh International Conference}} on
  {{Bio}}-{{Inspired Computing}}: {{Theories}} and {{Applications}}
  ({{BIC}}-{{TA}} 2012), Vol. 201, {Springer India}, {India}, 2013, pp. 15--26.
\newblock \href {https://doi.org/10.1007/978-81-322-1038-2_2}
  {\path{doi:10.1007/978-81-322-1038-2_2}}.

\bibitem{2014-Otero-Bootstrapanalysismultiple}
J.~Otero, L.~S{\'a}nchez, I.~Couso, A.~Palacios, Bootstrap analysis of multiple
  repetitions of experiments using an interval-valued multiple comparison
  procedure, Journal of Computer and System Sciences 80~(1) (2014) 88 -- 100.
\newblock \href {https://doi.org/http://dx.doi.org/10.1016/j.jcss.2013.03.009}
  {\path{doi:http://dx.doi.org/10.1016/j.jcss.2013.03.009}}.

\bibitem{2014-Yu-Blocked3x2CrossValidated}
W.~Yu, W.~Ruibo, J.~Huichen, L.~Jihong, Blocked 3x2 {{Cross}}-{{Validated}}
  {\emph{t}} -{{Test}} for {{Comparing Supervised Classification Learning
  Algorithms}}, Neural Computation 26~(1) (2014) 208--235.
\newblock \href {https://doi.org/10.1162/NECO_a_00532}
  {\path{doi:10.1162/NECO_a_00532}}.

\bibitem{2014-Garcia-statisticalanalysisparameters}
S.~Garc{\'i}a, J.~Derrac, S.~{Ram{\'i}rez-Gallego}, F.~Herrera, On the
  statistical analysis of the parameters' trend in a machine learning
  algorithm, Progress in Artificial Intelligence 3~(1) (2014) 51--53.
\newblock \href {https://doi.org/10.1007/s13748-014-0043-8}
  {\path{doi:10.1007/s13748-014-0043-8}}.

\bibitem{2014-Benavoli-BayesianWilcoxonsignedrank}
A.~Benavoli, G.~Corani, F.~Mangili, M.~Zaffalon, F.~Ruggeri, A {{Bayesian
  Wilcoxon}} signed-rank test based on the {{Dirichlet}} process, in:
  Proceedings of the 31th {{International Conference}} on {{Machine Learning}},
  {{ICML}} 2014, {{Beijing}}, {{China}}, 21-26 {{June}} 2014, 2014, pp.
  1026--1034.

\bibitem{2015-Benavoli-ImpreciseDirichletProcess}
A.~Benavoli, F.~Mangili, F.~Ruggeri, M.~Zaffalon, Imprecise {{Dirichlet Process
  With Application}} to the {{Hypothesis Test}} on the {{Probability That}}
  {{{\emph{X}}}} {$\leq$} {{{\emph{Y}}}}, Journal of Statistical Theory and
  Practice 9~(3) (2015) 658--684.
\newblock \href {https://doi.org/10.1080/15598608.2014.985997}
  {\path{doi:10.1080/15598608.2014.985997}}.

\bibitem{2015-Corani-Bayesianapproachcomparing}
G.~Corani, A.~Benavoli, A {{Bayesian}} approach for comparing cross-validated
  algorithms on multiple data sets, Machine Learning 100~(2-3) (2015) 285--304.
\newblock \href {https://doi.org/10.1007/s10994-015-5486-z}
  {\path{doi:10.1007/s10994-015-5486-z}}.

\bibitem{2015-Benavoli-StatisticalTestsJoint}
A.~Benavoli, C.~P. de~Campos, Statistical {{Tests}} for {{Joint Analysis}} of
  {{Performance Measures}}, in: Advanced {{Methodologies}} for {{Bayesian
  Networks}} - {{Second International Workshop}}, {{AMBN}} 2015, {{Yokohama}},
  {{Japan}}, {{November}} 16-18, 2015. {{Proceedings}}, 2015, pp. 76--92.
\newblock \href {https://doi.org/10.1007/978-3-319-28379-1_6}
  {\path{doi:10.1007/978-3-319-28379-1_6}}.

\bibitem{2015-Benavoli-Bayesiannonparametricprocedure}
A.~Benavoli, G.~Corani, F.~Mangili, M.~Zaffalon, A {{Bayesian}} nonparametric
  procedure for comparing algorithms, in: Proceedings of the 32nd
  {{International Conference}} on {{Machine Learning}}, {{ICML}} 2015,
  {{Lille}}, {{France}}, 6-11 {{July}} 2015, 2015, pp. 1264--1272.

\bibitem{2015-Wang-ConfidenceIntervalMeasure}
Y.~Wang, J.~Li, Y.~Li, R.~Wang, X.~Yang, Confidence {{Interval}} for {{F}}\_1
  {{Measure}} of {{Algorithm Performance Based}} on {{Blocked}} 3x2
  {{Cross}}-{{Validation}}, IEEE Transactions on Knowledge and Data Engineering
  27~(3) (2015) 651--659.
\newblock \href {https://doi.org/10.1109/TKDE.2014.2359667}
  {\path{doi:10.1109/TKDE.2014.2359667}}.

\bibitem{2015-Perolat-GeneralizingWilcoxonranksum}
J.~Perolat, I.~Couso, K.~Loquin, O.~Strauss, Generalizing the {{Wilcoxon}}
  rank-sum test for interval data, International Journal of Approximate
  Reasoning 56 (2015) 108--121.
\newblock \href {https://doi.org/10.1016/j.ijar.2014.08.001}
  {\path{doi:10.1016/j.ijar.2014.08.001}}.

\bibitem{2015-Singh-Statisticalvalidationmultiple}
P.~K. Singh, R.~Sarkar, M.~Nasipuri, Statistical validation of multiple
  classifiers over multiple datasets in the field of pattern recognition,
  International Journal of Applied Pattern Recognition 2~(1) (2015) 1--23.

\bibitem{2016-Gondara-Classifiercomparisonusing}
L.~Gondara, Classifier comparison using precision, arXiv preprint
  arXiv:1609.09471 (2016).

\bibitem{2016-Corani-Statisticalcomparisonclassifiers}
G.~Corani, A.~Benavoli, J.~Dem{\v s}ar, F.~Mangili, M.~Zaffalon, Statistical
  comparison of classifiers through {{Bayesian}} hierarchical modelling,
  Machine Learning (2016) 1--21.

\bibitem{2017-Eisinga-Exactpvaluespairwise}
R.~Eisinga, T.~Heskes, B.~Pelzer, M.~Te~Grotenhuis, Exact p-values for pairwise
  comparison of {{Friedman}} rank sums, with application to comparing
  classifiers, BMC Bioinformatics 18~(1) (Dec. 2017).
\newblock \href {https://doi.org/10.1186/s12859-017-1486-2}
  {\path{doi:10.1186/s12859-017-1486-2}}.

\bibitem{2017-Yu-NewKindNonparametric}
Z.~Yu, Z.~Wang, J.~You, J.~Zhang, J.~Liu, H.~Wong, G.~Han, A {{New Kind}} of
  {{Nonparametric Test}} for {{Statistical Comparison}} of {{Multiple
  Classifiers Over Multiple Datasets}}, IEEE Transactions on Cybernetics
  47~(12) (2017) 4418--4431.
\newblock \href {https://doi.org/10.1109/TCYB.2016.2611020}
  {\path{doi:10.1109/TCYB.2016.2611020}}.

\bibitem{2017-Eftimov-Comparingmultiobjectiveoptimization}
T.~Eftimov, P.~Koro{\v s}ec, B.~K. Seljak, Comparing multi-objective
  optimization algorithms using an ensemble of quality indicators with deep
  statistical comparison approach, in: 2017 {{IEEE Symposium Series}} on
  {{Computational Intelligence}} ({{SSCI}}), 2017, pp. 1--8.
\newblock \href {https://doi.org/10.1109/SSCI.2017.8280910}
  {\path{doi:10.1109/SSCI.2017.8280910}}.

\bibitem{2018-Calvo-BayesianInferenceAlgorithm}
B.~Calvo, J.~Ceberio, J.~A. Lozano, Bayesian {{Inference}} for {{Algorithm
  Ranking Analysis}}, in: Proceedings of the {{Genetic}} and {{Evolutionary
  Computation Conference Companion}}, {{GECCO}} '18, {ACM}, {New York, NY,
  USA}, 2018, pp. 324--325.
\newblock \href {https://doi.org/10.1145/3205651.3205658}
  {\path{doi:10.1145/3205651.3205658}}.

\bibitem{2019-Campelo-Samplesizeestimation}
F.~Campelo, F.~Takahashi, Sample size estimation for power and accuracy in the
  experimental comparison of algorithms, Journal of Heuristics 25~(2) (2019)
  305--338, 00002.
\newblock \href {https://doi.org/10.1007/s10732-018-9396-7}
  {\path{doi:10.1007/s10732-018-9396-7}}.

\bibitem{2019-Eftimov-novelstatisticalapproach}
T.~Eftimov, P.~Koro{\v s}ec, A novel statistical approach for comparing
  meta-heuristic stochastic optimization algorithms according to the
  distribution of solutions in the search space, Information Sciences 489
  (2019) 255--273, 00000.
\newblock \href {https://doi.org/10.1016/j.ins.2019.03.049}
  {\path{doi:10.1016/j.ins.2019.03.049}}.

\bibitem{2010-Pesarin-Permutationtestscomplex}
F.~Pesarin, L.~Salmaso, Permutation Tests for Complex Data: Theory,
  Applications and Software, {John Wiley \& Sons}, 2010.

\bibitem{2010-Nordstokke-newnonparametricLevene}
D.~W. Nordstokke, B.~D. Zumbo, A new nonparametric {{Levene}} test for equal
  variances, Psicol{\'o}gica 31~(2) (2010).

\bibitem{2010-Kasuya-Wilcoxonsignedrankstest}
E.~Kasuya, Wilcoxon signed-ranks test: Symmetry should be confirmed before the
  test, Animal Behaviour 79~(3) (2010) 765--767.
\newblock \href {https://doi.org/10.1016/j.anbehav.2009.11.019}
  {\path{doi:10.1016/j.anbehav.2009.11.019}}.

\bibitem{1946-Dixon-statisticalsigntest}
W.~J. Dixon, A.~M. Mood, The statistical sign test, Journal of the American
  Statistical Association 41~(236) (1946) 557--566.

\bibitem{1945-Wilcoxon-IndividualComparisonsRanking}
F.~Wilcoxon, Individual {{Comparisons}} by {{Ranking Methods}}, Biometrics
  Bulletin 1~(6) (1945) 80--83.
\newblock \href {https://doi.org/10.2307/3001968} {\path{doi:10.2307/3001968}}.

\bibitem{1981-Conover-Ranktransformationsbridge}
W.~J. Conover, R.~L. Iman, Rank transformations as a bridge between parametric
  and nonparametric statistics, The American Statistician 35~(3) (1981)
  124--129.

\bibitem{1965-Rhyne-TablesTreatmentsControl}
A.~L. Rhyne, R.~G.~D. Steel, Tables for a {{Treatments Versus Control Multiple
  Comparisons Sign Test}}, Technometrics 7~(3) (1965) 293--306.

\bibitem{1980-Iman-Approximationscriticalregion}
R.~L. Iman, J.~M. Davenport, Approximations of the critical region of the
  fbietkan statistic, Communications in Statistics - Theory and Methods 9~(6)
  (1980) 571--595, 00709.
\newblock \href {https://doi.org/10.1080/03610928008827904}
  {\path{doi:10.1080/03610928008827904}}.

\bibitem{1962-Hodges-RankMethodsCombination}
J.~L. Hodges, E.~L. Lehmann, Rank {{Methods}} for {{Combination}} of
  {{Independent Experiments}} in {{Analysis}} of {{Variance}}, The Annals of
  Mathematical Statistics 33~(2) (1962) 482--497.
\newblock \href {https://doi.org/10.1214/aoms/1177704575}
  {\path{doi:10.1214/aoms/1177704575}}.

\bibitem{1967-Quade-Rankanalysiscovariance}
D.~Quade, Rank analysis of covariance, Journal of the American Statistical
  Association 62~(320) (1967) 1187--1200.

\bibitem{2013-Berrar-Significancetestsconfidence}
D.~Berrar, J.~A. Lozano, Significance tests or confidence intervals: Which are
  preferable for the comparison of classifiers?, Journal of Experimental \&
  Theoretical Artificial Intelligence 25~(2) (2013) 189--206.
\newblock \href {https://doi.org/10.1080/0952813X.2012.680252}
  {\path{doi:10.1080/0952813X.2012.680252}}.

\bibitem{1982-Seldrup-Geigyscientifictables}
J.~Seldrup, C.~Lentner, K.~Diem, Geigy Scientific Tables: {{Introduction}} to
  Statistics, Statistical Tables, Mathematical Formulae, {Ciba-Geigy}, 1982.

\bibitem{2012-Good-Commonerrorsstatistics}
P.~I. Good, J.~W. Hardin, Common Errors in Statistics (and How to Avoid Them),
  {John Wiley \& Sons}, 2012.

\bibitem{2010-Kruschke-Bayesiandataanalysis}
J.~K. Kruschke, Bayesian data analysis, Wiley Interdisciplinary Reviews:
  Cognitive Science 1~(5) (2010) 658--676.

\bibitem{2002-Willems-robustHotellingtest}
G.~Willems, G.~Pison, P.~J. Rousseeuw, S.~Van~Aelst, A robust {{Hotelling}}
  test, Metrika 55~(1) (2002) 125--138, 00058.
\newblock \href {https://doi.org/10.1007/s001840200192}
  {\path{doi:10.1007/s001840200192}}.

\bibitem{2009-VillasenorAlva-generalizationShapiroWilk}
J.~A. Villasenor~Alva, E.~G. Estrada, A generalization of
  {{Shapiro}}\textendash{{Wilk}}'s test for multivariate normality,
  Communications in Statistics\textemdash{}Theory and Methods 38~(11) (2009)
  1870--1883, 00126.

\bibitem{2016-deCampos-JointAnalysisMultiple}
C.~P. {de Campos}, A.~Benavoli, Joint {{Analysis}} of {{Multiple Algorithms}}
  and {{Performance Measures}}, New Generation Computing 35~(1) (2016) 69--86.
\newblock \href {https://doi.org/10.1007/s00354-016-0005-8}
  {\path{doi:10.1007/s00354-016-0005-8}}.

\bibitem{2017-Kumar-Improvinglocalsearcha}
A.~Kumar, R.~K. Misra, D.~Singh, Improving the local search capability of
  {{Effective Butterfly Optimizer}} using {{Covariance Matrix Adapted Retreat
  Phase}}, in: 2017 {{IEEE Congress}} on {{Evolutionary Computation}}
  ({{CEC}}), 2017, pp. 1835--1842.
\newblock \href {https://doi.org/10.1109/CEC.2017.7969524}
  {\path{doi:10.1109/CEC.2017.7969524}}.

\bibitem{2017-Brest-Singleobjectiverealparameter}
J.~Brest, M.~S. Mau{\v c}ec, B.~Bo{\v s}kovi{\'c}, Single objective
  real-parameter optimization: Algorithm {{jSO}}, in: Evolutionary
  {{Computation}} ({{CEC}}), 2017 {{IEEE Congress}} On, {IEEE}, 2017, pp.
  1311--1318.

\bibitem{2017-Awad-Ensemblesinusoidaldifferential}
N.~H. Awad, M.~Z. Ali, P.~N. Suganthan, Ensemble sinusoidal differential
  covariance matrix adaptation with {{Euclidean}} neighborhood for solving
  {{CEC2017}} benchmark problems, in: Evolutionary {{Computation}} ({{CEC}}),
  2017 {{IEEE Congress}} On, {IEEE}, 2017, pp. 372--379.

\bibitem{2017-Mohamed-LSHADEsemiparameteradaptation}
A.~W. Mohamed, A.~A. Hadi, A.~M. Fattouh, K.~M. Jambi, {{LSHADE}} with
  semi-parameter adaptation hybrid with {{CMA}}-{{ES}} for solving {{CEC}} 2017
  benchmark problems, in: Evolutionary {{Computation}} ({{CEC}}), 2017 {{IEEE
  Congress}} On, {IEEE}, 2017, pp. 145--152.

\bibitem{2017-Jagodzinski-differentialevolutionstrategy}
D.~Jagodzi{\'n}ski, J.~Arabas, A differential evolution strategy, in:
  Evolutionary {{Computation}} ({{CEC}}), 2017 {{IEEE Congress}} On, {IEEE},
  2017, pp. 1872--1876.

\bibitem{2017-Sallam-Multimethodbasedorthogonal}
K.~M. Sallam, S.~M. Elsayed, R.~A. Sarker, D.~L. Essam, Multi-method based
  orthogonal experimental design algorithm for solving {{CEC2017}} competition
  problems, in: Evolutionary {{Computation}} ({{CEC}}), 2017 {{IEEE Congress}}
  On, {IEEE}, 2017, pp. 1350--1357.

\bibitem{2017-Bujok-Enhancedindividualdependentdifferential}
P.~Bujok, J.~Tvrd{\'i}k, Enhanced individual-dependent differential evolution
  with population size adaptation, in: Evolutionary {{Computation}} ({{CEC}}),
  2017 {{IEEE Congress}} On, {IEEE}, 2017, pp. 1358--1365.

\bibitem{2017-Biedrzycki-versionIPOPCMAESalgorithm}
R.~Biedrzycki, A version of {{IPOP}}-{{CMA}}-{{ES}} algorithm with midpoint for
  {{CEC}} 2017 single objective bound constrained problems, in: Evolutionary
  {{Computation}} ({{CEC}}), 2017 {{IEEE Congress}} On, {IEEE}, 2017, pp.
  1489--1494.

\bibitem{2017-LaTorre-comparisonthreelargescale}
A.~LaTorre, J.-M. Pe{\~n}a, A comparison of three large-scale global optimizers
  on the {{CEC}} 2017 single objective real parameter numerical optimization
  benchmark, in: Evolutionary {{Computation}} ({{CEC}}), 2017 {{IEEE Congress}}
  On, {IEEE}, 2017, pp. 1063--1070.

\bibitem{2017-Tangherloni-ProactiveParticlesSwarm}
A.~Tangherloni, L.~Rundo, M.~S. Nobile, Proactive {{Particles}} in {{Swarm
  Optimization}}: {{A}} settings-free algorithm for real-parameter single
  objective optimization problems, in: Evolutionary {{Computation}} ({{CEC}}),
  2017 {{IEEE Congress}} On, {IEEE}, 2017, pp. 1940--1947.

\bibitem{2017-Maharana-DynamicYinYangPair}
D.~Maharana, R.~Kommadath, P.~Kotecha, Dynamic {{Yin}}-{{Yang Pair
  Optimization}} and its performance on single objective real parameter
  problems of {{CEC}} 2017, in: Evolutionary {{Computation}} ({{CEC}}), 2017
  {{IEEE Congress}} On, {IEEE}, 2017, pp. 2390--2396.

\bibitem{2017-Kommadath-TeachingLearningBased}
R.~Kommadath, P.~Kotecha, Teaching {{Learning Based Optimization}} with focused
  learning and its performance on {{CEC2017}} functions, in: Evolutionary
  {{Computation}} ({{CEC}}), 2017 {{IEEE Congress}} On, {IEEE}, 2017, pp.
  2397--2403.

\bibitem{2017-Berrar-JeffreysLindleyParadoxLooming}
D.~Berrar, W.~Dubitzky, On the {{Jeffreys}}-{{Lindley Paradox}} and the
  {{Looming Reproducibility Crisis}} in {{Machine Learning}}, in: Data
  {{Science}} and {{Advanced Analytics}} ({{DSAA}}), 2017 {{IEEE International
  Conference}} On, {IEEE}, 2017, pp. 334--340.

\bibitem{1998-Chow-Precisstatisticalsignificance}
S.~L. Chow, Pr{\'e}cis of statistical significance: {{Rationale}}, validity,
  and utility, Behavioral and brain sciences 21~(2) (1998) 169--194.

\bibitem{2016-Melinscak-pvaluesevaluationbrain}
F.~Melinscak, L.~Montesano, Beyond p-values in the evaluation of
  brain\textendash{}computer interfaces: {{A Bayesian}} estimation approach,
  Journal of neuroscience methods 270 (2016) 30--45.

\bibitem{2019-Amrhein-Scientistsrisestatistical}
V.~Amrhein, S.~Greenland, B.~McShane, Scientists rise up against statistical
  significance, Nature 567~(7748) (2019) 305, 00002.
\newblock \href {https://doi.org/10.1038/d41586-019-00857-9}
  {\path{doi:10.1038/d41586-019-00857-9}}.

\bibitem{2016-Wasserstein-ASAStatementpValues}
R.~L. Wasserstein, N.~A. Lazar, The {{ASA}}'s {{Statement}} on p-{{Values}}:
  {{Context}}, {{Process}}, and {{Purpose}}, The American Statistician 70~(2)
  (2016) 129--133.
\newblock \href {https://doi.org/10.1080/00031305.2016.1154108}
  {\path{doi:10.1080/00031305.2016.1154108}}.

\bibitem{2019-Wasserstein-MovingWorld05}
R.~L. Wasserstein, A.~L. Schirm, N.~A. Lazar, Moving to a {{World Beyond}} ``p
  {$<$} 0.05'', The American Statistician 73~(sup1) (2019) 1--19, 00003.
\newblock \href {https://doi.org/10.1080/00031305.2019.1583913}
  {\path{doi:10.1080/00031305.2019.1583913}}.

\bibitem{2018-Silva-correspondencefrequentistBayesian}
I.~R. Silva, On the correspondence between frequentist and {{Bayesian}} tests,
  Communications in Statistics - Theory and Methods 47~(14) (2018) 3477--3487.
\newblock \href {https://doi.org/10.1080/03610926.2017.1359296}
  {\path{doi:10.1080/03610926.2017.1359296}}.

\bibitem{2017-Couso-ReconcilingBayesianFrequentist}
I.~Couso, A.~{\'A}lvarez-Caballero, L.~S{\'a}nchez, Reconciling {{Bayesian}}
  and {{Frequentist Tests}}: The {{Imprecise Counterpart}}, in: A.~Antonucci,
  G.~Corani, I.~Couso, S.~Destercke (Eds.), Proceedings of the {{Tenth
  International Symposium}} on {{Imprecise Probability}}: {{Theories}} and
  {{Applications}}, Vol.~62 of Proceedings of {{Machine Learning Research}},
  {PMLR}, 2017, pp. 97--108.

\end{thebibliography}

\end{document}